%% file: boundary_iou.tex
\documentclass[final]{cvpr}

\usepackage{times,epsfig,graphicx,amsmath,amssymb,bm}

\usepackage{booktabs,caption,subcaption,makecell,multirow,tabulary}
\usepackage[export]{adjustbox}
\usepackage[table]{xcolor}
\usepackage{collcell}
\makeatletter
\@namedef{ver@everyshi.sty}{}
\makeatother
\usepackage{pgf}

\PassOptionsToPackage{hyphens}{url}\usepackage[pagebackref=true,breaklinks=true,letterpaper=true,colorlinks,citecolor=citecolor,bookmarks=false]{hyperref}
\definecolor{citecolor}{RGB}{34,139,34}
\newcommand{\x}{\times}
\newcommand{\dt}[1]{\fontsize{7pt}{0.1em}\selectfont (#1)}
\newcommand{\bd}[1]{\textbf{#1}}
\newcommand{\bolden}[1]{{\boldmath\bfseries#1}}
\newcolumntype{x}[1]{>{\centering\arraybackslash}p{#1pt}}

\usepackage{pifont}%
\newcommand{\cmark}{\ding{51}}%
\newcommand{\xmark}{\ding{55}}%

\newlength\savewidth\newcommand\shline{\noalign{\global\savewidth\arrayrulewidth
  \global\arrayrulewidth 1pt}\hline\noalign{\global\arrayrulewidth\savewidth}}
\newcommand{\tablestyle}[2]{\setlength{\tabcolsep}{#1}\renewcommand{\arraystretch}{#2}\centering\footnotesize}

\def\eg{\emph{e.g}\onedot} 
\def\ie{\emph{i.e}\onedot} 
 
 \def\vs{\emph{vs}\onedot}
\def\wrt{w.r.t\onedot}

\newcommand{\gc}{G_{1}}
\newcommand{\gb}{G_{d}}
\newcommand{\pc}{P_{1}}
\newcommand{\pb}{P_{d}}

\newcommand{\app}{\raise.17ex\hbox{$\scriptstyle\sim$}}

\makeatletter
\renewcommand\paragraph{\@startsection{paragraph}{4}{\z@}%
  {.5em \@plus1ex \@minus.1ex}%
  {-.5em}%
  {\normalfont\normalsize\bfseries}}
\makeatother

\def\cvprPaperID{607} %
\def\confYear{CVPR 2021}

\begin{document}

\title{Boundary IoU: Improving Object-Centric Image Segmentation Evaluation}

\author{
  Bowen Cheng$^{1}$\thanks{Work done during an internship at Facebook AI Research.} \quad Ross Girshick$^{2}$ \quad Piotr Dollár$^{2}$ \quad Alexander C. Berg$^{2}$ \quad Alexander Kirillov$^{2}$\\[2mm]
  $^1$UIUC  \qquad $^2$Facebook AI Research (FAIR)
}

\maketitle

\begin{abstract}
  We present Boundary IoU (Intersection-over-Union), a new segmentation evaluation measure focused on boundary quality. We perform an extensive analysis across different error types and object sizes and show that Boundary IoU is significantly more sensitive than the standard Mask IoU measure to boundary errors for large objects and does not over-penalize errors on smaller objects. The new quality measure displays several desirable characteristics like symmetry \wrt prediction/ground truth pairs and balanced responsiveness across scales, which makes it more suitable for segmentation evaluation than other boundary-focused measures like Trimap IoU and F-measure.
  Based on Boundary IoU, we update the standard evaluation protocols for instance and panoptic segmentation tasks by proposing the Boundary AP (Average Precision) and Boundary PQ (Panoptic Quality) metrics, respectively. Our experiments show that the new evaluation metrics track boundary quality improvements that are generally overlooked by current Mask IoU-based evaluation metrics. We hope that the adoption of the new boundary-sensitive evaluation metrics will lead to rapid progress in segmentation methods that improve boundary quality.\footnote{Project page: \scriptsize \url{https://bowenc0221.github.io/boundary-iou}}
\end{abstract}

\section{Introduction}

The Common Task Framework~\cite{liberman2015}, in which standardized tasks, datasets, and evaluation metrics are used to track research progress, yields impressive results. For example, researchers working on the instance segmentation task, which requires an algorithm to delineate objects with pixel-level binary masks, have improved the standard Average Precision (AP) metric on COCO~\cite{lin2014coco} by an astonishing 86\% (relative) from 2015~\cite{dai2016instance} to 2019~\cite{li2020joint}.

However, this progress is not equal across all error modes, because different evaluation metrics are sensitive (or insensitive) to different types of errors. If a metric is used for a prolonged time, as in the Common Task Framework, then the corresponding sub-field most rapidly resolves the types of errors to which this metric is sensitive. Research directions that improve other error types typically advance more slowly, as such progress is harder to quantify.

This phenomenon is at play in instance segmentation, where, among the multitude of papers contributing to the impressive 86\% relative improvement in AP (\eg,~\cite{zhu2019deformable,cai2018cascade,bodla2017soft,huang2019mask,coco2018winner}), only a few address mask boundary quality. 

\begin{figure}[!t]
  \centering
  \bgroup
  \def\arraystretch{0.5}
  \setlength\tabcolsep{12pt}
  \begin{tabular}{cc}
  \includegraphics[width=0.40\linewidth]{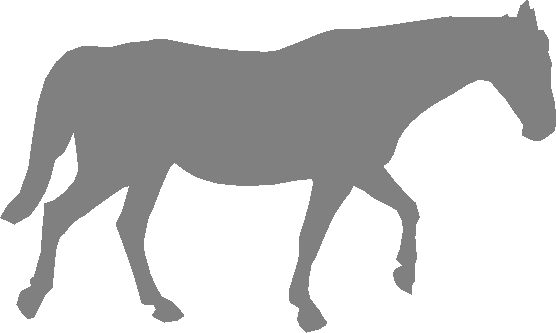} &
  \includegraphics[width=0.40\linewidth]{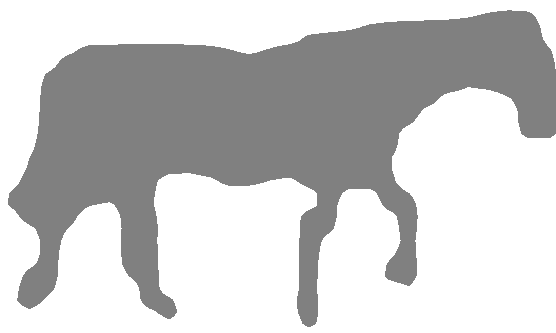} \\
  \scriptsize Ground Truth (LVIS~\cite{gupta2019lvis}) & \scriptsize Mask R-CNN~\cite{he2017mask}\\
  \includegraphics[width=0.40\linewidth]{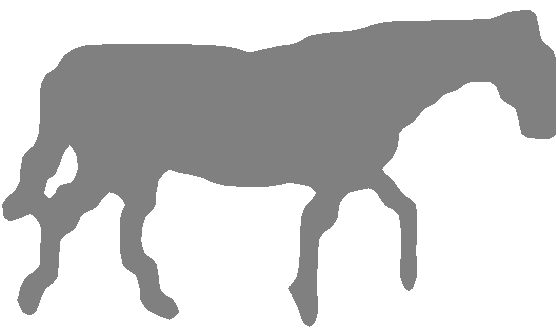} &
  \includegraphics[width=0.40\linewidth]{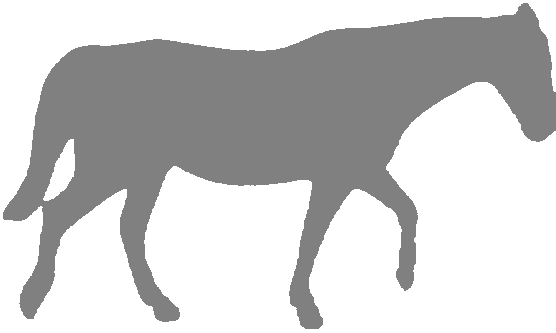}\\
  \scriptsize BMask R-CNN~\cite{ChengWHL20bmaskrcnn} & \scriptsize PointRend~\cite{kirillov2020pointrend} \\
  \end{tabular} \egroup

  \vspace{0.5mm}

  \tablestyle{5pt}{1.1}
  \begin{tabular}{lx{50}x{50}x{50}}
  & Mask R-CNN & BMask R-CNN & PointRend \\
  \shline
  Mask IoU & 89\% & 92\% \dt{+3\%} & 97\% \dt{+8\%}\phantom{0} \\
  Boundary IoU & 69\% & 78\% \bd{\dt{+9\%}} & 91\% \bd{\dt{+22\%}} \\
  \end{tabular}
  \vspace{-3mm}
  \caption{Given the bounding box for a horse, the mask predicted by Mask R-CNN scores a high Mask IoU value (89\%) relative to the ground truth despite having low-fidelity, blobby boundaries. The recently proposed BMask R-CNN~\cite{ChengWHL20bmaskrcnn} and PointRend~\cite{kirillov2020pointrend} methods predict masks with higher fidelity boundaries, yet these clear visual improvements only marginally improve Mask IoU (+3\% and +8\%, respectively). In contrast, our proposed \textbf{Boundary IoU} measure demonstrates greater sensitivity to boundary errors, and thus provides a clear, quantitative gradient that rewards improvements to boundary segmentation quality.}
  \label{fig:teaser}
  \vspace{-4mm}
\end{figure}

Note that mask boundary quality is an essential aspect of image segmentation, as various downstream applications directly benefit from more precise object segmentations~\cite{zhang2020perceiving,sengupta2020synthetic,sharma2020compositional}. However, the dominant family of Mask R-CNN-based methods~\cite{he2017mask} are well-known to predict low-fidelity, blobby masks (see Figure~\ref{fig:teaser}). This observation suggests that the current evaluation metrics may have limited sensitivity to mask prediction errors near object boundaries.

To understand why, we start by analyzing Mask Intersection-over-Union (Mask IoU), the underlying measure used in AP to compare predicted and ground truth masks. Mask IoU divides the intersection area of two masks by the area of their union. This measure values all pixels equally and, therefore, is less sensitive to boundary quality in \emph{larger objects}: the number of interior pixels grows quadratically in object size and can far exceed the number of boundary pixels, which only grows linearly. 
In this paper we aim to identify a measure for image segmentation that is sensitive to boundary quality across all scales.

Towards this goal we start by studying standard segmentation measures like Mask IoU and boundary-focused measures such as Trimap IoU~\cite{kohli2009robust,deeplabV2} and F-measure~\cite{martin2003empirical,csurka2013good,perazzi2016davis}. We study error-sensitivity characteristics of each measure by generating a variety of error types on top of the high-quality ground truth masks from the LVIS dataset~\cite{gupta2019lvis}. Our analysis confirms that Mask IoU is less sensitive to errors in larger objects. In addition, the analysis reveals limitations of existing boundary-focused measures, such as asymmetries and instability to small changes in mask quality.

Based on these insights we propose a new \textbf{Boundary IoU} measure. Boundary IoU is simple and easy to compute. Instead of considering all pixels, it calculates the intersection-over-union for mask pixels within a certain distance from the corresponding ground truth or prediction boundary contours. Our analysis demonstrates that Boundary IoU measures boundary quality of large objects well, unlike Mask IoU, and it does not over-penalize errors on small objects. An illustrative examples compares Boundary IoU to Mask IoU in Figure~\ref{fig:teaser}.

Boundary IoU enables new task-level evaluation metrics. For the task of instance segmentation~\cite{lin2014coco}, we propose \textbf{Boundary Average Precision} (Boundary AP), and for panoptic segmentation~\cite{kirillov2017panoptic}, we propose \textbf{Boundary Panoptic Quality} (Boundary PQ).

Boundary AP assesses all relevant aspects of instance segmentation, simultaneously taking into account categorization, localization, and segmentation quality, unlike prior boundary-focused metrics for instance segmentation like AF~\cite{liang2020polytransform} that ignore false positive rates. We test Boundary AP on three common datasets: COCO~\cite{lin2014coco}, LVIS~\cite{gupta2019lvis}, and Cityscapes~\cite{Cordts2016Cityscapes}. With real predictions from recent instance segmentation methods that directly aim to improve boundary quality~\cite{kirillov2020pointrend,ChengWHL20bmaskrcnn}, we verify that Boundary AP tracks improvements better than Mask AP. With synthetic predictions, we show that Boundary AP is significantly more sensitive to large-object boundary quality than Mask AP.

For panoptic segmentation, we apply Boundary PQ to the COCO~\cite{kirillov2017panoptic} and Cityscapes~\cite{Cordts2016Cityscapes} panoptic datasets. We test the new metric with synthetic predictions and show that it is more sensitive than the previous metric based on Mask IoU. Finally, we evaluate the performance of various recent instance and panoptic segmentation models with the new evaluation metrics to ease comparison for future research.

These new metrics reveal improvements in boundary quality that are generally ignored by Mask IoU-based evaluation metrics. We hope that the adoption of these new boundary-sensitive evaluations can enable faster progress towards segmentation models with better boundary quality.

\section{Related Work and Preliminaries}

Image segmentation tasks like semantic, instance, or panoptic segmentation are evaluated by comparing segmentation masks predicted by a system to ground truth masks provided by annotators. Modern evaluation metrics for these tasks are based on segmentation quality measures that evaluate consistency between ground truth object shape $G$ and prediction shape $P$ represented by binary masks of a fixed resolution.
We define the most common segmentation quality measures and the new Boundary IoU measure in Table~\ref{tab:measurement_summary} using the unified notation presented in Table~\ref{tab:notations}. We split the measures into mask- and boundary-based types and discuss their differences next.

\begin{table*}[t!]
\tablestyle{6.5pt}{1.4}
\begin{tabular}{l|l|c|c|c|c}
Name & Type & Definition & Symmetric & Preference & Insensitivity \\
\shline\rule{0mm}{5mm}
Pixel accuracy & mask-based & \makecell{\rule{0mm}{5mm}$\dfrac{|G \cap P|}{|G|}$} & \xmark & \makecell{larger\\prediction} & $-$ \\[2mm]
\hline\rule{0mm}{5mm}
Mask IoU & mask-based & $\dfrac{|G \cap P|}{|G \cup P|}$ & \cmark & $-$ & \makecell{boundary\\errors} \\[2mm]
\hline\rule{0mm}{5mm}
Trimap IoU & boundary-based & $\dfrac{|\gb \cap (G \cap P)|}{|\gb \cap (G \cup P)|} = \dfrac{|(\gb \cap G) \cap P|}{|(\gb \cap G) \cup (\gb \cap P)|}$ & \xmark & \makecell{larger\\prediction} & \makecell{errors far from\\ground truth boundary} \\[2.5mm]
\hline\rule{0mm}{5mm}
F-measure & boundary-based & 
\makecell{\rule{0mm}{5mm}
$\dfrac{2 \cdot \tilde{p} \cdot \tilde{r}}{\tilde{p} + \tilde{r}}$\vspace{1mm}, \,
$\tilde{p} = \dfrac{|\pc \cap \gb|}{|\pc|}$, \, $\tilde{r} = \dfrac{|\gc \cap \pb|}{|\gc|}$
} & \cmark & $-$ & \makecell{errors on\\small objects} \\
\hline\rule{0mm}{5mm}
\textbf{Boundary IoU} & boundary-based & $\dfrac{|(\gb \cap G) \cap (\pb \cap P)|}{|(\gb \cap G) \cup (\pb \cap P)|}$ & \cmark & $-$ & \makecell{errors far from predicted\\ and ground truth boundaries}
\end{tabular}
\caption{Existing segmentation measures and the new Boundary IoU defined with the unified notation from Table~\ref{tab:notations}. For each measure we detail several properties. \textbf{Symmetric}: whether the swap of ground truth and prediction masks changes measure's value. \textbf{Preference}: whether the measure biases better scores to a certain type of prediction. \textbf{Insensitivity}: types of errors the measure is less sensitive to.}
\label{tab:measurement_summary}
\vspace{-2mm}
\end{table*}

\begin{table}[t!]
  \begin{center}
  \tablestyle{5pt}{1.2}
  \begin{tabular}{l|l}
  Notation & Definition \\
  \shline
  $G$ & ground truth binary mask \\
  $P$ & prediction binary mask \\
  \hline
  $\gc$, $\pc$ & set of pixels on the contour line of the binary mask \\
  $\gb$, $\pb$ & set of pixels in the boundary region of the binary mask \\[0.5mm]
  \hline
  $d$ & pixel width of the boundary region
  \end{tabular}
  \caption{Notation used in this paper. We define a contour as the 1d line comprised of the set of mask pixels that touches the background. The boundary is a 2D region consisting of pixels within pixel distance $d$ from the contour pixels. A boundary region can be constructed by dilating the contour line by $d$ pixels.}
  \label{tab:notations}
  \end{center}
  \end{table}

\paragraph{Mask-based segmentation measures}
take into account all pixels of an object mask. The first PASCAL VOC semantic segmentation track in 2007~\cite{pascal-voc-2007} used Pixel Accuracy measure to evaluate predictions. For each class it calculates the ratio of correctly labeled ground truth pixels (see Table~\ref{tab:measurement_summary}).
Pixel accuracy is not symmetric and biased toward prediction masks that are larger than ground truth masks. Subsequently, PASCAL VOC~\cite{everingham2015pascal} switched its evaluation to the Mask Intersection-over-Union (Mask IoU) measure.

Mask IoU segmentation consistency measure divides the number of pixels in the intersection of the prediction and ground truth masks by the number of pixels in their union (see Table~\ref{tab:measurement_summary}).
The measure is widely used in the evaluation metrics for most popular semantic, instance, and panoptic segmentation tasks~\cite{everingham2015pascal,lin2014coco,kirillov2017panoptic} and datasets~\cite{Cordts2016Cityscapes,caesar2016coco,zhou2017ade20k,lin2014coco}. Unlike Pixel Accuracy, Mask IoU is symmetric, however, as we will show in this paper, it demonstrates unbalanced responsiveness to the boundary quality across object sizes.

\paragraph{Boundary-based segmentation measures} evaluate segmentation quality by estimating contour alignment between predicted and ground truth masks. Unlike mask-based measures, these measures only evaluate the pixels that lie directly on the masks' contours or in their close proximity.

Trimap IoU~\cite{kohli2009robust,deeplabV2} is a boundary-based segmentation measure that calculates IoU in a narrow band of pixels within a pixel distance $d$ from the contour of the ground truth mask (see Table~\ref{tab:measurement_summary}).
In contrast to Mask IoU, Trimap IoU reacts similarly to comparable pixel errors across object scales because it calculates IoU only for pixels around the contour. However, unlike Mask IoU, the measure is not symmetric and favors predictions whose masks are larger than the corresponding ground truth masks. Moreover, the measure ignores prediction errors that appear outside the band around the ground truth contour.

F-measure was initially proposed for edge detection~\cite{martin2004learning}, but it is also used to evaluate segmentation quality~\cite{csurka2013good,perazzi2016davis}. $\text{F-measure} = 2 \cdot p \cdot r / \left(p + r\right)$, where $p$ and $r$ denote precision and recall. In the original definition, 
$p$ and $r$ are calculated by matching prediction and ground truth contour pixels within the pixel distance threshold $d$ via bipartite matching. However, the matching process is computationally expensive for high resolution masks and large datasets and, therefore,~\cite{csurka2013good,perazzi2016davis} proposed an approximation procedure to compute the precision and recall by allowing duplicate matches, which we denote by $\tilde{p}$ and $\tilde{r}$.
In this case, $\tilde{p}$ computes the ratio of pixels in the prediction contour that lie within a distance $d$ from the ground truth contours, whereas $\tilde{r}$ computes a similar ratio for the pixels of the ground truth contour, see Table~\ref{tab:measurement_summary}. In the rest of the paper we use the approximate formulation of F-measure. The measure is symmetric and tolerates small contour misalignments that can be attributed to ambiguity, however, it ignores significant errors when the object size is comparable to $d$. For example, this occurs with reasonable choices of $d$ and small objects commonly found in datasets (\eg, COCO~\cite{lin2014coco}).

Trimap and F-measure are often used to evaluate boundary quality for semantic segmentation tasks in an ad-hoc fashion. For example, Trimap IoU is used as an extra evaluation to show boundary quality improvement~\cite{deeplabV2,marin2019efficient}, but it is not reported by most segmentation methods. In the next section we will study both measures in detail and analyze their behavior across different error types and object sizes.

\section{Sensitivity Analysis}\label{sec:sensitivity}

In~\S\ref{sec:analysis} and~\S\ref{sec:boundaryiou} we will compare several mask consistency measures by observing how a measure's value changes in response to errors of different magnitudes. We will observe and interpret these curves to draw conclusions about the behavior of these measures, a methodology that we refer to as \emph{sensitivity analysis}.

To enable a systematic comparison, we \emph{simulate} a set of common segmentation errors across different mask sizes by generating pseudo-predictions from ground truth annotations. This approach allows us explicitly control the type and severity of the errors used in the analysis. Moreover, the use of pseudo-predictions avoids any bias toward specific segmentation models which makes the analysis more robust and general. A potential limitation of this approach is that simulated errors may not fully represent errors created by real models. We aim to counteract this limitation by using a diverse set of error types. Figure~\ref{fig:error_types} depicts an example of each error type we consider:\\
  \indent $\bullet$ \textbf{Scale error.} Dilation/erosion are applied to the ground truth masks. The error severity is controlled by the kernel radius of the morphological operations.\\
  \indent $\bullet$ \textbf{Boundary localization error.} Random Gaussian noise is added to the coordinate of each vertex in polygons that represent ground truth masks. The error severity is controlled by the standard deviation (std) of the Gaussian noise.\\
  \indent $\bullet$ \textbf{Object localization error.} Ground truth masks are shifted with random offsets. The error severity is controlled by the pixel length of the shift.\\
  \indent $\bullet$ \textbf{Boundary approximation error.} The \texttt{simplify} function from Shapely~\cite{shapely} removes vertices from the polygons that represent ground truth masks while keeping the simplified polygons as close to the original ones as possible. The error severity is controlled by the error tolerance parameter of the \texttt{simplify} function.\\
  \indent $\bullet$ \textbf{Inner mask error.} Holes of random shape are added to ground truth masks. The error severity is controlled by the number of holes added. While this error type is not common for modern segmentation approaches, we include it to assess the effect of interior mask errors.

\input{fig_tex/vis_error_type/single_fig}

\paragraph{Implementation details.}
For the analysis, we randomly sample instance masks from the LVIS v0.5~\cite{gupta2019lvis} validation set. The dataset is selected due to its high-quality annotations. Using these masks, for each segmentation error type we create multiple sets of pseudo-predictions by varying the severity of the error. To analyze a segmentation measure, we report its mean and standard deviation across a set of pseudo-predictions that represent a given error type of a fixed severity. We will also compare segmentation measures across different object sizes by generating a separate set of pseudo-predictions using ground truth objects within a specific mask area range. For all boundary-based measures that use pixel distance parameter, $d$, we set it to $2\%$ of the image diagonal for fair comparison.

\section{Analysis of Existing Segmentation Measures}\label{sec:analysis}

First, we analyze the standard Mask IoU segmentation consistency measure from both theoretical and empirical perspectives. Then, we study two existing alternatives -- Trimap IoU and F-measure boundary-based measures.

\subsection{Mask IoU}

\paragraph{Theoretical analysis.} Mask IoU is scale-invariant \wrt object area. For a fixed Mask IoU value, a larger object will have more incorrect pixels and the change in incorrect pixel count grows in proportional to the change in object area (as Mask IoU is a ratio of areas). However, when scaling up a typical object, the number of interior pixels grows quadratically, whereas the number of contour pixels only grows linearly. These different asymptotic growth rates cause Mask IoU to tolerate a larger number of misclassified pixels \emph{per each unit of contour length} for a larger object.

\begin{figure}[t]
  \centering
  \bgroup
  \def\arraystretch{0.2}
  \setlength\tabcolsep{0.2pt}
  \begin{tabular}{cc}
  \includegraphics[height=0.116\textheight,trim={0cm 0cm 0cm 3cm},clip]{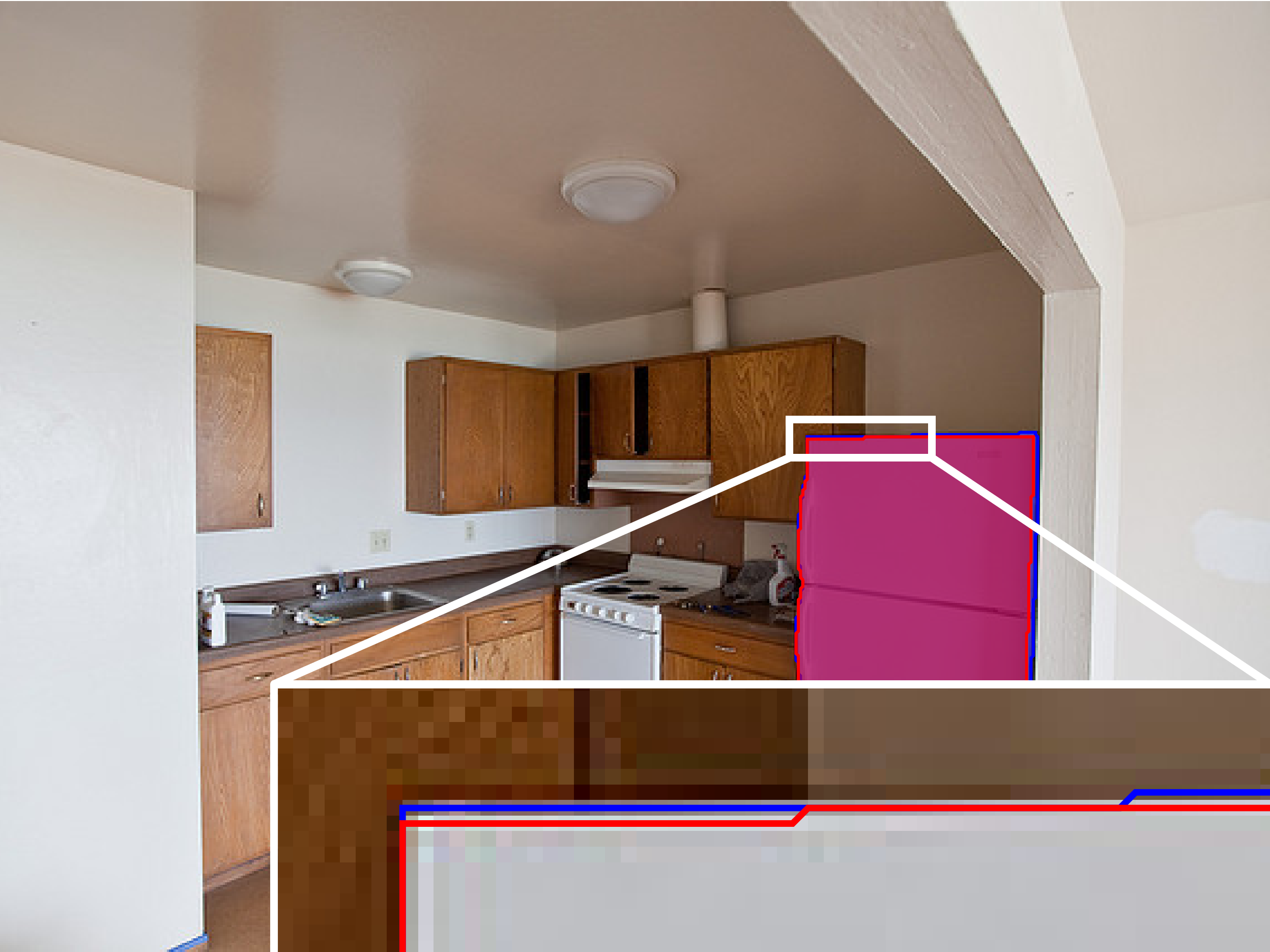} &
  \includegraphics[height=0.116\textheight,trim={0cm 0cm 0cm 3cm},clip]{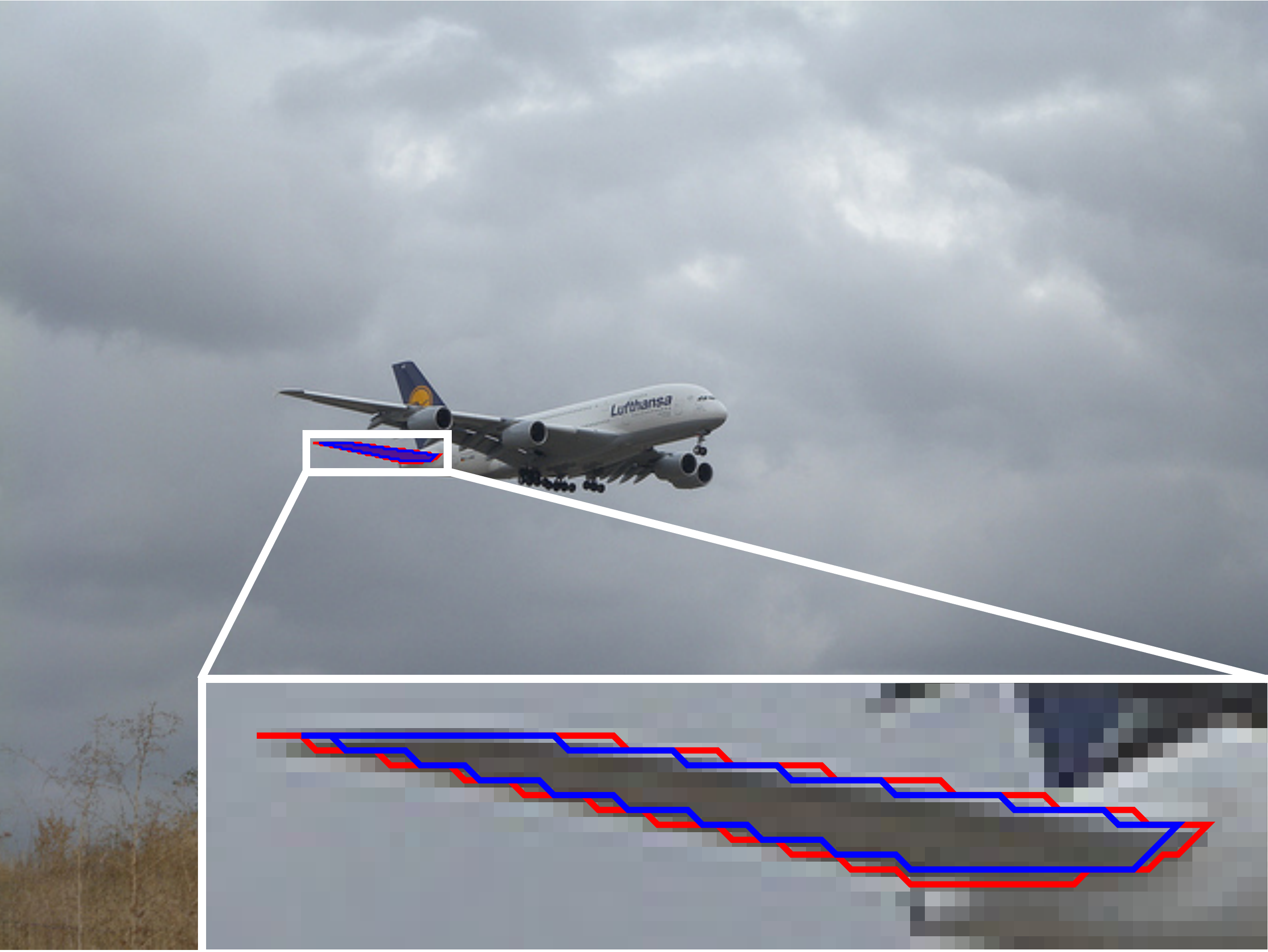} \\
  \end{tabular} \egroup
  \caption{The two objects from LVIS~\cite{gupta2019lvis} dataset annotated independently twice. While the fridge (left) has more than 100 times larger area than the wing (right), on the crop of the same resolution the discrepancy between annotations is visually similar. This simple example supports the observation that the boundary quality in the ground truth is independent of the object size. Mask IoU counts all pixels equally and gives a higher score to fridge, 0.97, vs wing, 0.81, due to the larger internal region with consistent labeling. The bias toward larger objects render Mask IoU inadequate for boundary quality evaluation across object sizes. In contrast, the new Boundary IoU yields much closer scores (0.87 \vs 0.81).}
  \label{fig:boundary_ambiguity}
\end{figure}

\paragraph{Empirical analysis.} This property corresponds to an assumption that boundary localization error in ground truth annotations (\ie, intrinsic annotation ambiguity) also grows with the object size. However, a classic study on multi-region segmentation~\cite{martin2003empirical} shows that the pixel distance between two contours of the same object labeled by different annotators seldomly exceeds 1\% of the image diagonal, irrespective of the object size. We confirm this observation by exploring double annotations that are provided for a subset of images in LVIS~\cite{gupta2019lvis}. In Figure~\ref{fig:boundary_ambiguity} we present a random pair of objects with significant size difference. While one of the objects is 100 times larger, the boundary discrepancy within the cropped part, which has the same resolution, is similar between the two objects. Observed results suggest that boundary ambiguity is fixed and independent of objects area. This is likely a consequence of the annotation tool, which includes the ability to zoom while drawing contours.

\input{fig_tex/analysis/trimap_iou_and_f_measure/single_fig}

Using simulated scale errors (described in~\S\ref{sec:sensitivity}) we confirm Mask IoU's bias in favor of large objects. The dilation/erosion of the ground truth mask by a fixed number of pixels significantly decreases Mask IoU for small objects while Mask IoU grows as object area increases (see Figure~\ref{fig:mask_iou_problem}). Note that Mask IoU's insensitivity to boundary errors on large objects cannot be addressed by simply increasing the lowest Mask IoU threshold in evaluation metrics like AP or PQ. Such a change does not remedy the bias and will lead to relative over-penalization for smaller objects.

\subsection{Boundary-Based Measures} Next, we will analyze the boundary-based measures Trimap IoU and F-measure. These measures focus on pixels within a distance $d$ from object contours. The parameter $d$ is usually fixed on the dataset~\cite{deeplabV2} or image level~\cite{perazzi2016davis} which results in these measures treating boundary localization errors independently of the size of the object. By matching the natural characteristic of the ground truth segmentation data, these boundary-based measures are better suited to evaluate improvements in boundary quality across object sizes.

\paragraph{Trimap IoU} computes IoU for a region around the ground truth boundary only (\ie, the region is independent of the prediction), and therefore is not symmetric: swapping the prediction and ground truth masks will give a different score. In Figure~\ref{fig:trimap_iou_problem} we show that this asymmetry favors predictions that are larger than ground truth masks. For larger (dilated) pseudo-predictions Trimap IoU does not drop below some positive value irrespective of the error severity, whereas for smaller (eroded) pseudo-predictions it drops to 0. Moreover, the measure ignores any errors outside of the ground truth boundary region, penalizing inner mask errors less than Mask IoU (see the appendix for details).

\paragraph{F-Measure} matches the pixels of the predicted and ground truth contours if they are within the pixels distance threshold $d$. Hence, it ignores small contour misalignments that can be attributed to ambiguity. While robustness to ambiguity is good in principle, in Figure~\ref{fig:f_measure_problem} we observe that F-measure can be nearly discontinuous, rapidly stepping from 1 to 0 when the error severity changes by a small amount. Sharp response curves can lead to task metrics with high variance. In comparison, Mask IoU is more continuous. Further, $d$ may be large relative to small objects, causing F-measure to award significant errors a perfect score.

\paragraph{Discussion.} Given the limitations presented above, we conclude that neither Trimap IoU nor F-measure can replace Mask IoU as the main segmentation consistency measure for a broad range of evaluation metrics. At the same time, Mask IoU is biased towards large objects in a way that discourages improvements to boundary segmentations. Next, we propose \emph{Boundary IoU} as a new measure to evaluate segmentation boundary-quality that does not have any of the previously mentioning limitations.

\begin{figure}[t]
  \centering
  \includegraphics[width=0.90\linewidth]{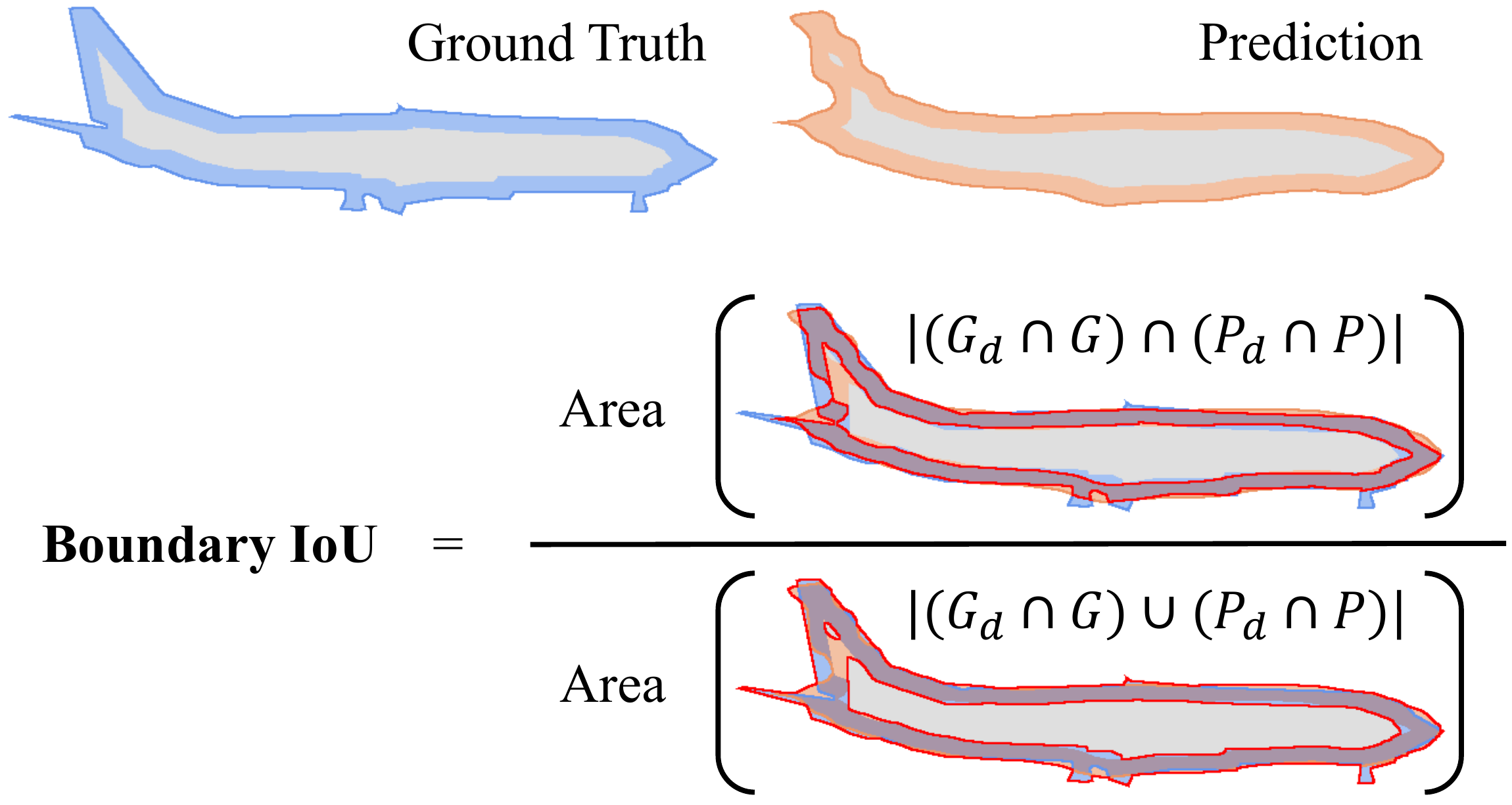}
  \vspace{-3mm}
  \caption{\textbf{Boundary IoU computation illustration}. (top) Ground truth and prediction masks. For both, we highlight \emph{mask} pixels that are within distance $d$ from the contours. (bottom) Boundary IoU segmentation consistency measures computes the intersection-over-union between the highlighted regions. In this example, Boundary IoU is 0.73, whereas Mask IoU is 0.91.}
\label{c}
\vspace{-2mm}
\end{figure}

\section{Boundary IoU}\label{sec:boundaryiou}

\input{fig_tex/analysis/boundary_iou/separate_fig}

In this section we introduce a new segmentation measure and compare it with existing consistency measures using simulated errors. The new measure should have a weaker bias toward large objects than Mask IoU. Furthermore, we aim for a measure that neither over-penalizes nor ignores errors in small objects similarly to Mask IoU.

Guided by these principals, we propose the \emph{Boundary IoU} segmentation consistency measure. The new measure is simultaneously simple and satisfies the principals charted above. Given two masks $G$ and $P$, Boundary IoU first computes the set of the original masks' pixels that are within distance $d$ from each contour, and then computes the intersection-over-union of these two sets (see Figure~\ref{c}). Using the notation from Table~\ref{tab:notations}:
\begin{align}
  \text{Boundary IoU}(G, P) = \frac{|(\gb \cap G) \cap (\pb \cap P)|}{|(\gb \cap G) \cup (\pb \cap P)|},
\end{align}
where boundary regions $\gb$ and $\pb$ are the sets of all pixels within $d$ pixels distance from the ground truth and prediction contours respectively. Note, that the new measure evaluates only \emph{mask} pixels that are within pixel distance $d$ from the contours. A simpler version with IoU calculated directly for boundary regions $\gb$ and $\pb$ loses information about sharp contour corners that are smoothed by considering all pixels within distance $d$ from the contours.

The distance to the contour parameter $d$ in Boundary IoU controls the sensitivity of the measure. With $d$ large enough to include all pixels inside the prediction and ground truth masks, Boundary IoU becomes equivalent to Mask IoU. With a smaller parameter $d$ Boundary IoU ignores interior mask pixels, which makes it more sensitive to the boundary quality than Mask IoU for larger objects. For smaller objects, Boundary IoU is very close or even equivalent to Mask IoU depending on the parameter $d$ which prevents it from over-penalizing errors in smaller objects where the number of inner pixels is comparable with the number of pixels close to the contours.

In Figure~\ref{fig:biou_large_objects} and Figure~\ref{fig:biou_across_scales} we present the results of our analysis for Boundary IoU. The analysis shows that Boundary IoU is less biased than Mask IoU towards large object sizes across all considered error types (Figure~\ref{fig:biou_large_objects}). Varying object sizes while keeping error severities constant (Figure~\ref{fig:biou_across_scales}), Boundary IoU behaves identically to Mask IoU for smaller objects avoiding over-penalization for any error type. For larger objects Boundary IoU reduces the bias that Mask IoU exhibits and its value grows more slowly with the object area given fixed error severities.

\paragraph{Comparison with Trimap IoU.} We note that the new measure appears quite similar to Trimap IoU (see Table~\ref{tab:measurement_summary}). However, unlike Trimap IoU, Boundary IoU considers pixels close to the contours of both prediction and ground truth together. This simple change remedies two main limitations of Trimap IoU. The new measure is symmetric and penalizes the errors that appear away from the ground truth boundary (see Figure~\ref{fig:biou_large_objects} (a), (b), and (f)).

\paragraph{Comparison with F-measure.} F-measure uses hard matching between the contours of the predicted and ground truth masks. If the distance between contour pixels is within $d$ pixels, then both precision and recall are perfect, but once the distance goes beyond $d$ the matching does not happen at all. In contrast, Boundary IoU evaluates consistency in a soft manner. Intersection over union is 1.0 if two contours are perfectly aligned and as the contours diverge Boundary IoU gradually decreases. In the appendix, we also compare Boundary IoU with a soft generalization of F-measure that averages multiple scores across different parameters $d$. Our analysis shows that it under-penalizes errors in small objects in comparison with Mask IoU and Boundary IoU.

\paragraph{The pixel distance parameter \bolden{$d$}.} If $d$ is large enough Boundary IoU is equivalent to Mask IoU. On the other hand, if $d$ is too small, Boundary IoU severely penalizes even the smallest misalignment ignoring possible ambiguity of the contours. To select $d$ that does not over-penalize possible ambiguity of the contours, we use Boundary IoU to measure the consistency of two expert annotators who independently delineated the same objects. The creators of LVIS~\cite{gupta2019lvis} have collected such expert annotations for images in COCO~\cite{lin2014coco} and ADE20k~\cite{zhou2017ade20k} datasets. Both datasets have images of similar resolution and we find that median Boundary IoU between the annotations of the two experts exceeds 0.9 for both datasets when $d$ equals 2\% of an image diagonal (15 pixels distance on average). For Cityscapes~\cite{Cordts2016Cityscapes} that has higher resolution images and excellent annotation quality we suggest to use the same distance in pixels which results in $d$ set to $0.5\%$ of an image diagonal for the dataset.

For other datasets, we suggest two considerations for selecting the pixel distance $d$: (1) the annotation consistency sets the lower bound on $d$, and (2) $d$ should be selected based on the performance of current methods and decreased as performance improves.

\paragraph{Limitations of Boundary IoU.} The new measure does not evaluate mask pixels that are further than $d$ pixels away from the corresponding ground truth or prediction boundary. Therefore, it can award a perfect score to non-identical masks. For example, Boundary IoU is perfect for a disc-shaped mask and a ring-shaped mask that has the same center and outer radius as the disk, plus the inner radius that is exactly $d$ pixels smaller than the outer one (we show this example in the appendix). For these two masks, all non-matching pixels of the disc-shaped mask are further than $d$ pixels away from its boundary. To penalize such cases, we suggest a simple combination of Mask IoU and Boundary IoU by taking their minimum. In our experiments with both real and simulated predictions, we found that Boundary IoU is smaller or equal to Mask IoU in the absolute majority of cases ($99.9\%$) and the inequality is violated when a prediction with accurate boundaries misses interior part of an object (similar to the toy example above).

\section{Applications}\label{sec:apps}

The most common evaluation metrics for instance and panoptic segmentation tasks are Average Precision (AP or Mask AP)~\cite{lin2014coco} and Panoptic Quality (PQ or Mask PQ)~\cite{kirillov2017panoptic} respectively. Both metrics use Mask IoU and inherit its bias toward large objects and, subsequently, the insensitivity to the boundary quality observed before in~\cite{kirillov2020pointrend,wang2019object}.

We update the evaluation metrics for these tasks by replacing Mask IoU with \textbf{min(Mask IoU, Boundary IoU)} as suggested in the previous section. We name the new evaluation metrics Boundary AP and Boundary PQ. The change is simple to implement and we demonstrate that the new metrics are more sensitive to the boundary quality while able to track other types of improvements in predictions as well.

We hope that adoption of the new evaluation metrics will allow rapid progress of boundary quality in segmentation methods. We present our results for instance segmentation in the main text and refer to the appendix for our analysis of Boundary PQ.

\paragraph{Boundary AP for instance segmentation.} The goal of the instance segmentation task is to delineate each object with a pixel-level mask. An evaluation metric for the task is simultaneously assessing multiple aspects such as categorization, localization, and segmentation quality. To compare different evaluation metrics, we conduct our experiments with both synthetic predictions and real instance segmentation models. Synthetic predictions allow us to assess the segmentation quality aspect in isolation, whereas real predictions provide insights into the ability of Boundary AP to track all aspects of the instance segmentation task.

We compare Mask AP and Boundary AP on the COCO instance segmentation dataset~\cite{lin2014coco} in the main text. In addition, our findings are supported by similar experiments on Cityscapes~\cite{Cordts2016Cityscapes} and LVIS~\cite{gupta2019lvis} presented in the appendix. Detailed description of all datasets can be found in the appendix, along with Boundary AP evaluation for various recent and classic models on all three datasets. These results can be used as a reference to simplify the comparison for future methods.

\input{tables/ins_seg_exp1}

\input{tables/ins_seg_exp2}

\subsection{Evaluation on Synthetic Predictions}

Using synthetic predictions we evaluate the segmentation quality aspect of instance segmentation in isolation without a bias that any particular model can have. We simulate predictions by capping the effective resolution of each mask. First, we downscale cropped ground truth masks to a $28 \x 28$ resolution\footnote{This is a popular prediction resolution used in practice~\cite{he2017mask}.} mask with continuous values, we then upscale it back using bilinear interpolation, and finally binarize it. Such synthetic masks are close to the ground truth masks for smaller objects, however the discrepancy grows with object size. In Table~\ref{tab:coco_gt_mask_resolution} we report overall AP and AP\textsubscript{$S$}, AP\textsubscript{$M$}, and AP\textsubscript{$L$} for object size splits defined in COCO~\cite{lin2014coco}. Mask AP follows the behavior of Mask IoU, showing little sensitivity to the error growth between AP$_S$ and AP$_L$. In contrast, Boundary AP successfully captures the difference with significantly lower score for larger objects. In the appendix, we provide an example of the synthetic predictions and more results using different effective resolutions.

\subsection{Evaluation on Real Predictions}

We use outputs of existing segmentation models to further study Boundary AP. Unless specified, to isolate the segmentation quality from categorization and localization errors for purposes of analysis, we supply ground truth boxes to these methods and assign a random confidence score to each box. We use Detectron2~\cite{wu2019detectron2} with a ResNet-50~\cite{he2016deep} backbone unless otherwise specified. 

\paragraph{Mask AP \vs Boundary AP.} Table~\ref{tab:inference_with_gt_box_a} shows both Mask AP and Boundary AP for the standard Mask R-CNN model~\cite{he2017mask}. Mask R-CNN is well-known to predict blobby masks with significant visual defects for larger objects (see Figure~\ref{fig:teaser}). Nevertheless, Mask AP\textsubscript{$L$} is larger than Mask AP\textsubscript{$S$}. In contrast, we observe that Boundary AP\textsubscript{$L$} is smaller than Boundary AP\textsubscript{$S$} for Mask R-CNN suggesting that the new measure is more sensitive to the boundary quality of the large objects. Note that in this experiment the use of ground truth boxes removes any categorization and localization errors that are usually larger for small objects. 

\paragraph{Segmentation \vs categorization and localization.} A general evaluation metric for instance segmentation should track the improvements in all aspects of the task including segmentation, categorization, and localizations. In Table~\ref{tab:inference_with_gt_box_b} we first evaluate Mask R-CNN with several backbones (ResNet-50, ResNet-101, and ResNeXt-101-32$\times$8d~\cite{xie2017aggregated}), again supplying ground truth boxes. Note that both Mask AP and Boundary AP do not change significantly with different backbones, suggesting that more powerful backbones do not directly influence the segmentation quality. Next, we evaluate Boundary AP requiring each model to predict its own boxes as is standard. We observe that Boundary AP is able to track improvements from better localization and categorization similarly to Mask AP.

\paragraph{Mask quality improvements.} We explore Boundary AP's ability to capture the improvements in segmentation quality by the methods designed for this purpose in Tables~\ref{tab:inference_with_gt_box_c} and~\ref{tab:inference_with_gt_box_d}. To compare the segmentation quality aspect across models we again supply ground truth boxes to each model.

PointRend~\cite{kirillov2020pointrend} was developed to improve pixel-level prediction quality of models like Mask R-CNN and can produce predictions of varying resolution. PointRend significantly improves mask quality, while this can be measured via mask AP, it is more pronounced in Boundary AP, especially for large objects and for a higher resolution PointRend variant. See Table~\ref{tab:inference_with_gt_box_c} for details.

Boundary-preserving Mask R-CNN~\cite{ChengWHL20bmaskrcnn} (BMask R-CNN) improves boundary quality by adding a direct boundary supervision loss and increasing the resolution of feature maps used in its mask head. In Table~\ref{tab:inference_with_gt_box_d} Boundary AP shows that BMask R-CNN with its $28 \x 28$ output resolution outperforms PointRend for small objects, whereas for larger objects the $224 \x 224$ resolution output of PointRend is preferable, which matches a subjective visual quality assessment (see an example in Figure~\ref{fig:teaser}). We hope that the improved sensitivity of the new Boundary AP metric will lead to a rapid progress in the methods that improve boundary quality for instance segmentation.

\section{Conclusion}
Unlike the standard Mask IoU, Boundary IoU segmentation quality measure provides a clear, quantitative gradient that rewards improvements to boundary segmentation quality. We hope that the new measure will challenge our community to develop new methods with high-fidelity mask predictions. In addition, Boundary IoU allows a more granular analysis of segmentation-related errors for the complex multifaceted tasks like instance and panoptic segmentation. Incorporation of the measure in the performance analysis tools like TIDE~\cite{bolya2020tide} can provide better insights into specific error types of instance segmentation models.

\newpage
\appendix
\begin{center}{\bf \Large Appendix}\end{center}
\input{appendix}

{\small
\bibliographystyle{ieee_fullname}
\bibliography{egbib}
}

\end{document}

%% file: fig_tex/vis_error_type/single_fig.tex
\begin{figure}
    \centering
        \centering
        \bgroup 
        \def\arraystretch{0.2} 
        \setlength\tabcolsep{3pt}
        \begin{tabular}{ccc}
        \includegraphics[height=0.10\textheight,trim={4cm 3.5cm 1cm 3.5cm},clip]{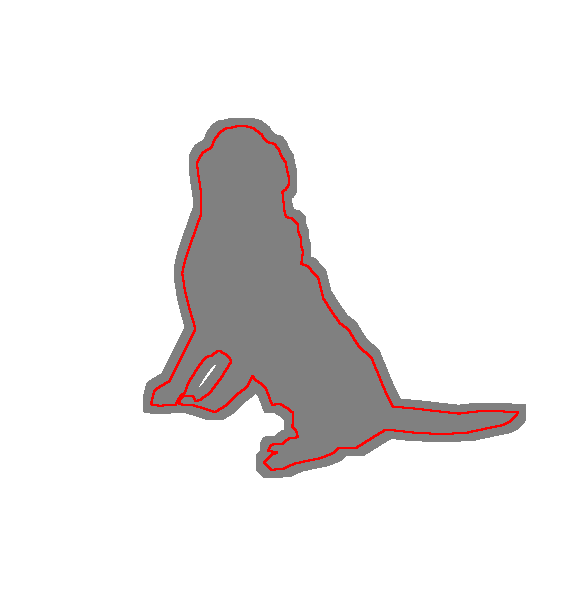} &
        \includegraphics[height=0.10\textheight,trim={4cm 3.5cm 1cm 3.5cm},clip]{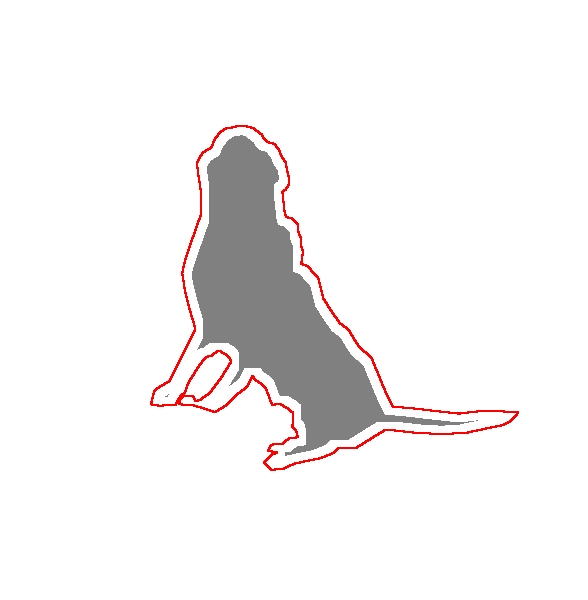} & 
        \includegraphics[height=0.10\textheight,trim={4cm 3.5cm 1cm 3.5cm},clip]{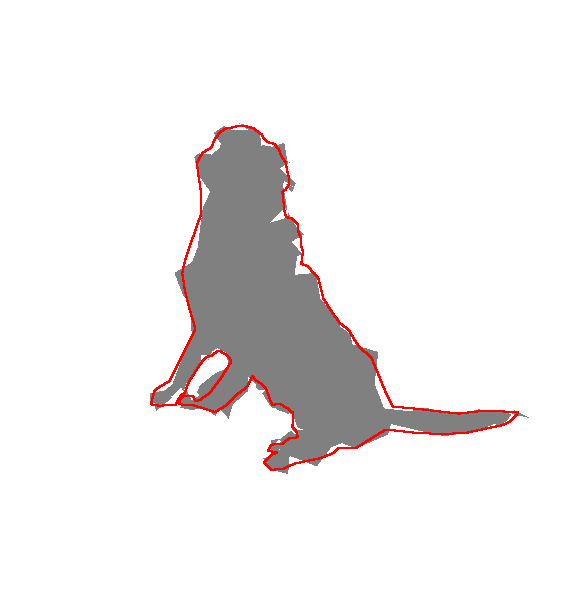} \\
        \scriptsize (a) Scale (dilation) & \scriptsize (b) Scale (erosion) & \scriptsize (c) Boundary localization \\
        \includegraphics[height=0.10\textheight,trim={4cm 3.5cm 1cm 3.5cm},clip]{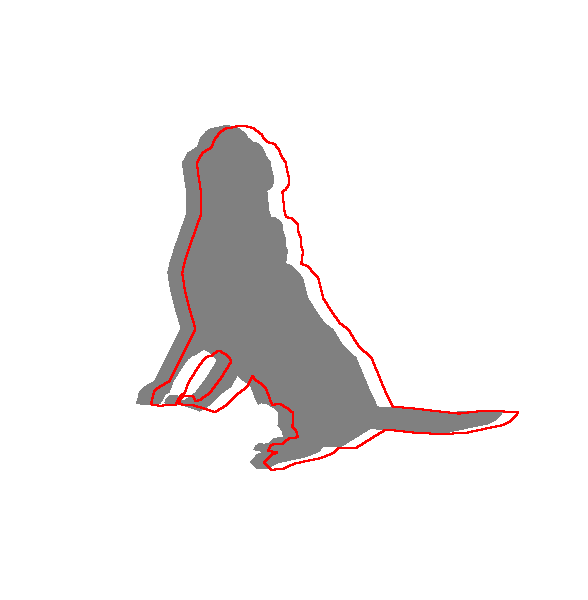} & 
        \includegraphics[height=0.10\textheight,trim={4cm 3.5cm 1cm 3.5cm},clip]{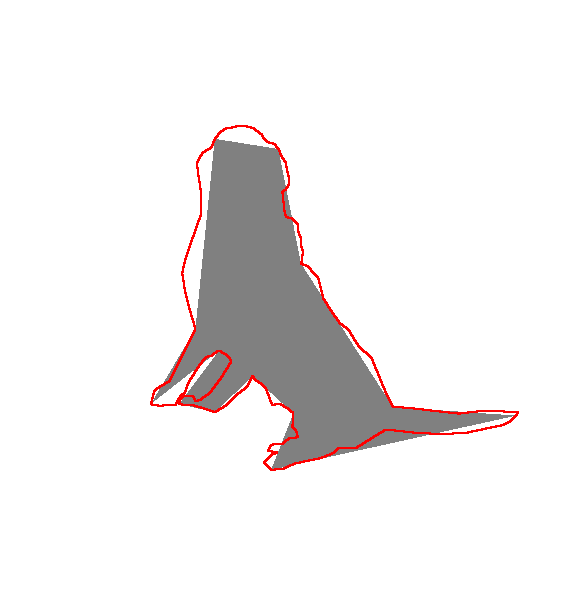} &
        \includegraphics[height=0.10\textheight,trim={4cm 3.5cm 1cm 3.5cm},clip]{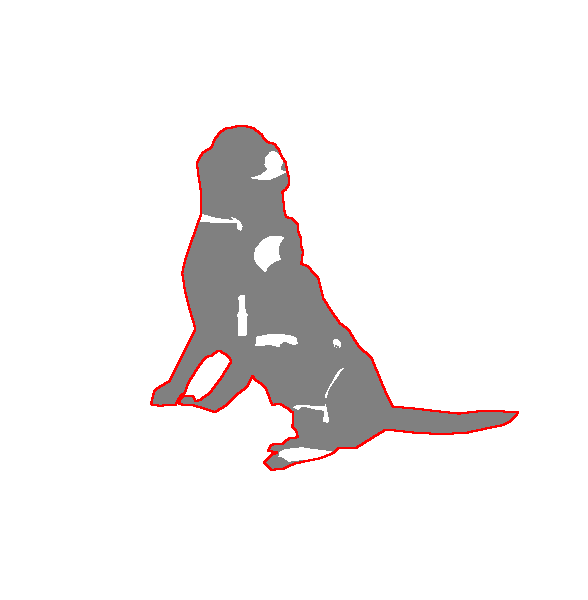} \\
        \scriptsize (d) Object localization & \scriptsize (e) Boundary approximation & \scriptsize (f) Inner mask\\
        \end{tabular} \egroup 
    \caption{\textbf{Examples of error types} generated from a single ground truth mask. The red contour represents the ground truth contour. \emph{Scale errors}: (a) dilation and (b) erosion of the mask. \emph{Boundary localization error}: (c) adding random Gaussian noise to each vertex in polygons. \emph{Object localization error}: (d) shifting masks. \emph{Boundary approximation error}: (e) simplifying polygons. \emph{Inner mask error}: (f) adding holes to masks.}
    \label{fig:error_types}
\end{figure}

%% file: fig_tex/analysis/trimap_iou_and_f_measure/single_fig.tex
\begin{figure*}
    \centering
    \begin{subfigure}[t]{0.33\linewidth}
    \centering
    \bgroup 
    \def\arraystretch{0.5} 
    \setlength\tabcolsep{0.5pt}
    \begin{tabular}{ccc}
    \multirow{3}{*}{\rotatebox[origin=c]{90}{\scriptsize Measure value \hspace{4pt}}} & \scriptsize Dilation (5 pixels) & \scriptsize Erosion (5 pixels) \\
    & \includegraphics[width=0.47\linewidth,trim={1.0cm 1.1cm 0cm 0cm},clip]{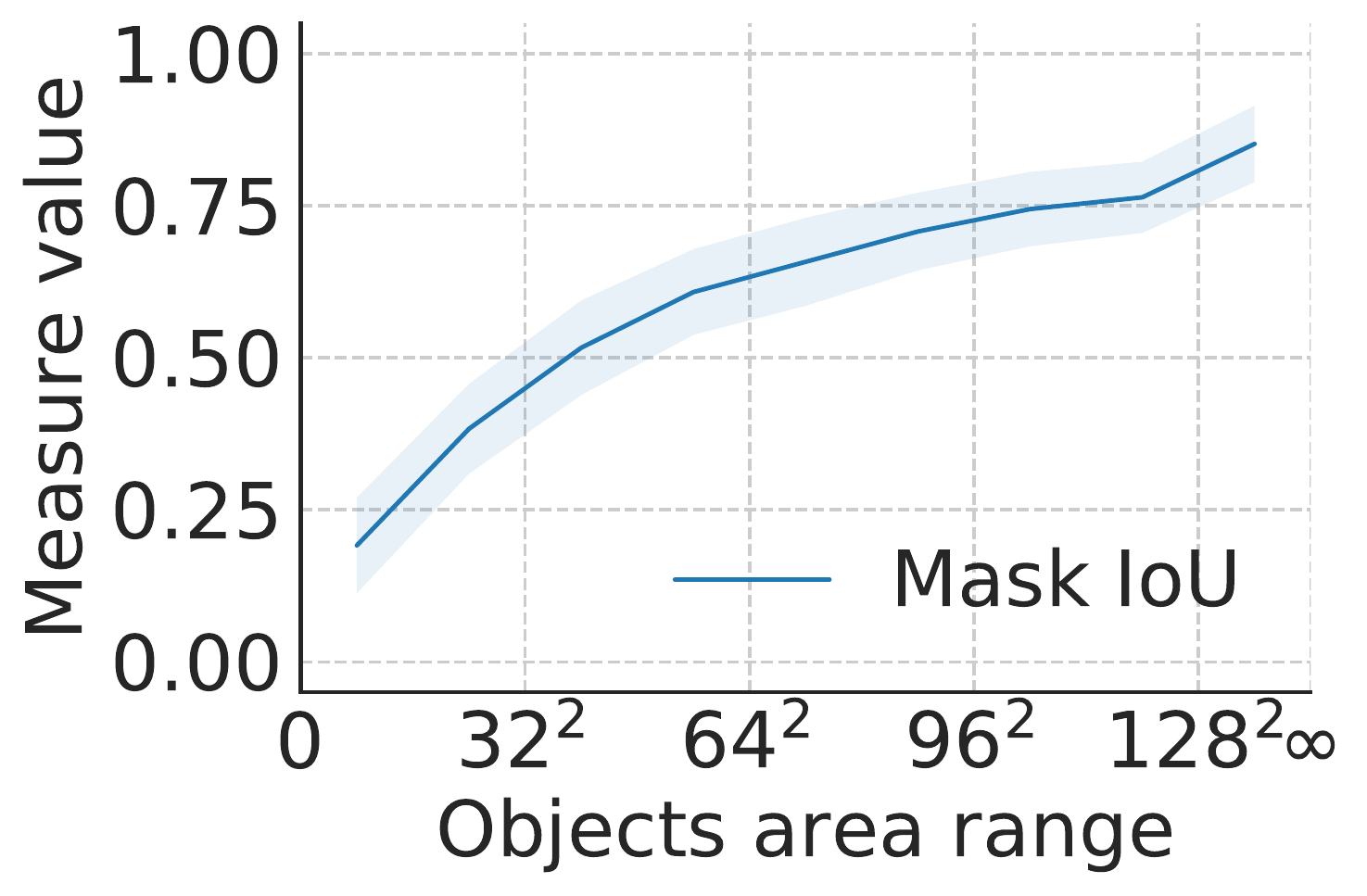} & \includegraphics[width=0.47\linewidth,trim={1.0cm 1.1cm 0cm 0cm},clip]{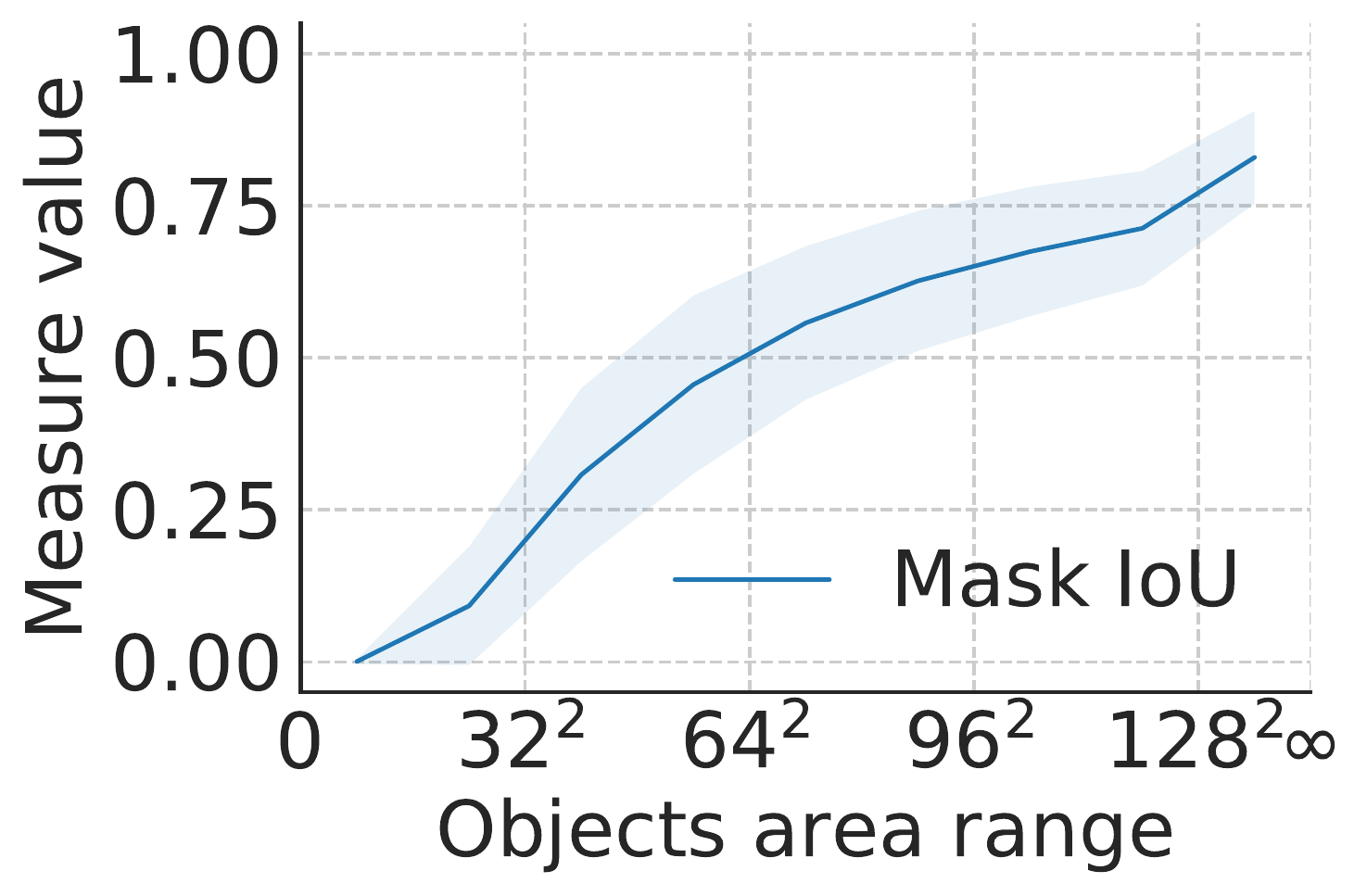} \\
    & \multicolumn{2}{c}{\scriptsize Objects area bins with $16^2$ increment} \\
    \end{tabular} \egroup 
    \vspace{-2mm}
    \subcaption{Mask IoU is biased toward large objects. The pseudo-predictions for larger objects receive higher score under a fixed error severity.}
    \label{fig:mask_iou_problem}
    \end{subfigure}
    \hfill
    \begin{subfigure}[t]{0.31\linewidth}
    \centering
    \bgroup 
    \def\arraystretch{0.5} 
    \setlength\tabcolsep{0.5pt}
    \begin{tabular}{cc}
    \scriptsize object area $> 96^2$ & \scriptsize object area $\le 32^2$  \\
    \includegraphics[width=0.5\linewidth,trim={1.0cm 1.1cm 0cm 0cm},clip]{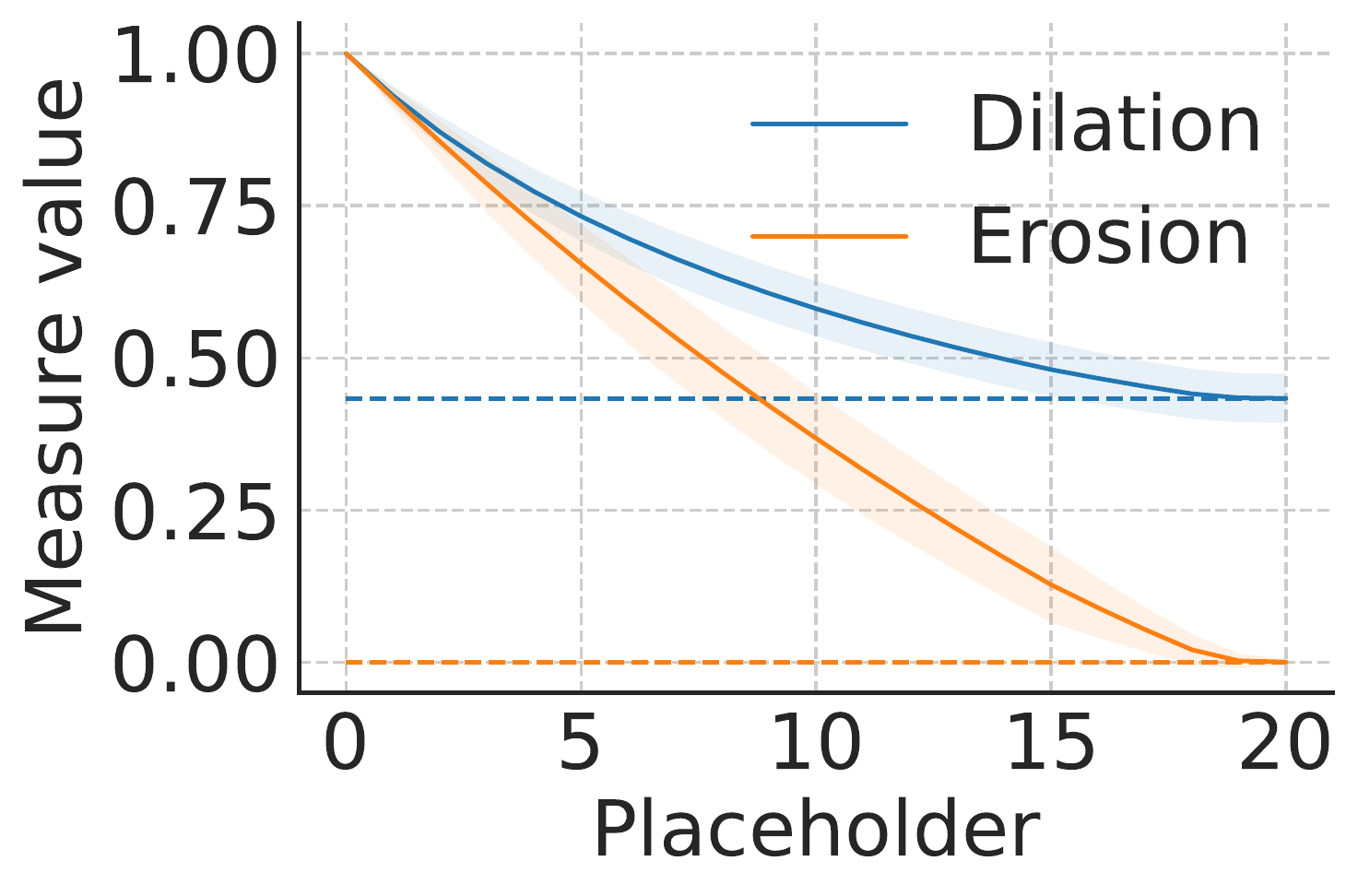} & \includegraphics[width=0.5\linewidth,trim={1.0cm 1.1cm 0cm 0cm},clip]{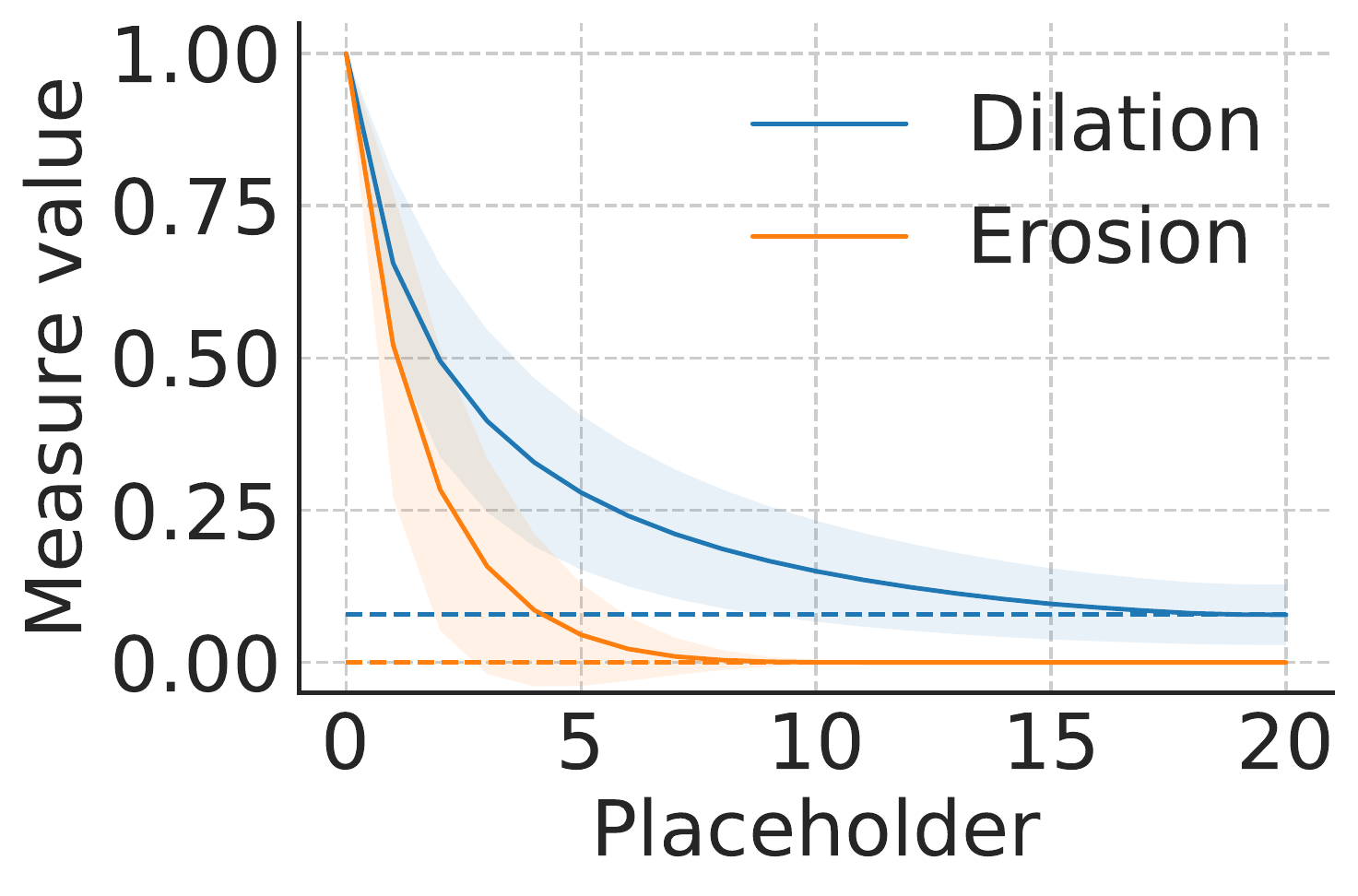} \\
    \multicolumn{2}{c}{\scriptsize Dilation/erosion (pixels)} \\
    \end{tabular} \egroup 
    \vspace{-2mm}
    \subcaption{Trimap IoU is not symmetric. It favors predictions larger than ground truth masks (\eg dilated pseudo-predictions).}
    \label{fig:trimap_iou_problem}
    \end{subfigure}
    \hfill
    \begin{subfigure}[t]{0.31\linewidth}
    \centering
    \bgroup 
    \def\arraystretch{0.5} 
    \setlength\tabcolsep{0.5pt}
    \begin{tabular}{cc}
    \scriptsize object area $> 96^2$ & \scriptsize object area $\le 32^2$  \\
    \includegraphics[width=0.5\linewidth,trim={1.0cm 1.1cm 0cm 0cm},clip]{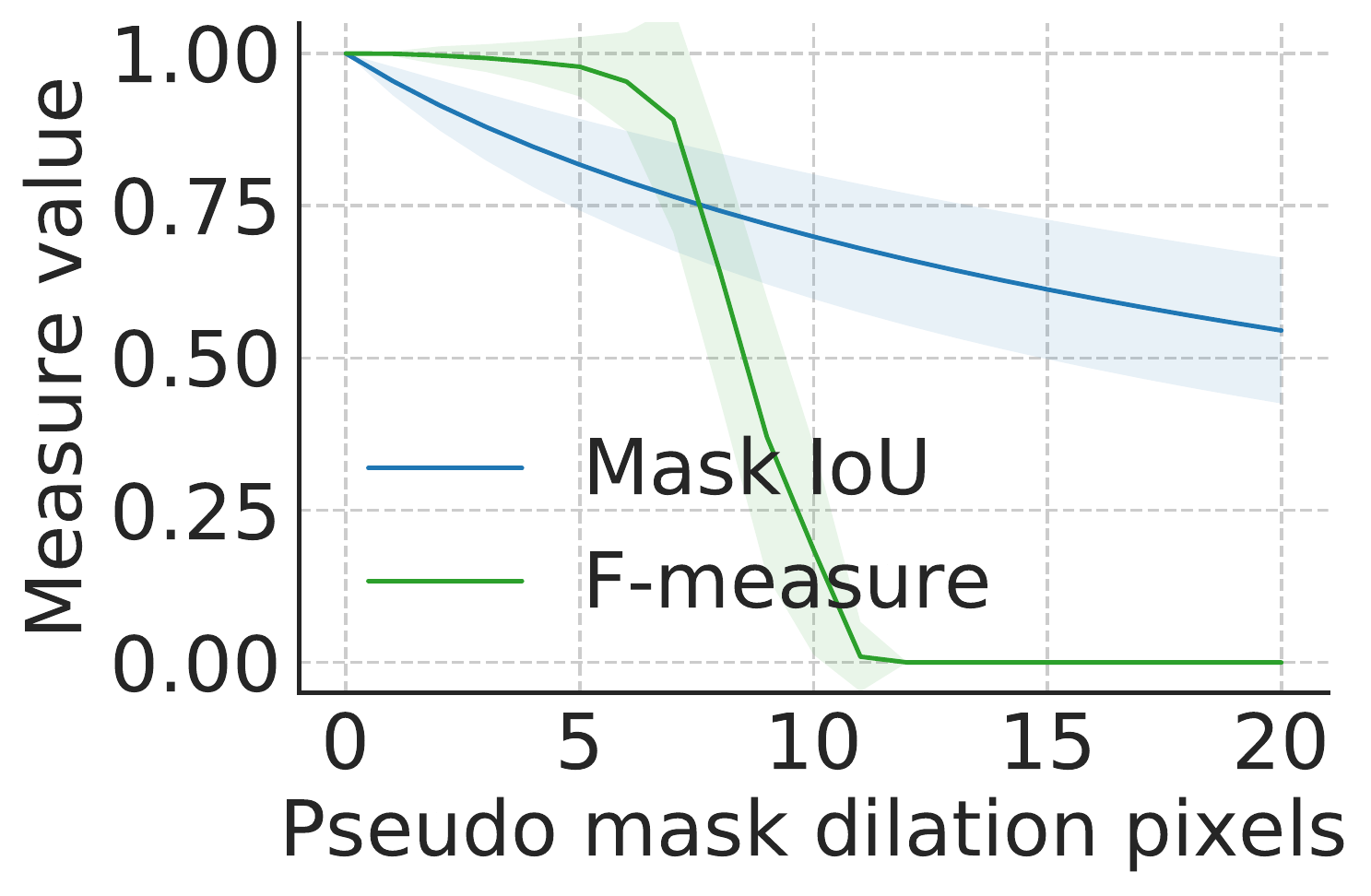} & \includegraphics[width=0.5\linewidth,trim={1.0cm 1.1cm 0cm 0cm},clip]{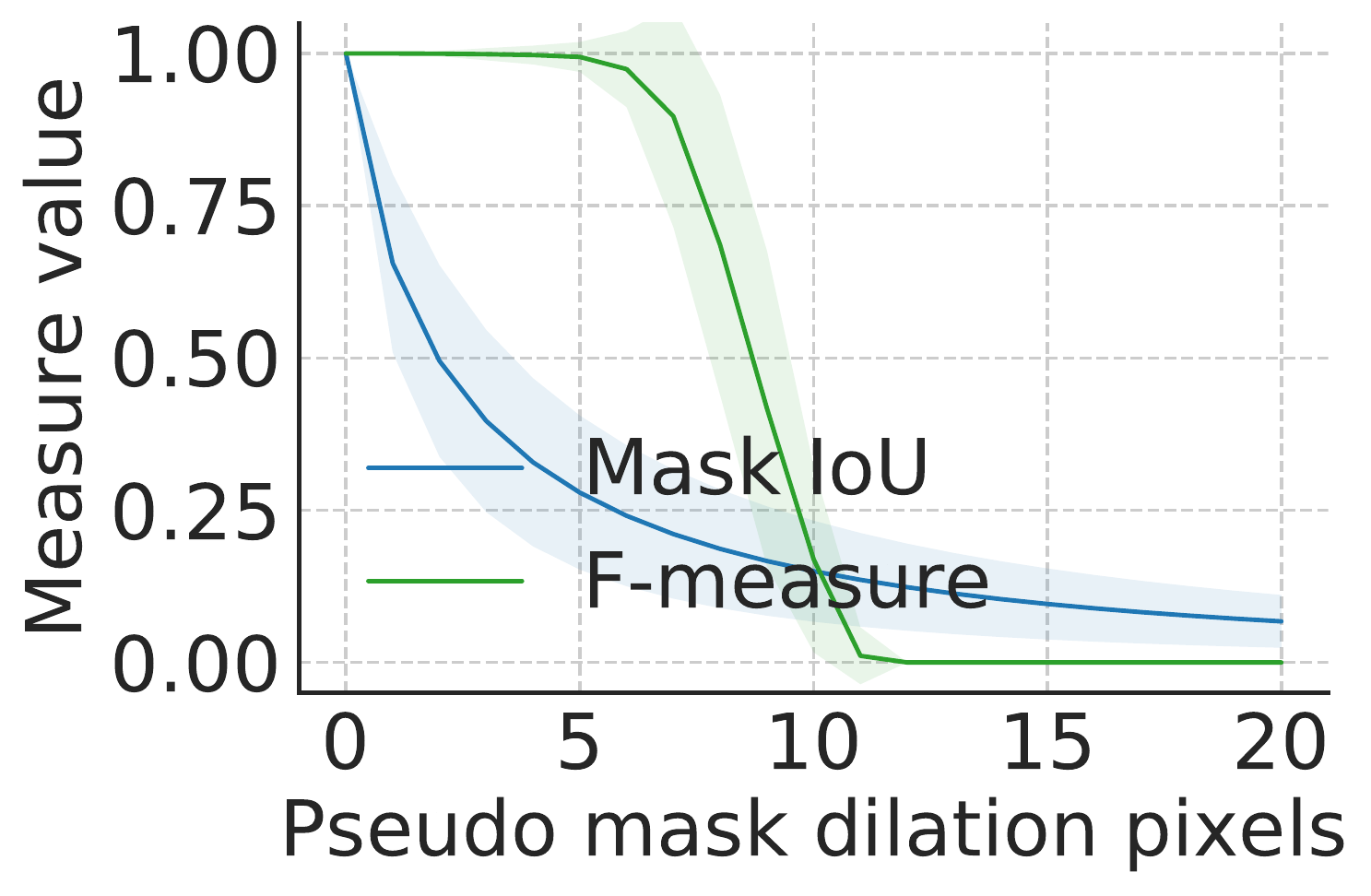} \\
    \multicolumn{2}{c}{\scriptsize Dilation (pixels)} \\
    \end{tabular} \egroup 
    \vspace{-2mm}
    \subcaption{F-measure completely ignores small contour misalignments and rapidly drops to zero within a short range of severities.}
    \label{fig:f_measure_problem}
    \end{subfigure}
    \caption{Sensitivity analysis across object scales for Mask IoU~(\ref{sub@fig:mask_iou_problem}), Trimap IoU~(\ref{sub@fig:trimap_iou_problem}), and F-measure~(\ref{sub@fig:f_measure_problem}) with scaling error type.}
    \label{fig:measures_problem}
\vspace{-2mm}
\end{figure*}

%% file: fig_tex/analysis/boundary_iou/separate_fig.tex
\begin{figure*}
    \vspace{-1mm}
    \centering

    \bgroup 
    \def\arraystretch{0.5} 
    \setlength\tabcolsep{0.5pt}
    \begin{tabular}{c cccccc}
    & \scriptsize (a) Scale (dilation) & \scriptsize (b) Scale (erosion) & \scriptsize (c) Boundary localization & \scriptsize (d) Object localization & \scriptsize (e) Boundary approximation & \scriptsize (f) Inner mask holes \\
    \multirow{3}{*}{\rotatebox[origin=c]{90}{\scriptsize Measure value \hspace{-65pt}}}  & \includegraphics[width=0.16\linewidth,trim={1.0cm 1.1cm 0cm 0cm},clip]{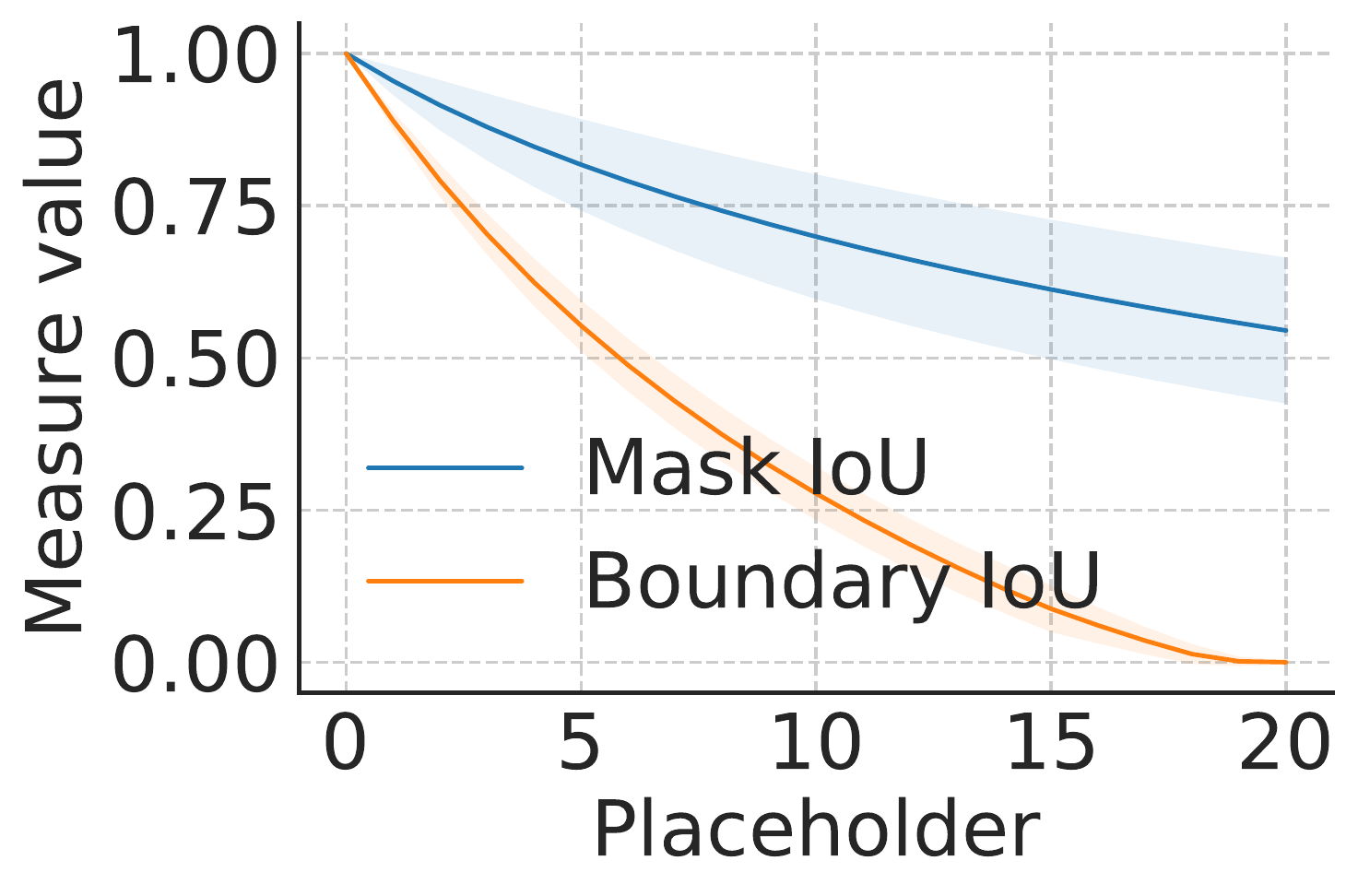} &
    \includegraphics[width=0.16\linewidth,trim={1.0cm 1.1cm 0cm 0cm},clip]{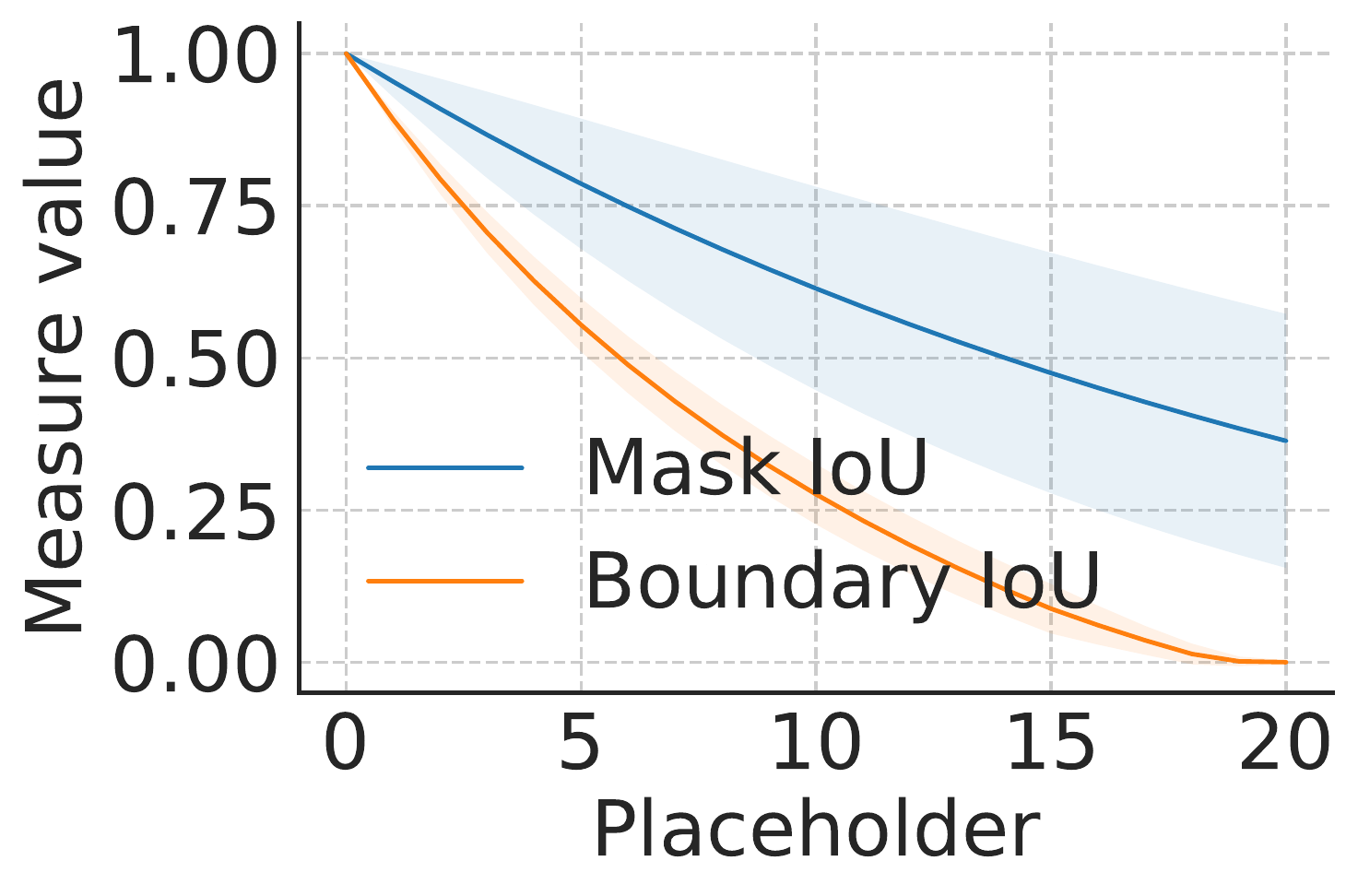} &
    \includegraphics[width=0.16\linewidth,trim={1.0cm 1.1cm 0cm 0cm},clip]{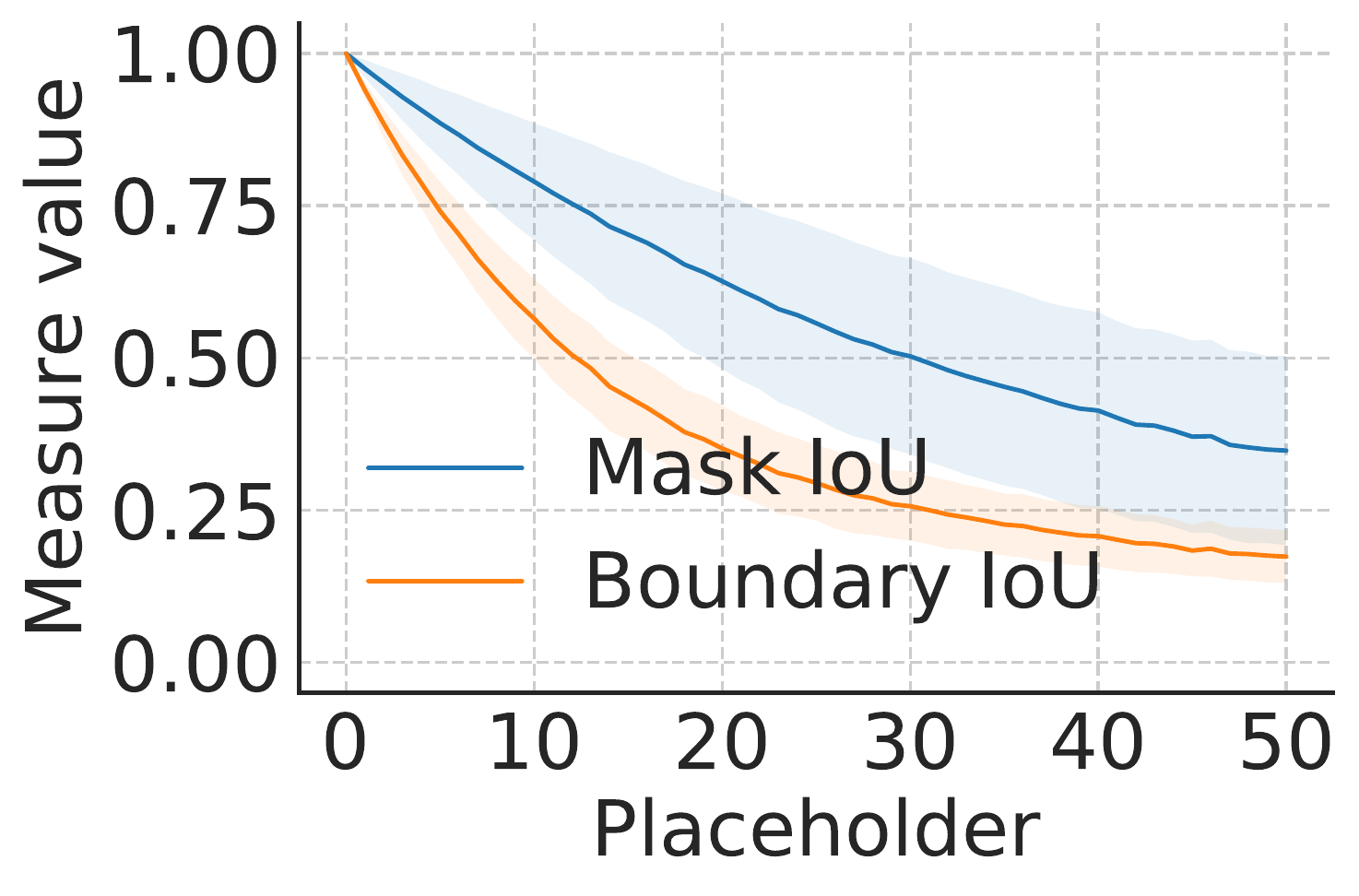} &
    \includegraphics[width=0.16\linewidth,trim={1.0cm 1.1cm 0cm 0cm},clip]{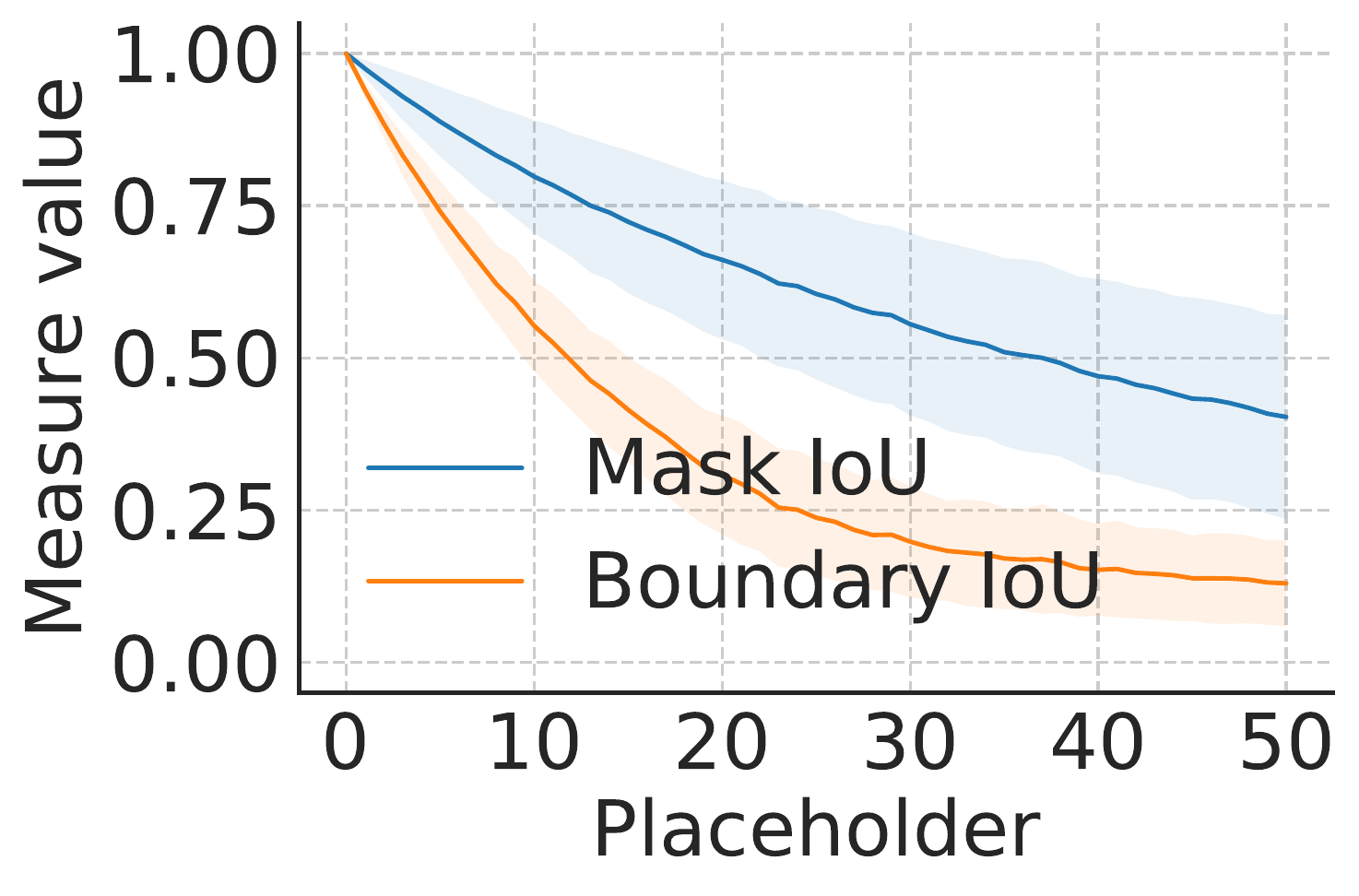} &
    \includegraphics[width=0.16\linewidth,trim={1.0cm 1.1cm 0cm 0cm},clip]{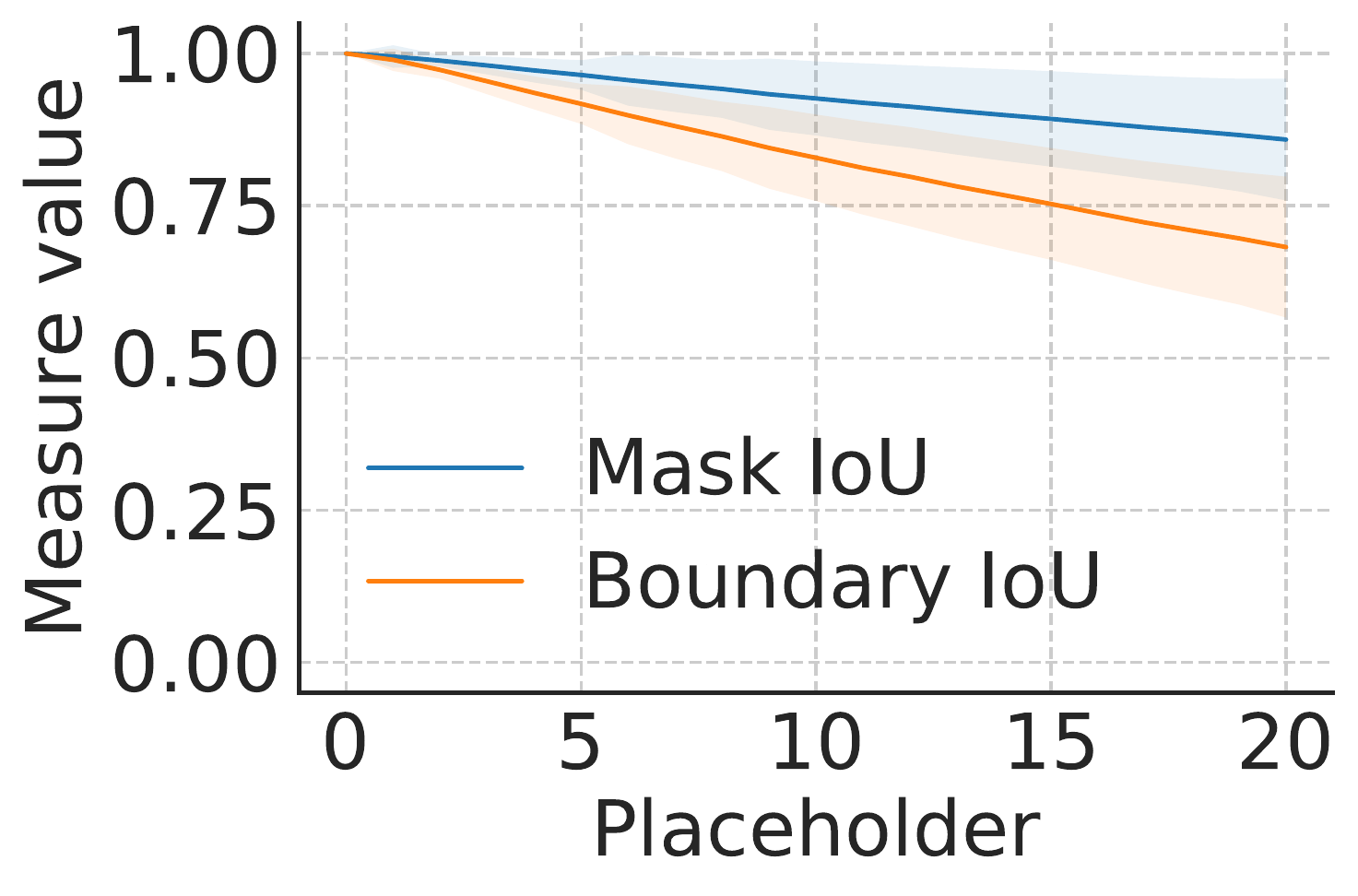} &
    \includegraphics[width=0.16\linewidth,trim={1.0cm 1.1cm 0cm 0cm},clip]{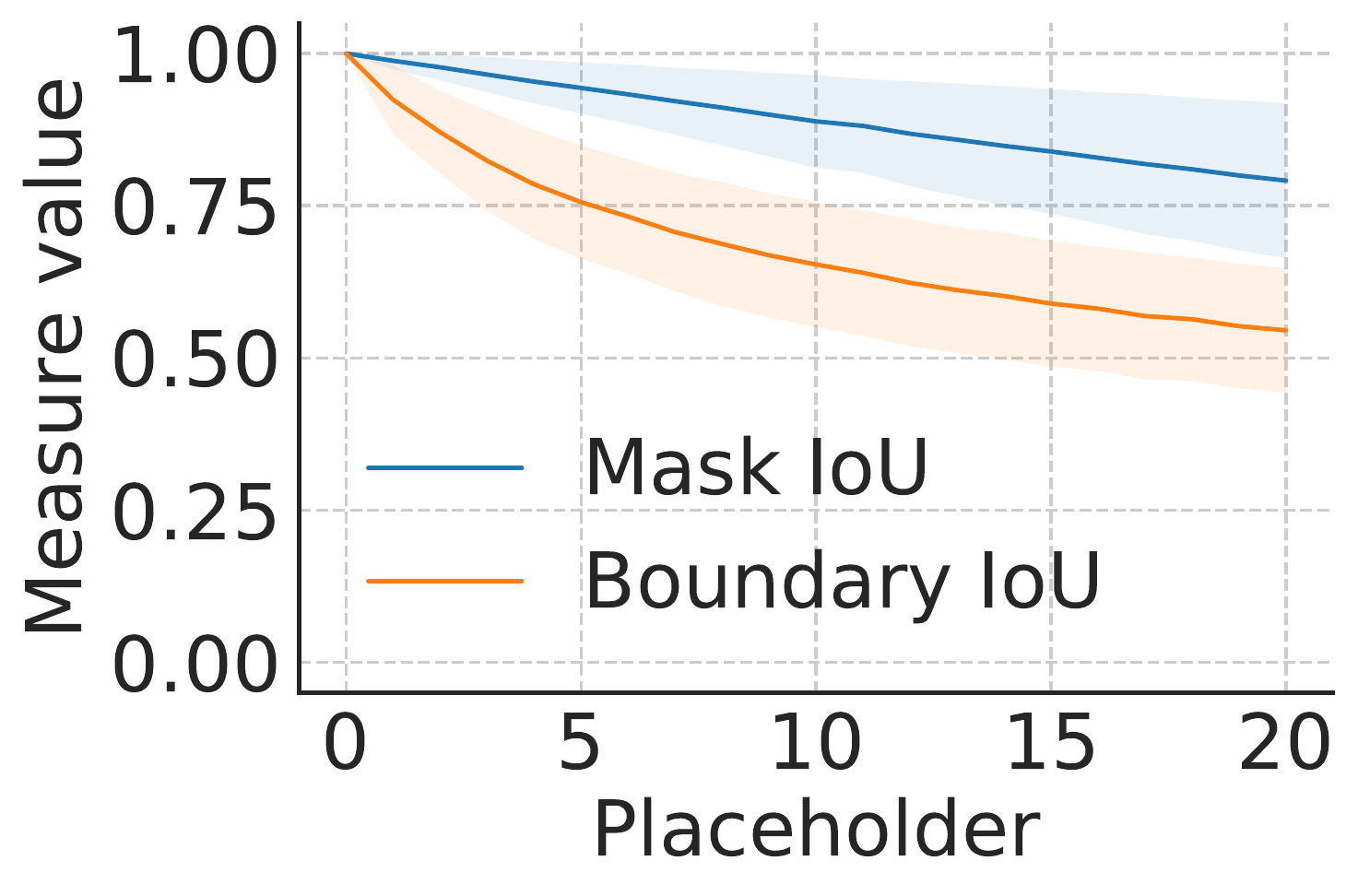} \\
    & \scriptsize Dilation (pixels) & \scriptsize Erosion (pixels) & \scriptsize Noise std (pixels) & \scriptsize Offset (pixels) & \scriptsize Error tolerance & \scriptsize Number of holes \\
    \end{tabular} \egroup 
    \vspace{-2mm}
    \caption{Boundary IoU sensitivity curves \emph{across error severities}. We use pseudo-predictions for objects with area $> 96^2$. For each error type, Boundary IoU makes better use of the 0-1 value range demonstrating an improved ability to differentiate between error severity.}
    \vspace{-1mm}
    \label{fig:biou_large_objects}
\end{figure*}

\begin{figure*}
    \centering

    \bgroup 
    \def\arraystretch{0.5} 
    \setlength\tabcolsep{0.5pt}
    \begin{tabular}{c cccccc}
    & \multicolumn{1}{p{0.16\linewidth}}{\centering \scriptsize (a) Scale \\ Dilation = 5 pixels}
    & \multicolumn{1}{p{0.16\linewidth}}{\centering \scriptsize (b) Scale \\ Erosion = 5 pixels}
    & \multicolumn{1}{p{0.16\linewidth}}{\centering \scriptsize (c) Boundary localization \\ Noise std = 10 pixels}
    & \multicolumn{1}{p{0.16\linewidth}}{\centering \scriptsize (d) Object localization \\ Offset = 10 pixels}
    & \multicolumn{1}{p{0.16\linewidth}}{\centering \scriptsize (e) Boundary approximation \\ Error tolerance = 5}
    & \multicolumn{1}{p{0.16\linewidth}}{\centering \scriptsize (f) Inner mask holes \\ Number of holes = 5} \\
    \multirow{3}{*}{\rotatebox[origin=c]{90}{\scriptsize Measure value \hspace{-65pt}}} & \includegraphics[width=0.16\linewidth,trim={1.0cm 1.1cm 0cm 0cm},clip]{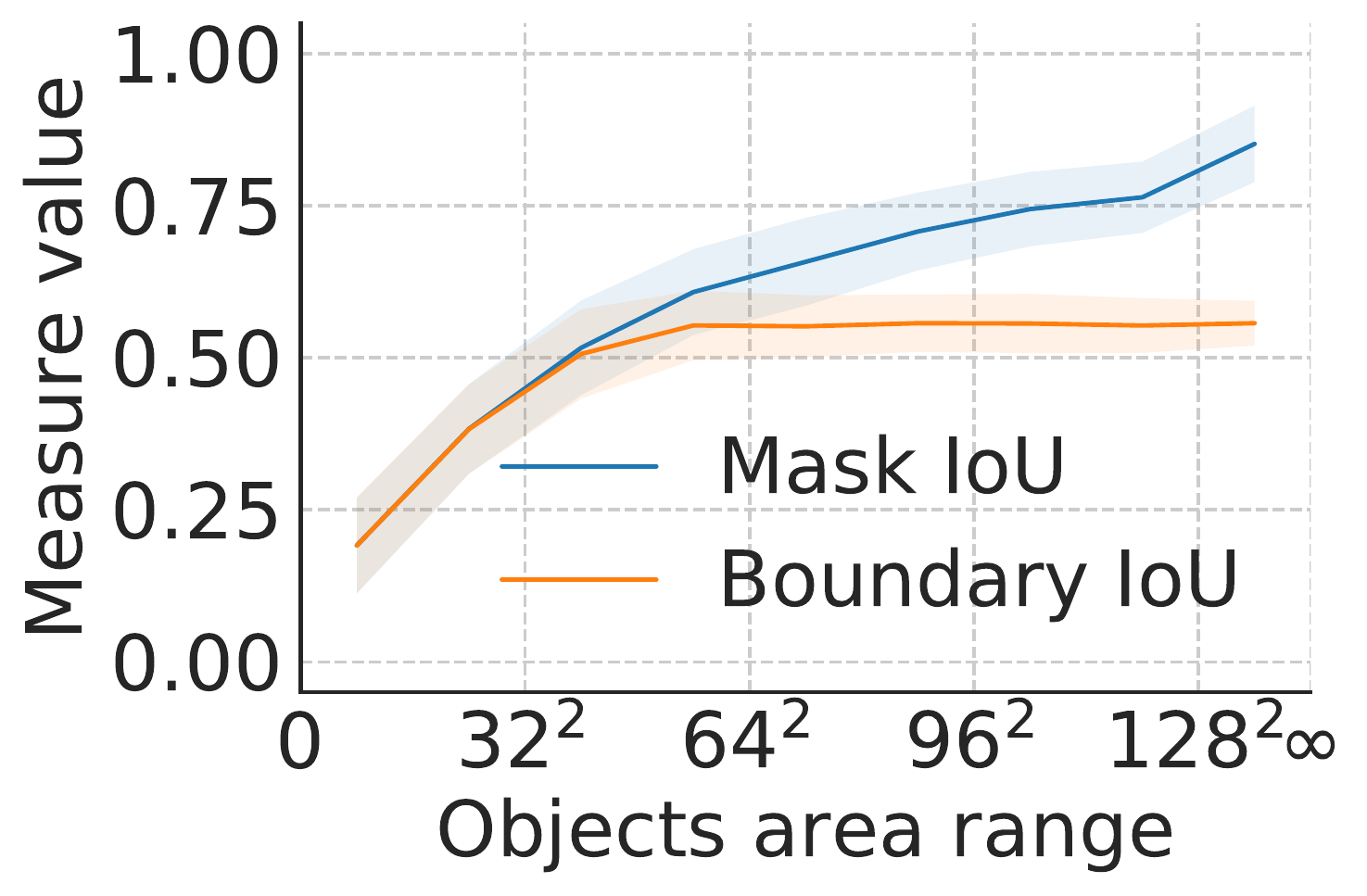} &
    \includegraphics[width=0.16\linewidth,trim={1.0cm 1.1cm 0cm 0cm},clip]{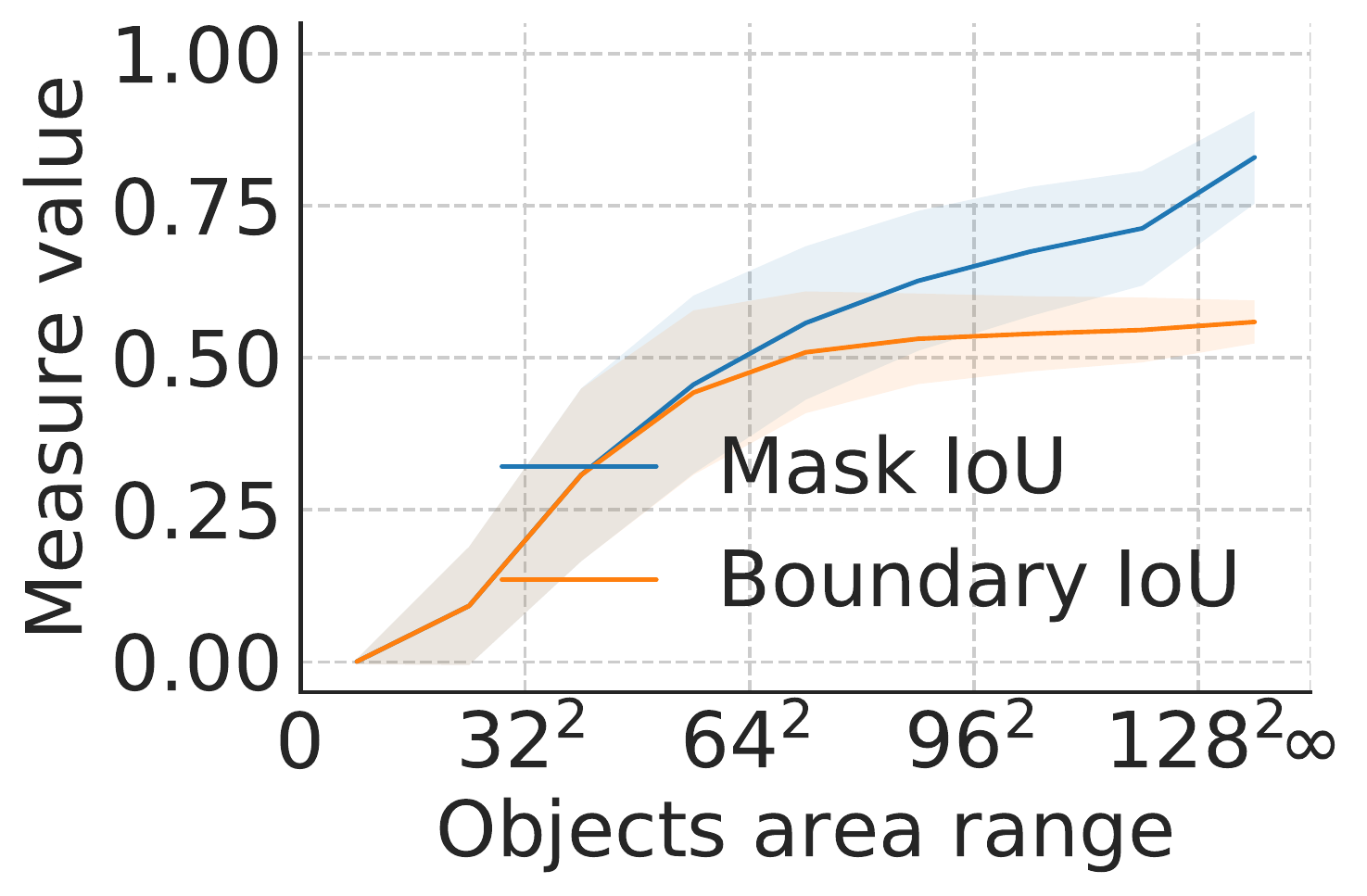} &
    \includegraphics[width=0.16\linewidth,trim={1.0cm 1.1cm 0cm 0cm},clip]{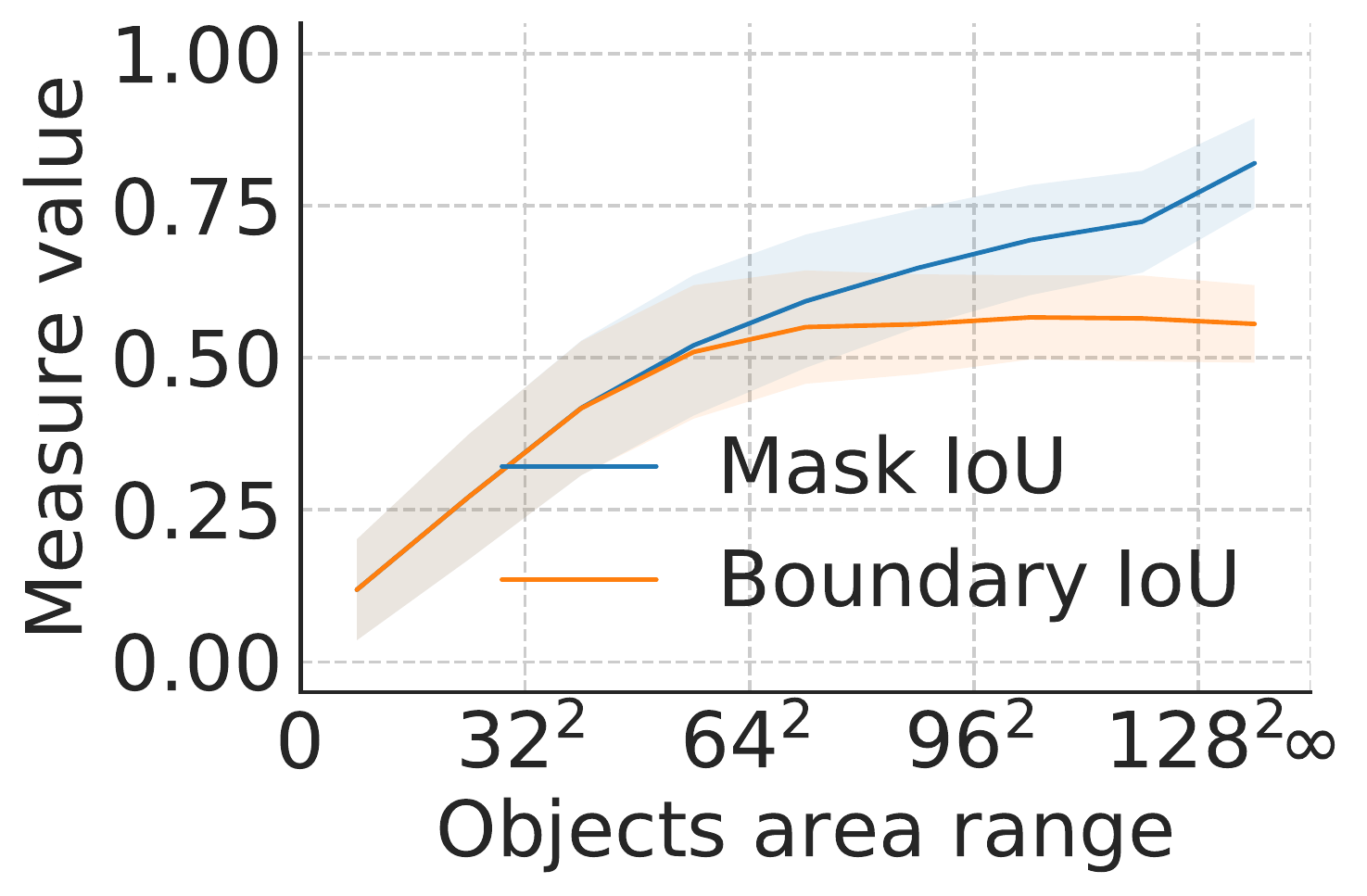} &
    \includegraphics[width=0.16\linewidth,trim={1.0cm 1.1cm 0cm 0cm},clip]{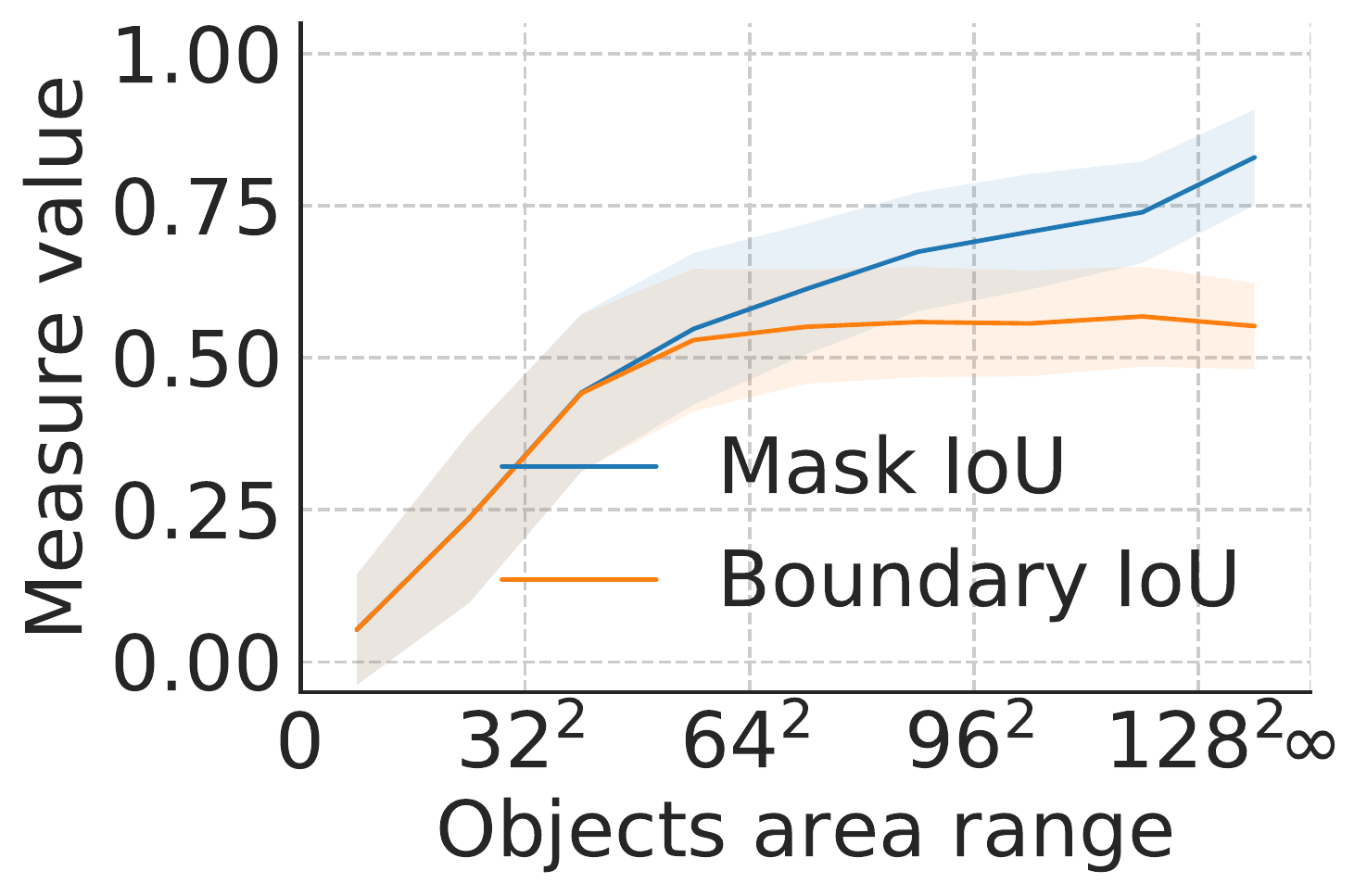} &
    \includegraphics[width=0.16\linewidth,trim={1.0cm 1.1cm 0cm 0cm},clip]{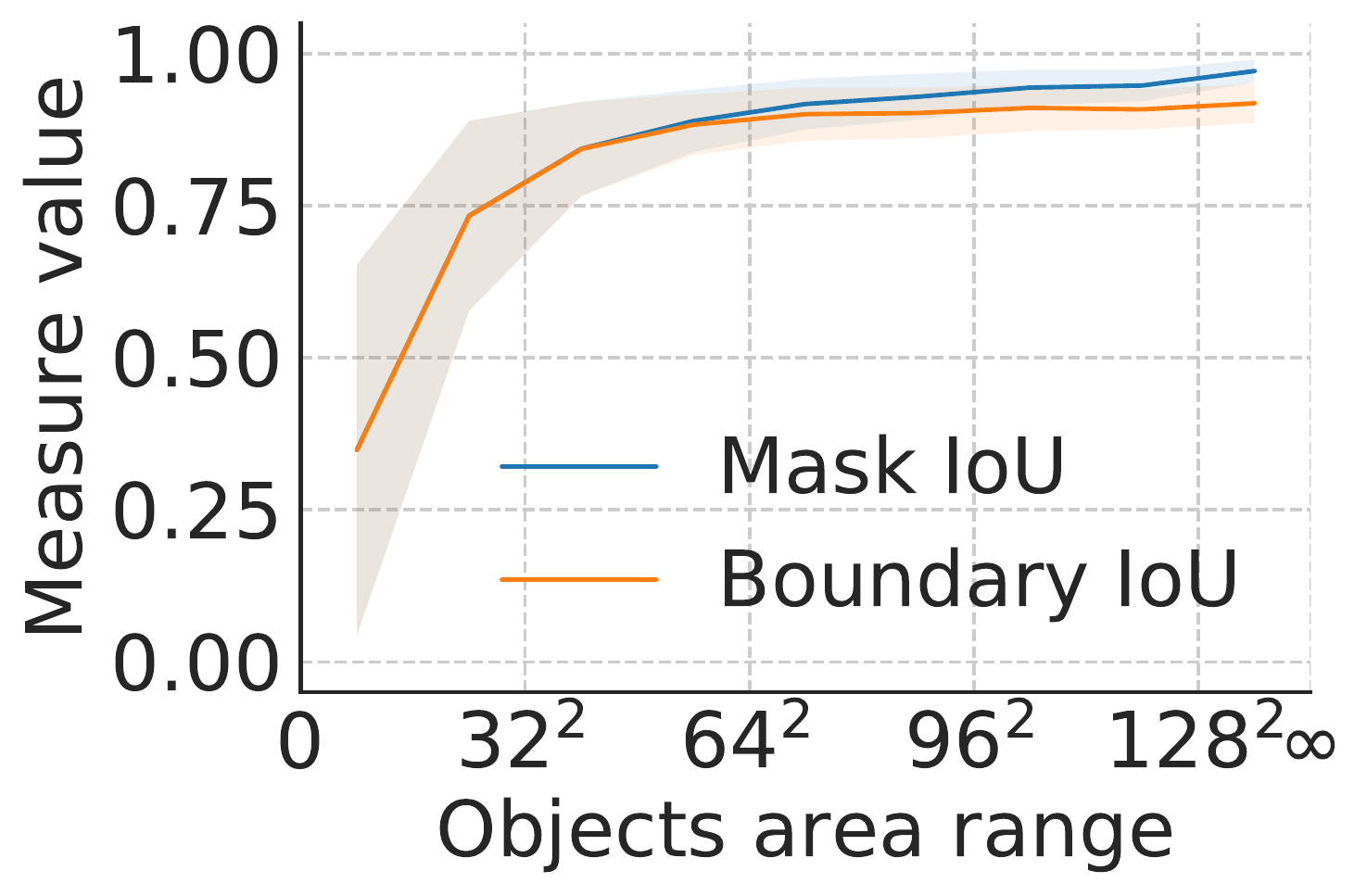} &
    \includegraphics[width=0.16\linewidth,trim={1.0cm 1.1cm 0cm 0cm},clip]{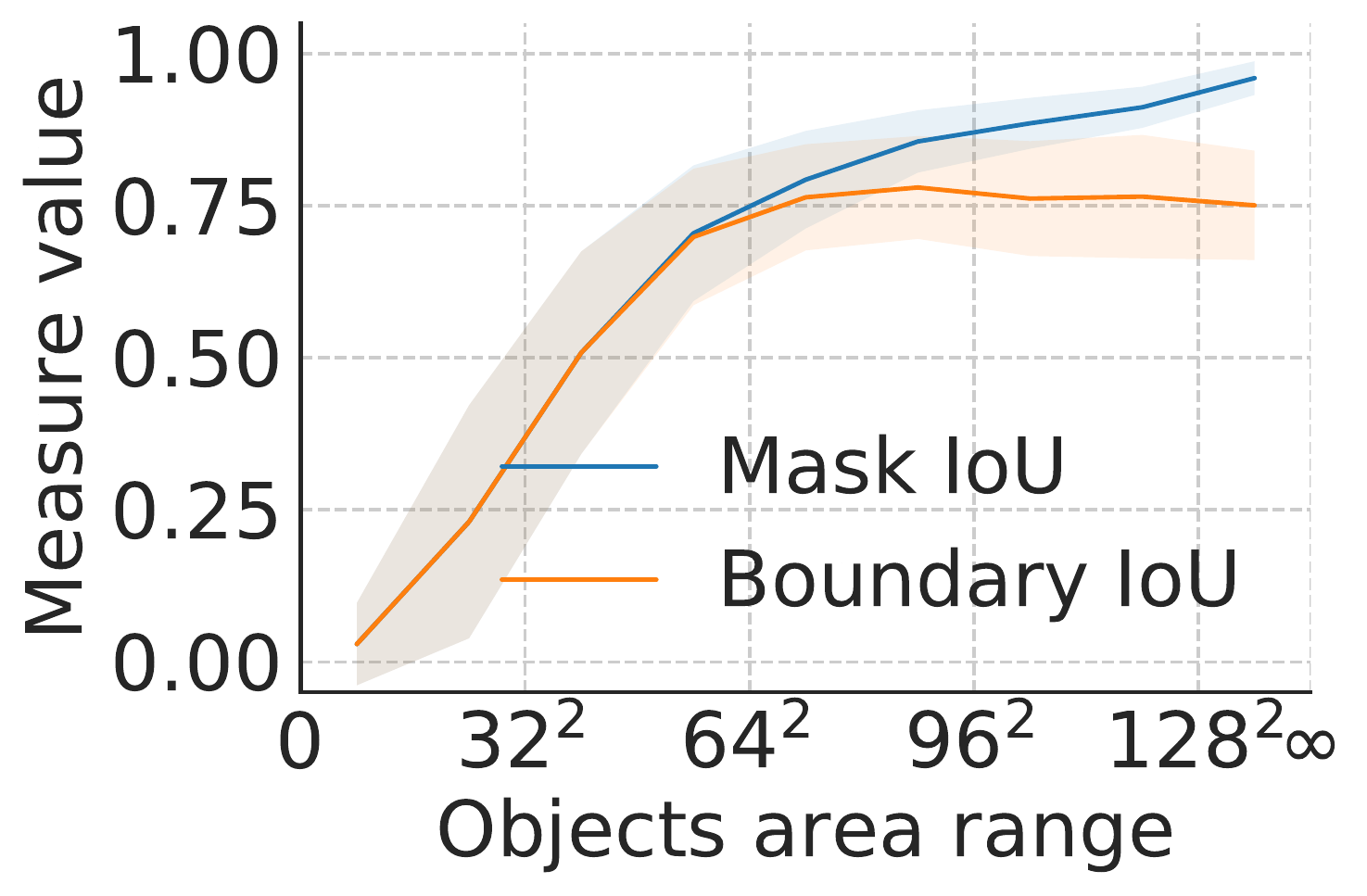} \\
    & \multicolumn{6}{c}{\scriptsize Binned object sizes}
    \end{tabular} \egroup 
    \vspace{-1mm}
    \caption{Boundary IoU sensitivity curves \emph{across object sizes}. We use pseudo-predictions for objects of different sizes with \emph{fixed} error severities. Objects are binned by their area with $16^2$ increment. For larger objects, Boundary IoU remains flat given the fixed error severity, while Mask IoU demonstrates a clear preferential bias for large objects. For small objects, both metrics have similar curves indicating that neither over-penalizes small objects \wrt to the other.}
    \label{fig:biou_across_scales}
\end{figure*}

%% file: tables/ins_seg_exp1.tex
\begin{table}[t]
  \tablestyle{6pt}{1.1}
  \begin{tabular}{c|x{22}x{22}x{22}x{22}}
    Evaluation metric & AP & AP$_S$ & AP$_M$ & AP$_L$ \\
    \shline
    Mask AP & 96.5 & 98.9 & 95.7 & 95.0 \\
    Boundary AP & 85.9 & 98.9 & 93.0 & 73.0
  \end{tabular}
  \vspace{-2mm}
  \caption{Boundary AP and Mask AP on COCO \texttt{val} set for synthetic $28 \x 28$ predictions generated from the ground truth. Unlike Mask AP$_L$, Boundary AP$_L$ successfully captures the lack of fidelity in the synthetic prediction for large objects (area $> 96^2$).}
  \vspace{-2mm}
\label{tab:coco_gt_mask_resolution}
\end{table}

%% file: tables/ins_seg_exp2.tex
\begin{table*}[!t]
  \begin{subtable}[t]{0.47\textwidth}
  \tablestyle{2pt}{1.1}
  \begin{tabular}{c c|x{22}x{22}x{22}x{22}}
  \multicolumn{6}{c}{~}\\
    Method & Evaluation metric & AP & AP$_S$ & AP$_M$ & AP$_L$ \\
    \shline
    Mask R-CNN & Mask AP &     52.5 & 44.9 & 55.0 & 66.0 \\
    Mask R-CNN & Boundary AP & 36.1 & 44.8 & 46.5 & 25.9 \\
    \multicolumn{6}{c}{~}\\
  \end{tabular}
  \caption{AP of Mask R-CNN \textbf{with ground truth boxes}. Mask R-CNN makes blobby predictions with large defects around boundaries for large objects (see Figure~\ref{fig:teaser}). Boundary AP successfully captures these errors with much lower AP\textsubscript{$L$} than Mask AP.}
  \label{tab:inference_with_gt_box_a}
  \end{subtable}\hspace{3mm}
  \begin{subtable}[t]{0.495\textwidth}
  \tablestyle{2pt}{1.1}
  \begin{tabular}{c c|x{34}x{34}|x{34}x{34}}
    &  & \multicolumn{2}{c|}{Ground truth boxes} & \multicolumn{2}{c}{Predicted boxes} \\
    Method & Backbone & AP\scriptsize{$^{\text{mask}}$} & AP\scriptsize{$^{\text{boundary}}$} & AP\scriptsize{$^{\text{mask}}$} & AP\scriptsize{$^{\text{boundary}}$} \\
    \shline
    Mask R-CNN & R50 &  53.6 \phantom{\dt{+0.1}} & 37.7 \phantom{\dt{+0.1}} & 37.2 \phantom{\dt{+0.1}} & 23.1 \phantom{\dt{+0.1}} \\
    Mask R-CNN & R101 & 53.8 \bd{\dt{+0.2}}      & 38.2 \bd{\dt{+0.5}}      & 38.6 \dt{+1.4}           & 24.5 \dt{+1.4} \\
    Mask R-CNN & X101 & 53.4 \dt{-0.2}           & 38.1 \dt{+0.4}           & 39.5 \bd{\dt{+2.3}}      & 25.4 \bd{\dt{+2.3}} \\
  \end{tabular}
  \caption{Larger backbones do not improve segmentation quality significantly (\ie, Boundary AP stays roughly the same when ground truth boxes are used). With real box predictions, Boundary AP tracks the categorization and localization improvements similarly to Mask AP.}
  \label{tab:inference_with_gt_box_b}
  \end{subtable}\vspace{2mm}

  \begin{subtable}[t]{0.4825\textwidth}
  \tablestyle{2pt}{1.1}
  \begin{tabular}{c | x{34}x{34}x{34}x{34}x{34}}
    Method & AP\scriptsize{$^{\text{mask}}$} & AP\scriptsize{$^{\text{boundary}}$} & AP\scriptsize{$^{\text{boundary}}_S$} & AP\scriptsize{$^{\text{boundary}}_M$} & AP\scriptsize{$^{\text{boundary}}_L$} \\
    \shline
    Mask R-CNN    & 52.5 \phantom{\dt{+0.1}} & 36.1 \phantom{\dt{+0.1}} & 44.8 \phantom{\dt{+0.1}} & 46.5 \phantom{\dt{+0.1}} & 25.9 \phantom{\dt{+0.1}} \\
    PointRend-28  & 57.0 \dt{+4.5}           & 41.6 \dt{+5.5}           & 48.2 \dt{+3.4}           & 52.3 \dt{+5.8}           & 33.3 \dt{+7.4} \\
    PointRend-224 & 57.2 \bd{\dt{+4.7}}      & 42.1 \bd{\dt{+6.0}}      & 48.3 \bd{\dt{+3.5}}      & 52.5 \bd{\dt{+6.0}}      & 34.4 \bd{\dt{+8.5}} \\
  \end{tabular}
  \caption{PointRend (with either $28 \x 28$ or $224 \x 224$ output resolution) is designed to give higher-quality output masks than Mask R-CNN (which has $28 \x 28$ output resolution). Boundary AP captures these improvements well, especially for the higher-resolution output variant of PointRend and for AP\textsubscript{$L$}.}  
  \label{tab:inference_with_gt_box_c}
  \end{subtable}\hspace{3mm}
  \begin{subtable}[t]{0.4825\textwidth}
  \tablestyle{2pt}{1.1}
  \begin{tabular}{c | x{34}x{34}x{34}x{34}x{34}}
    Method & AP\scriptsize{$^{\text{mask}}$} & AP\scriptsize{$^{\text{boundary}}$} & AP\scriptsize{$^{\text{boundary}}_S$} & AP\scriptsize{$^{\text{boundary}}_M$} & AP\scriptsize{$^{\text{boundary}}_L$} \\
    \shline
    Mask R-CNN &  52.5 \phantom{\dt{+0.1}} & 36.1 \phantom{\dt{+0.1}} & 44.8 \phantom{\dt{+0.1}} & 46.5 \phantom{\dt{+0.1}} & 25.9 \phantom{\dt{+0.1}} \\
    PointRend &   57.2 \dt{+4.7}           & 42.1 \dt{+6.0}           & 48.3 \dt{+3.5}           & 52.5 \dt{+6.0}           & 34.4 \bd{\dt{+8.5}} \\
    BMask R-CNN & 57.4 \bd{\dt{+4.9}}      & 42.3 \bd{\dt{+6.2}}      & 48.7 \bd{\dt{+4.1}}      & 52.7 \bd{\dt{+6.2}}      & 33.9 \dt{+8.0} \\
  \end{tabular}
  \caption{Boundary-preserving Mask R-CNN (BMask R-CNN) which uses $28 \x 28$ output \vs PointRend which uses $224 \x 224$ output. The Boundary AP metric reveals that BMask R-CNN outperforms PointRend for small objects but trails it for large objects where the high output resolution of PointRend improves boundary quality.}
  \label{tab:inference_with_gt_box_d}
  \end{subtable}
\vspace{-2mm}
\caption{Comparison of Mask AP and Boundary AP on COCO \texttt{val} set for different instance segmentation models \textbf{fed with ground truth boxes unless specified otherwise}. Using ground truth boxes disentangles segmentation errors from localization and classification errors.}
\label{tab:inference_with_gt_box}
\end{table*}

%% file: appendix.tex
\section{Additional Measure Analysis}\label{sec:sensitivity_appendix}

\paragraph{Trimap IoU} computes IoU for a band around the ground truth boundary and, therefore, it ignores errors away from the ground truth boundary (\eg inner mask prediction errors). We generate pseudo-predictions with such errors by adding holes of random shapes to ground truth masks. In Figure~\ref{fig:trimap_iou_hole} we show that Trimap IoU penalizes inner mask prediction errors less than Mask IoU.

\input{fig_tex/analysis/trimap_iou/hole}

\paragraph{F-measure} matches the pixels of the predicted and ground truth contours if they are within the pixels distance threshold $d$. In the experiments presented in the main text we observe that this strategy makes F-measure ignore scale type errors for smaller objects. The Mean F-measure (mF-measure) modification ameliorates this limitation by averaging several F-measures with different threshold parameters $d$. Figure~\ref{fig:mf_measure_dilation_erosion} demonstrates the sensitivity curves of this measure for the scale (dilation) error type. For mF-measure we use $d$ from $0.1\%$ to $2.1\%$ image diagonal with $0.4\%$ increment (from 1 pixel to 17 pixels on average) to compare it with Boundary IoU that uses single $d$ set to $2\%$ image diagonal. We observe that mF-measure behaves similarly to Boundary IoU for large objects, however it under-penalizes errors in small objects where Boundary IoU matches Mask IoU behavior. Furthermore, mF-measure is substantially slower than Boundary IoU as it requires to perform the matching of prediction/ground truth pairs several times for different thresholds $d$.

\input{fig_tex/analysis/mf_measure/dilation_erosion}

\begin{figure}[h!]
    \centering
    \bgroup 
    \def\arraystretch{0.2} 
    \setlength\tabcolsep{0.2pt}
    \begin{tabular}{c}
    \includegraphics[width=0.9\linewidth]{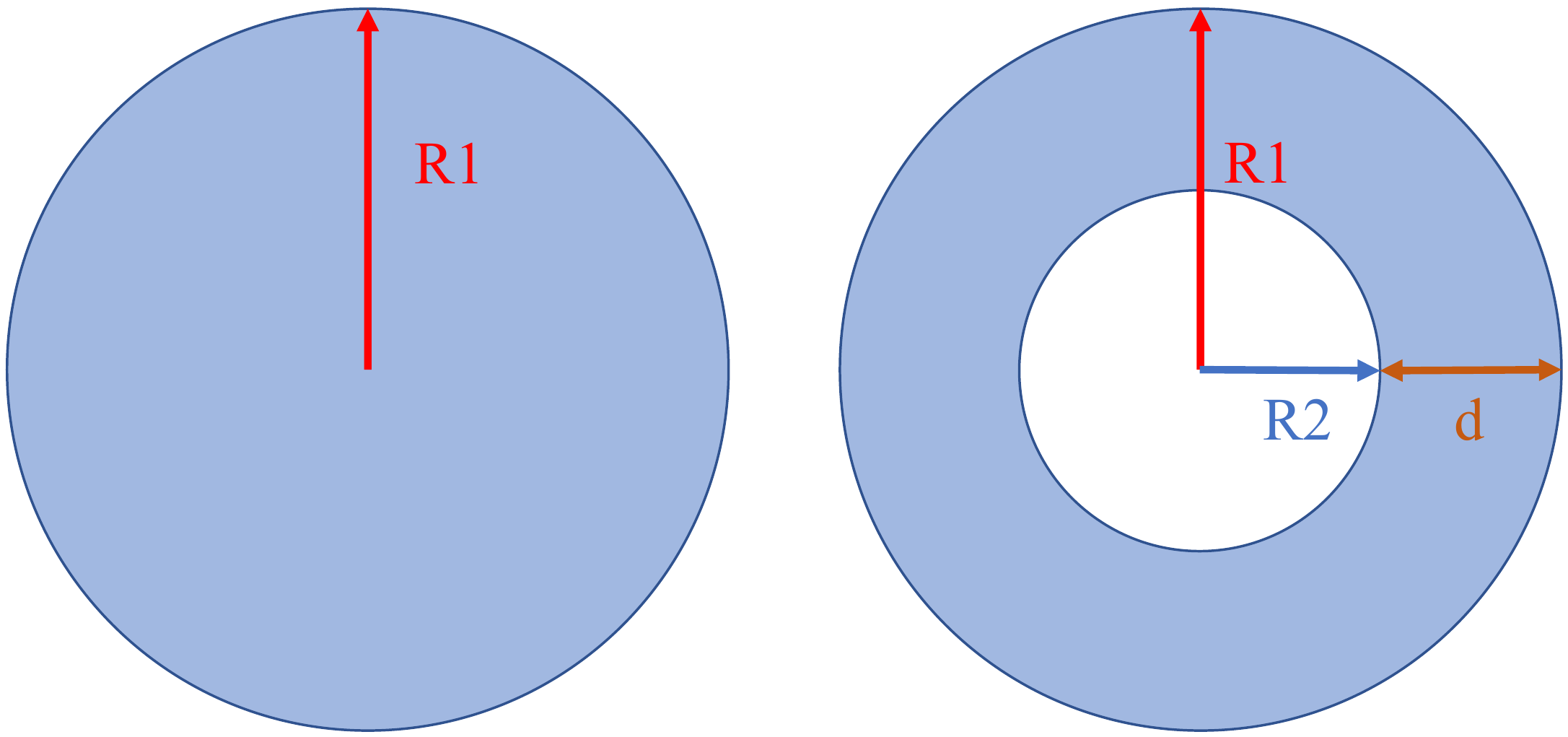} \\
    \end{tabular} \egroup
     
    \caption{Boundary IoU gives a perfect score for two non-identical masks: a disc mask and a ring mask that has the same center and outer radius as the disk, plus the inner radius that is exactly $d$ pixels smaller than the outer one.}
    \label{fig:boundary_iou_limitation}
\end{figure}

\paragraph{Boundary IoU} can award a perfect score for two non-identical masks~(see Figure~\ref{fig:boundary_iou_limitation}). As discussed in the main text, we observe that Boundary IoU is smaller or equal to Mask IoU in the absolute majority of cases and the inequality is violated when prediction misses interior part of an object~(similar to the toy example in Figure~\ref{fig:boundary_iou_limitation}). To mitigate this limitation, we propose a simple combination of Mask IoU and Boundary IoU by taking their minimum for real-world segmentation evaluation metrics.

\input{tables/coco_gt_mask_resolution}

\section{Application}\label{sec:applications}

\subsection{Instance Segmentation}\label{sec:instance}

\paragraph{Datasets.} We evaluate instance segmentation on three datasets: COCO~\cite{lin2014coco}, LVIS~\cite{gupta2019lvis} and Cityscapes~\cite{Cordts2016Cityscapes}.

\noindent \textit{COCO}~\cite{lin2014coco} is the most popular instance segmentation benchmark for common objects. It contains 80 categories. There are 118k images for training, 5k images for validation and 20k images for testing.

\noindent \textit{LVIS}~\cite{gupta2019lvis} is a federated dataset with more than 1000 categories. It shares the same set of images as COCO but the dataset has higher quality ground truth masks. We use LVISv0.5 version of the dataset. Following~\cite{kirillov2020pointrend}, we  construct the LVIS$^*$v0.5 dataset which keeps only the 80 COCO categories from LVISv0.5. LVIS$^*$v0.5 allows us to compare models trained on COCO using higher quality mask annotations from LVIS (\ie AP$^*$ in~\cite{kirillov2020pointrend}).

\noindent \textit{Cityscapes}~\cite{Cordts2016Cityscapes} is a street-scene high-resolution dataset. There are 5k images annotated with high quality pixel-level annotations and 8 classes with instance-level segmentation.

\begin{figure}[h!]
    \centering
    \bgroup 
    \def\arraystretch{0.2} 
    \setlength\tabcolsep{1pt}
    \begin{tabular}{cccc}
    \includegraphics[width=0.24\linewidth]{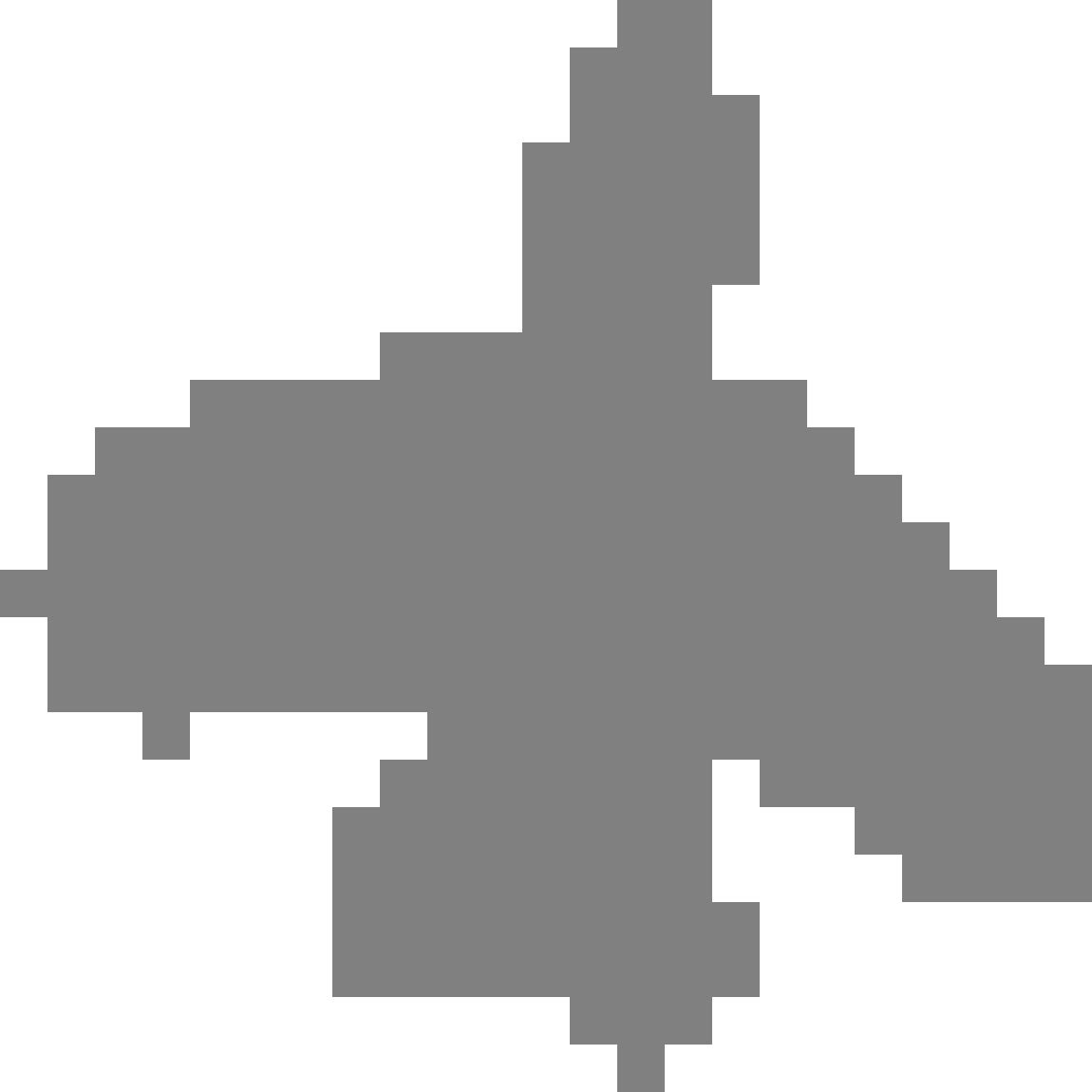} &
    \includegraphics[width=0.24\linewidth]{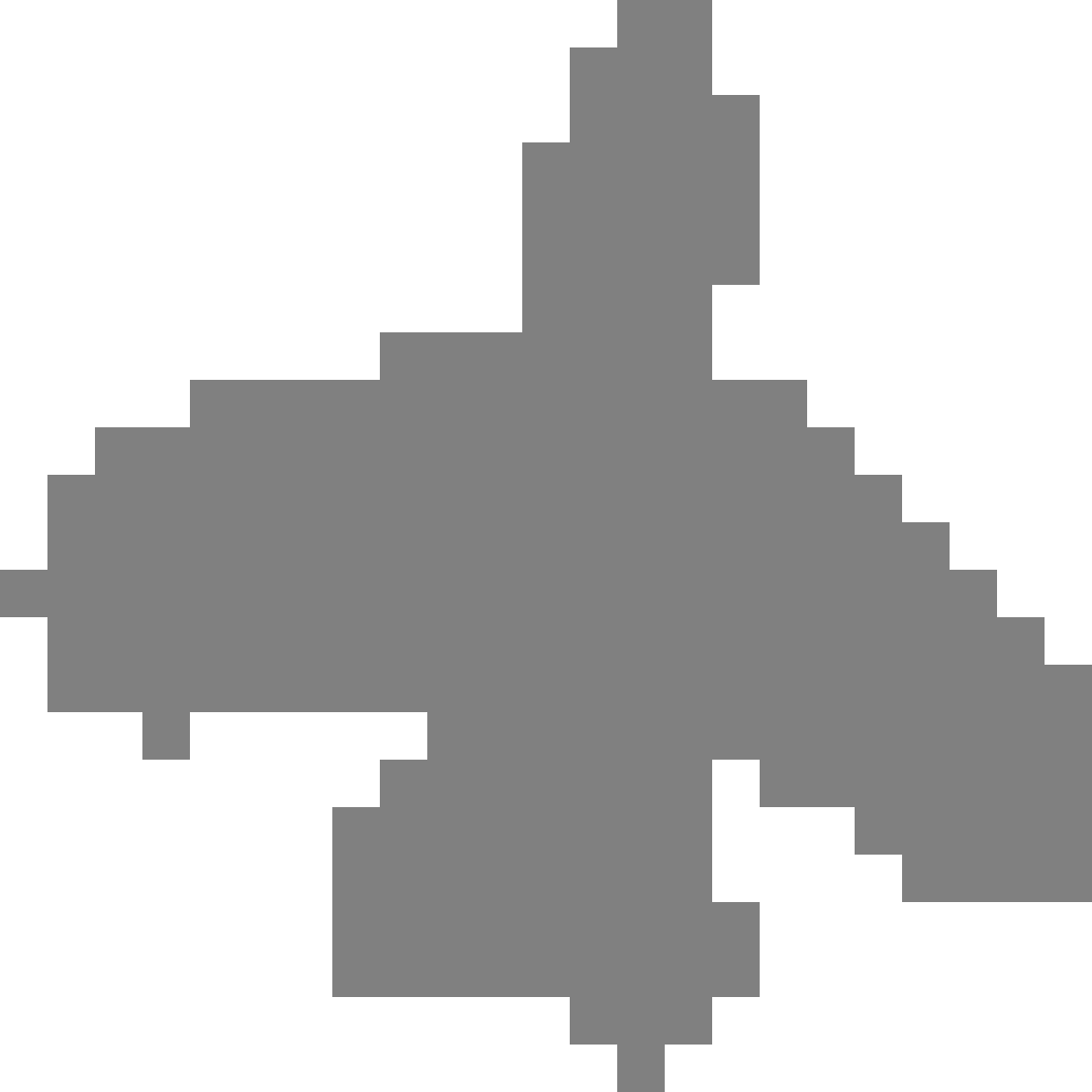} &
    \includegraphics[width=0.24\linewidth]{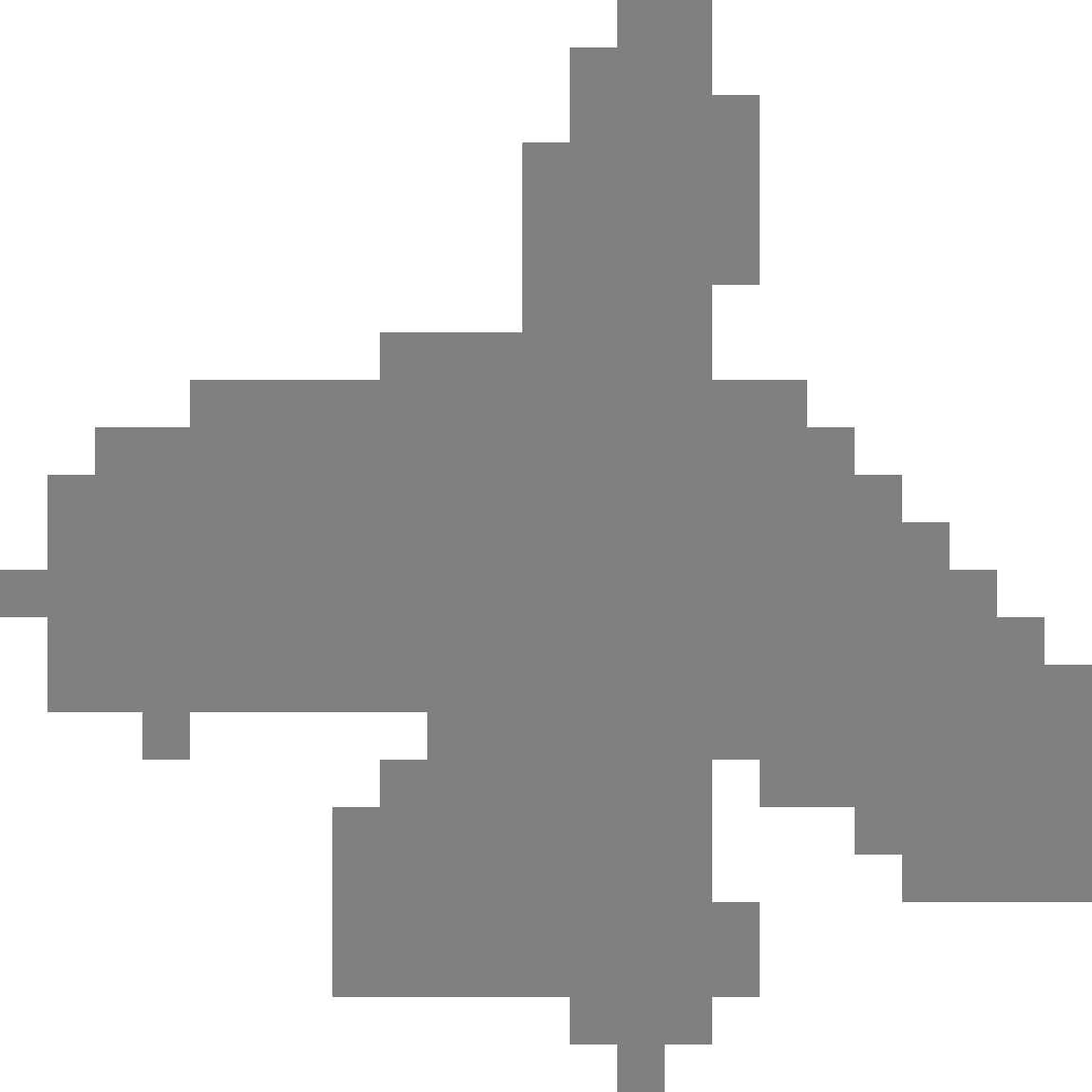} &
    \includegraphics[width=0.24\linewidth]{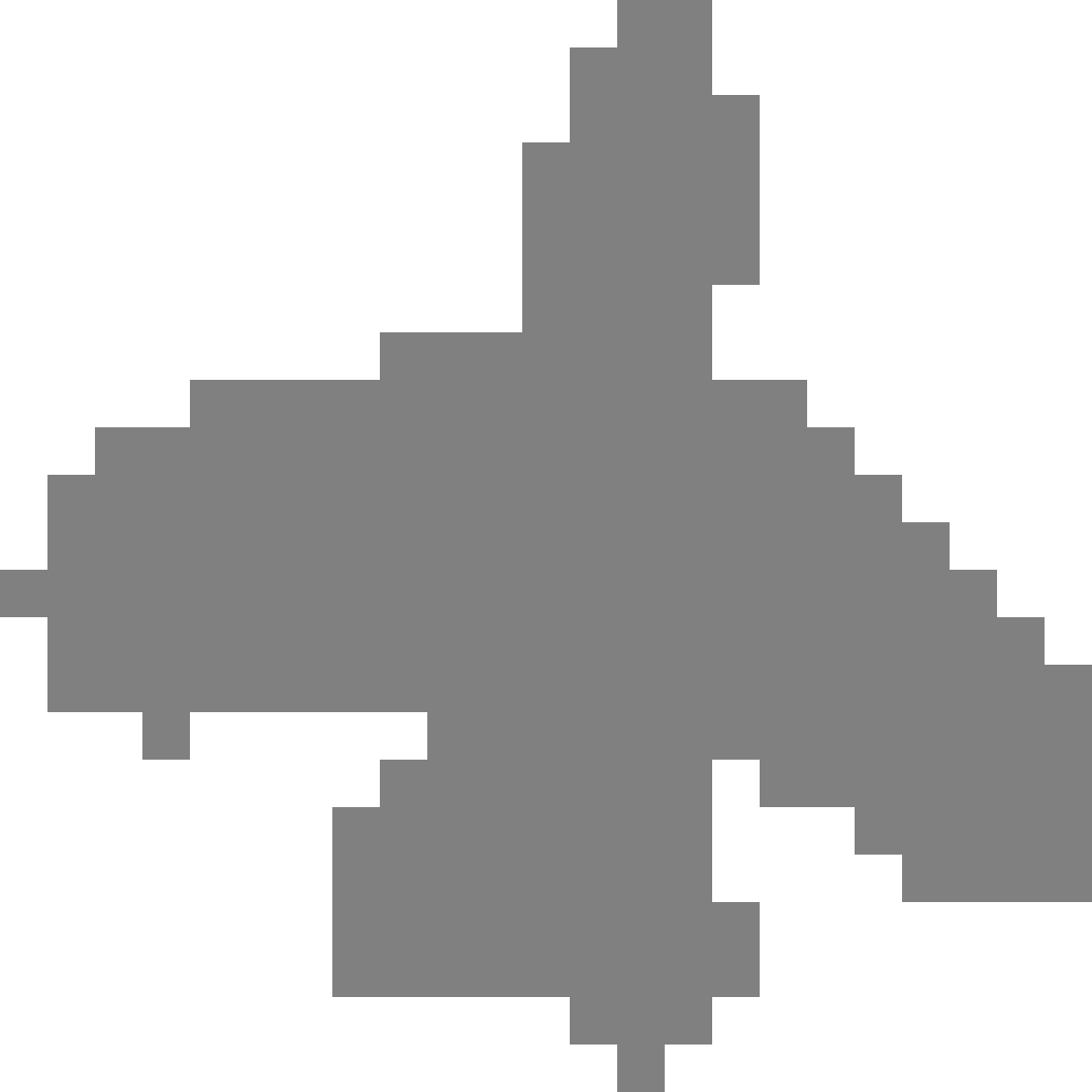} \\
    \includegraphics[width=0.24\linewidth]{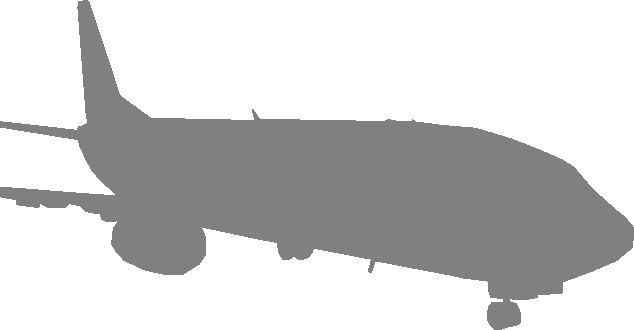} &
    \includegraphics[width=0.24\linewidth]{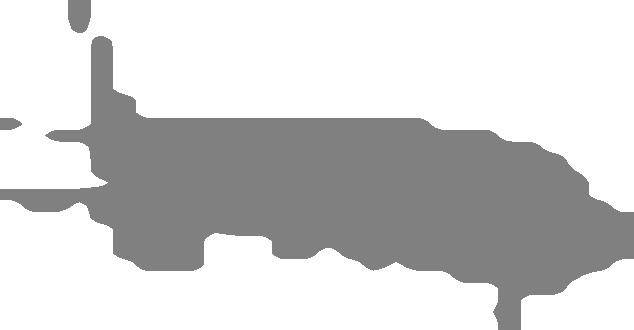} &
    \includegraphics[width=0.24\linewidth]{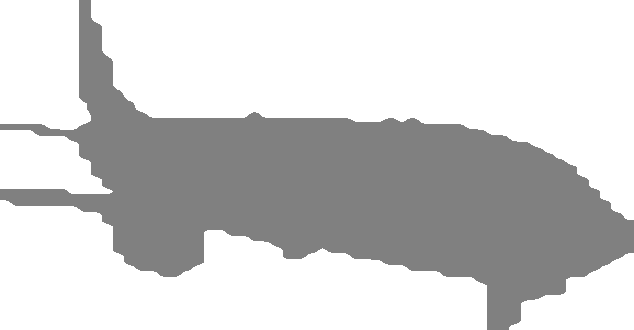} &
    \includegraphics[width=0.24\linewidth]{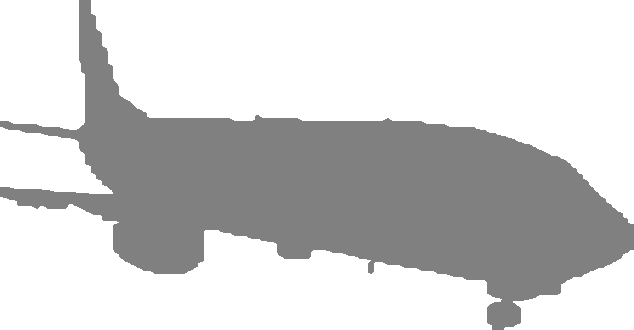} \\
    \scriptsize Ground Truth & \scriptsize $28 \x 28$ & \scriptsize $56 \x 56$ & \scriptsize $112 \x 112$ \\
    \end{tabular} \egroup
     
    \caption{\bd{Synthetic predictions visualization.} First, we downscale cropped ground truth masks to a fixed resolution (from $28 \x 28$ to $112 \x 112$) mask with continuous values, we then upscale it back using bilinear interpolation, and finally binarize it. Synthetic prediction with low effective resolution are close to the ground truth masks for smaller objects (top row), however the discrepancy grows with object size (bottom row).}
    \label{fig:synthetic_predictions}
\end{figure}

\paragraph{Evaluation on synthetic predictions.} We simulate predictions by capping the effective resolution of each mask. First, we downscale cropped ground truth masks to a fixed resolution mask with continuous values, we then upscale it back using bilinear interpolation, and finally binarize it. Figure~\ref{fig:synthetic_predictions} show visualization of the synthetic predictions with different effective resolutions. In Table~\ref{tab:coco_gt_mask_resolution_detail} we compare Mask AP and Boundary AP for the synthetic predictions with different synthetic scales across different datasets.

\paragraph{Evaluation on real predictions.} In addition to the experiments with COCO in the main text, we evaluate Mask R-CNN~\cite{he2017mask}, PointRend~\cite{kirillov2020pointrend}, and Boundary-preserving Mask R-CNN (BMask R-CNN)~\cite{ChengWHL20bmaskrcnn} on LVIS$^*$v0.5 and Cityscapes in Table~\ref{tab:inference_with_gt_box_detail}. For each method we feed ground truth boxes to isolate the segmentation quality aspect of the instance segmentation task. On all datasets we observe that Boundary AP better captures improvements in the mask quality. %

\input{tables/gt_box}

\input{tables/coco_instance}
\input{tables/lvis_instance}

\paragraph{Reference Boundary AP evaluation.}
We provide Boundary AP evaluation for various recent and classic models on COCO (Table~\ref{tab:coco_instance_seg}), LVIS (Table~\ref{tab:lvis_instance_seg}), and Cityscapes (Table~\ref{tab:cityscapes_instance_seg}) datasets. We do not train any models ourselves and use the Detectron2 framework~\cite{wu2019detectron2} or official implementations instead. These results can be used as a reference to simplify the comparison for future methods.

\input{tables/cityscapes_instance}

\subsection{Panoptic Segmentation}\label{sec:panoptic}

The standard evaluation metric for panoptic segmentation is panoptic quality (PQ or Mask PQ)~\cite{kirillov2017panoptic}, defined as:
$$\small{\text{PQ}} = \underbrace{\frac{\sum_{(p, g) \in \text{TP}} \text{IoU}(p, g)}{\vphantom{\frac{1}{2}}|\text{TP}|}}_{\text{segmentation quality (SQ)}} \x \underbrace{\frac{|\text{TP}|}{|\text{TP}| + \frac{1}{2} |\text{FP}| + \frac{1}{2} |\text{FN}|}}_{\text{recognition quality (RQ) }}$$

Mask IoU is presented in two places: (1) calculating the average Mask IoU for true positives in Segmentation Quality (SQ) component and (2) matching prediction and ground truth masks to split them into true positives, false positives, and false negatives. Similarly to Boundary AP, we replace Mask IoU with \textbf{$\text{min}(\text{Mask IoU}, \text{Boundary IoU})$} in both places and refer the new metric as Boundary PQ.

\input{tables/coco_panoptic}

\paragraph{Datasets.}
We use two popular datasets with panoptic annotation: COCO panoptic~\cite{kirillov2017panoptic} and Cityscapes~\cite{Cordts2016Cityscapes}.

\noindent \textit{COCO panoptic~\cite{kirillov2017panoptic}} combines annotations from COCO instance segmentation~\cite{lin2014coco} and COCO stuff segmentation~\cite{caesar2016coco} into a unified panoptic format with no overlaps. COCO panoptic has 80 things and 53 stuff categories.

\noindent \textit{Cityscapes~\cite{Cordts2016Cityscapes}} has 8 thing and 11 stuff categories.

Similar to the instance segmentation task, we set dilation width to $2\%$ image diagonal for COCO panoptic and $0.5\%$ image diagonal for Cityscapes.

\paragraph{Analysis with synthetic predictions.}
Following our experimental setup for instance segmentation, we evaluate Boundary PQ on low-fidelity synthetic predictions generated from ground truth annotations to avoid any potential bias toward a specific model. The synthetic predictions are generated by downscaling ground truth panoptic segmentation maps for each image and then upscaling it back using nearest neighbor interpolation in both cases. This image-level generation process ensures a unified treatment of both things and stuff segments following the idea behind the panoptic segmentation task. 

In Table~\ref{tab:coco_panoptic_gt_output_resolution}, we report Panoptic Quality and its two components: Segmentation Quality (SQ) and Recognition Quality (RQ) for synthetic predictions with various downscaling ratios across different datasets. Similar to our findings for AP, Boundary PQ better tracks boundary quality improvements than Mask PQ for panoptic segmentation. Furthermore, we find that the difference between Boundary PQ and Mask PQ is mainly caused by the difference in SQ. This observation confirms that Boundary IoU better tracks the mask quality of predictions and does not significantly change other aspects like the matching procedure between prediction and ground truth segments.

\input{tables/coco_panoptic_gt_output_resolution}

\paragraph{References Boundary PQ evaluation.}
We provide Boundary PQ evaluation for various models on COCO panoptic and Cityscapes datasets in Table~\ref{tab:coco_panoptic_seg}. We do not train any models ourselves and use models trained by their authors. These results can be used as a reference to simplify the comparison for future methods.

%% file: fig_tex/analysis/trimap_iou/hole.tex
\begin{figure}[h!]
    \centering
    \bgroup 
    \def\arraystretch{0.2} 
    \setlength\tabcolsep{2pt}
    \begin{tabular}{cc}
    \multirow{2}{*}{\rotatebox[origin=c]{90}{\small Measure value \hspace{-110pt}}} & \includegraphics[width=0.55\linewidth,trim={1.0cm 1.1cm 0cm 0cm},clip]{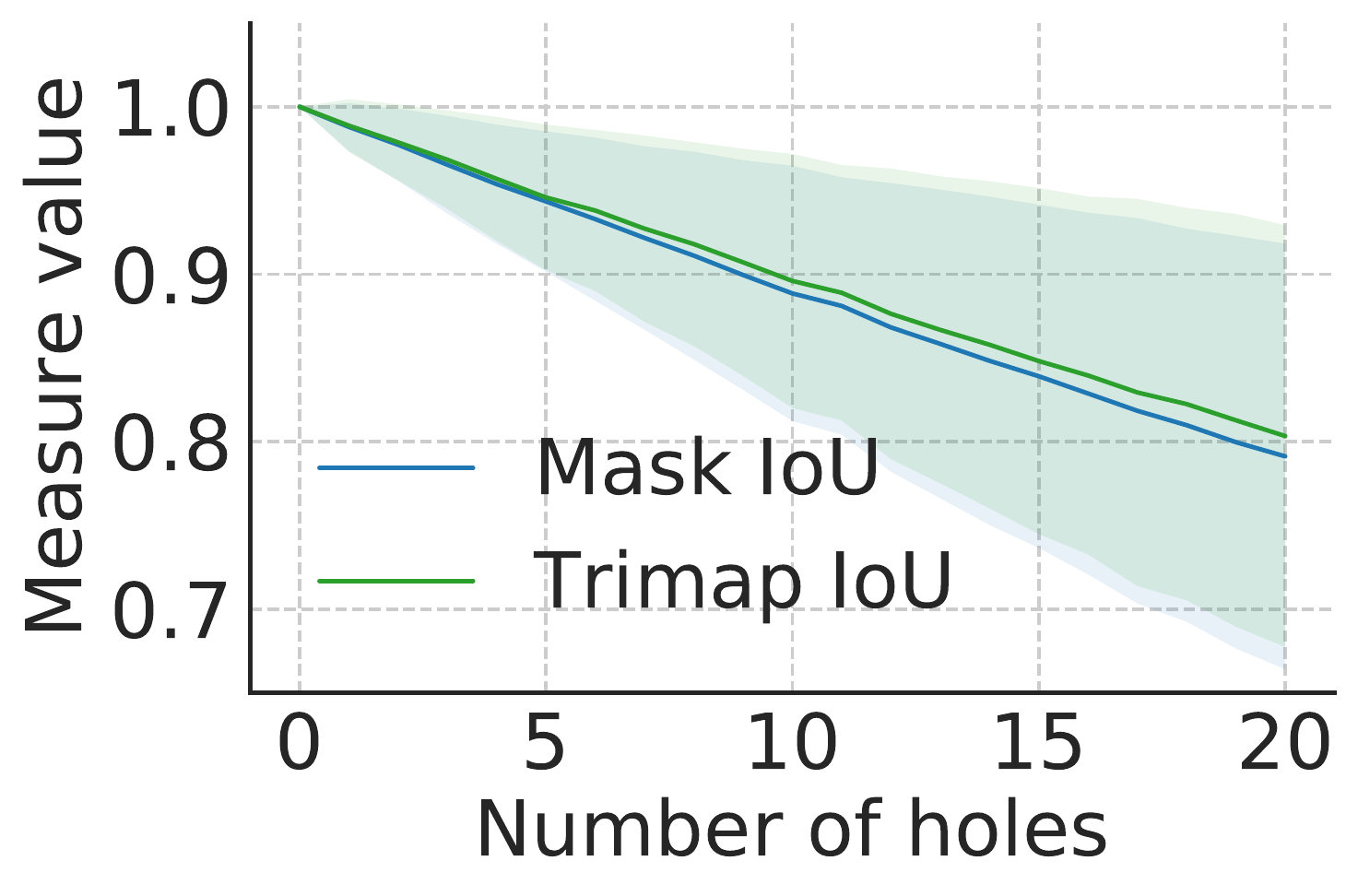} \\
    & \small Number of holes
    \end{tabular} \egroup
     
    \caption{\bd{Sensitivity analysis for Trimap IoU.} The measure penalizes inner mask errors less than Mask IoU.}
    \label{fig:trimap_iou_hole}
\end{figure}

%% file: fig_tex/analysis/mf_measure/dilation_erosion.tex
\begin{figure}[h!]
    \centering
    \bgroup 
    \def\arraystretch{0.6} 
    \setlength\tabcolsep{2pt}
    \begin{tabular}{ccc}
    \multirow{3}{*}{\rotatebox[origin=c]{90}{\small Measure value \hspace{8pt}}} & \small object area $> 96^2$ & \small object area $\le 32^2$  \\
    & \includegraphics[width=0.49\linewidth,trim={1.0cm 1.1cm 0cm 0cm},clip]{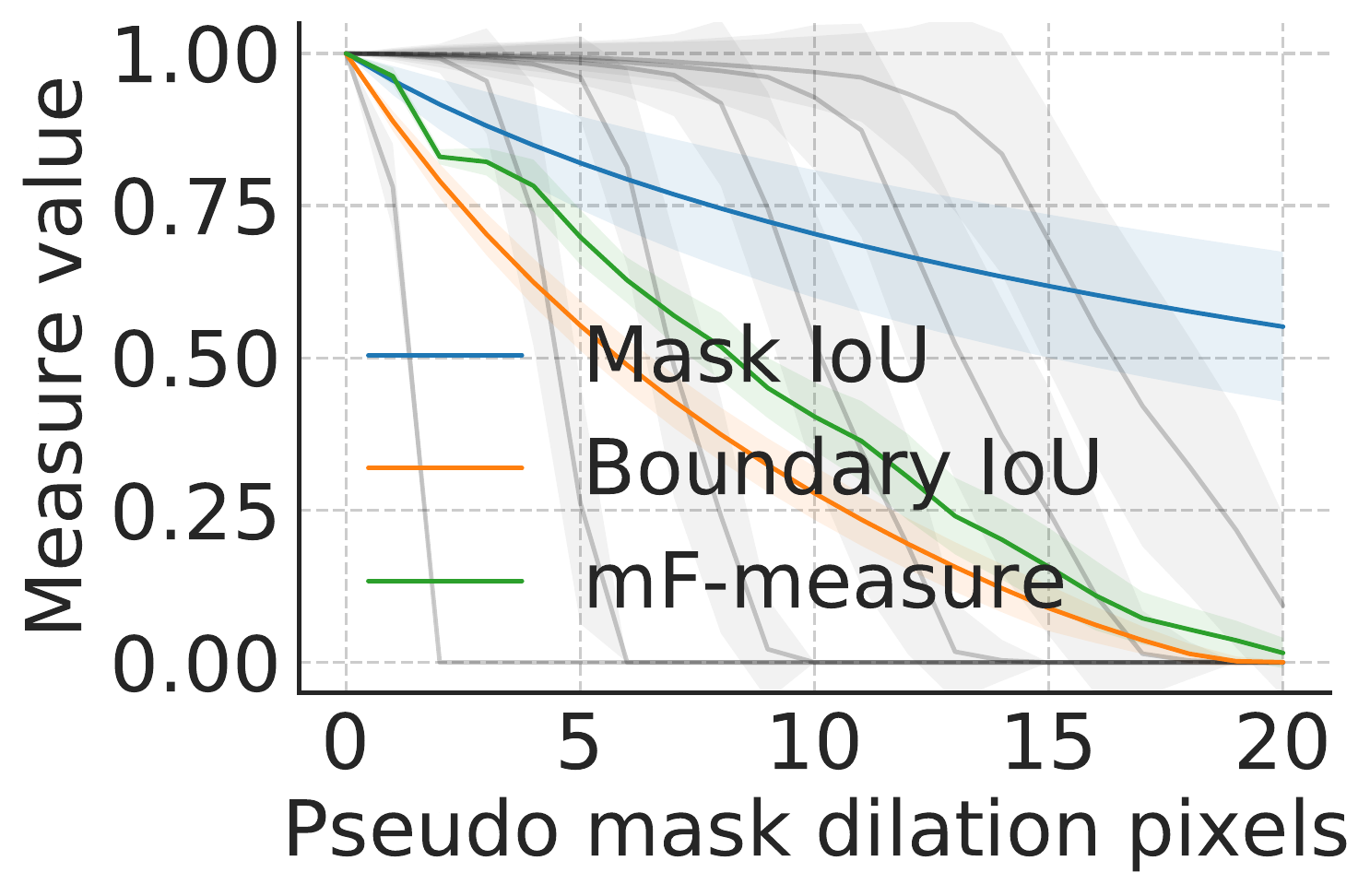} &
    \includegraphics[width=0.49\linewidth,trim={1.0cm 1.1cm 0cm 0cm},clip]{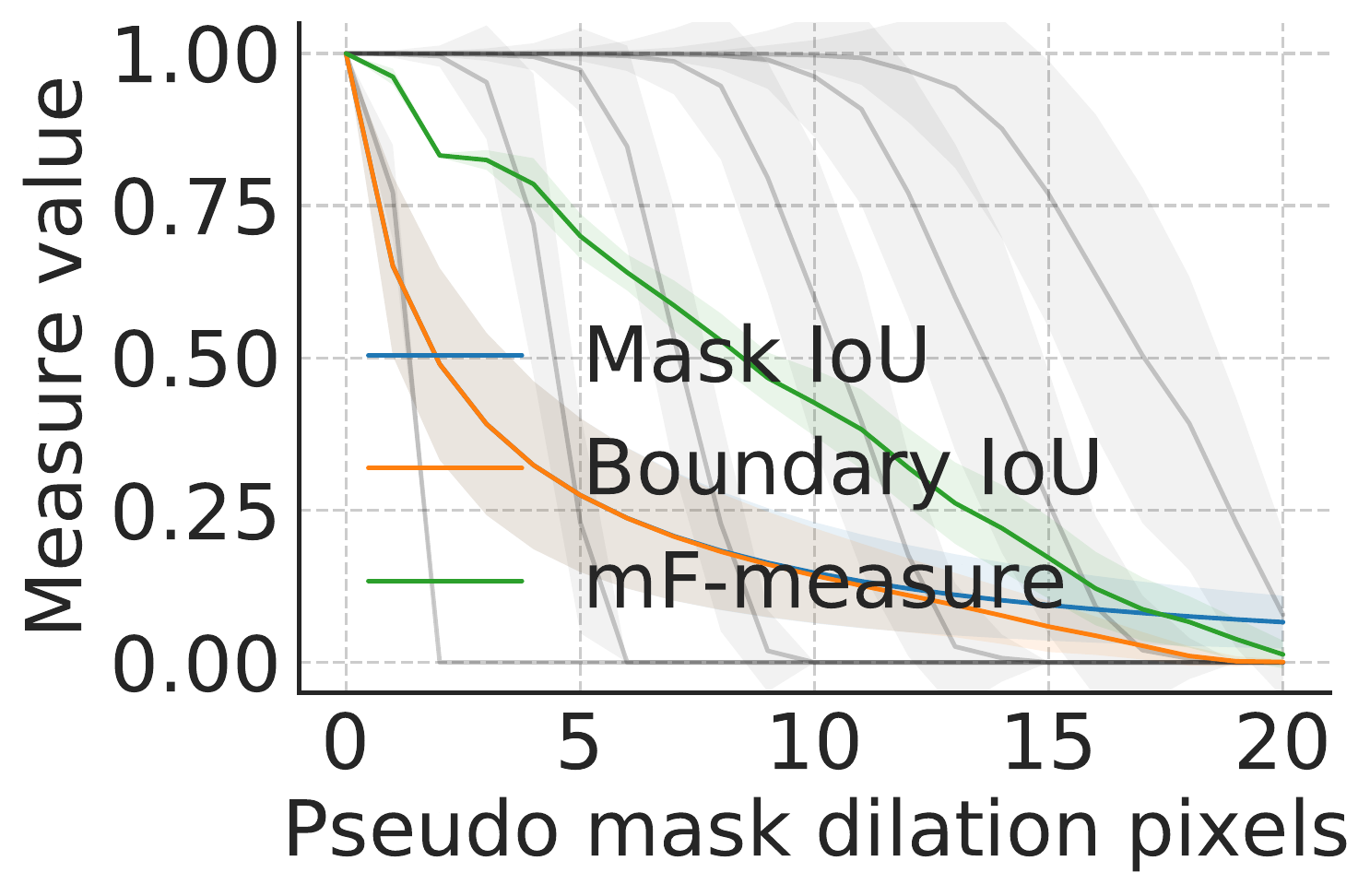} \\
    & \multicolumn{2}{c}{\small Dilation (pixels)} \\
    \end{tabular} \egroup 
    \caption{\bd{Sensitivity analysis for mean F-measure (mF-measure).} The measure under-penalizes scale type (dilation) errors in small objects in comparison with Boundary IoU that matches Mask IoU behavior for such objects. F-measure curves for different threshold parameters $d$ are shown in gray.}
    \label{fig:mf_measure_dilation_erosion}
\end{figure}

%% file: tables/coco_gt_mask_resolution.tex
\begin{table*}[h!]
  \centering
  \tablestyle{3pt}{1.2}
  \begin{tabular}{cc|x{22}x{22}x{22}x{22}|x{22}x{22}x{22}x{22}|x{22}x{22}x{22}x{22}}
    &  & \multicolumn{4}{c|}{COCO~\cite{lin2014coco}} & \multicolumn{4}{c|}{LVIS$^*$v0.5~\cite{gupta2019lvis}} & \multicolumn{4}{c}{Cityscapes~\cite{Cordts2016Cityscapes}} \\
    Mask resolution & Evaluation metric & AP & AP$_S$ & AP$_M$ &  AP$_L$ & AP & AP$_S$ & AP$_M$ &  AP$_L$ & AP & AP$_S$ & AP$_M$ &  AP$_L$ \\
    \shline \rule{0mm}{5mm}
    \multirow{2}{*}{$28\x28$} & Mask AP & 96.5 & 98.9 & 95.7 & 95.1 & 94.3 & 96.7 & 93.7 & 93.9 & 93.5 & 98.4 & 91.6 & 91.4 \\
    & Boundary AP & 85.9 & 98.9 & 93.0 & 73.0 & 85.5 & 96.7 & 91.4 & 73.1 & 75.9 & 98.4 & 86.4 & 55.9 \\[3mm]
    \hline \rule{0mm}{5mm}
    \multirow{2}{*}{$56\x56$} & Mask AP & 99.5 & 99.8 & 99.4 & 99.3 & 98.2 & 98.4 & 99.0 & 98.9 & 98.9 & 99.7 & 99.0 & 98.7 \\
    & Boundary AP & 95.2 & 99.8 & 99.3 & 89.5 & 94.6 & 98.4 & 98.8 & 89.9 & 90.9 & 99.7 & 97.7 & 80.5 \\[3mm]
    \hline \rule{0mm}{5mm}
    \multirow{2}{*}{$112\x112$} & Mask AP & 99.9 & 100.0 & 99.9 & 99.9 & 98.8 & 98.5 & 99.7 & 99.9 & 99.8 & 99.9 & 99.9 & 99.9 \\
    & Boundary AP & 99.0 & 100.0 & 99.9 & 97.9 & 98.1 & 98.5 & 99.7 & 98.3 & 97.2 & 99.9 & 99.9 & 94.9 \\
  \end{tabular}

\caption{Boundary AP and Mask AP on COCO \texttt{val} set, LVIS$^*$v0.5 \texttt{val} set and Cityscapes \texttt{val} set for synthetic $28 \x 28$, $56 \x 56$, and $112 \x 112$ predictions generated from the ground truth. Unlike Mask AP, Boundary AP$_L$ successfully captures the lack of fidelity in the synthetic prediction with lower effective resolution for large objects that have area $> 96^2$.}
\label{tab:coco_gt_mask_resolution_detail}
\end{table*}

%% file: tables/gt_box.tex
\begin{table}[!h]
  \begin{subtable}{1.0\linewidth}
    \centering
    \tablestyle{1.5pt}{1.1}
    \begin{tabular}{c | x{34}x{34}x{38}x{34}x{38}}
      Method & AP\scriptsize{$^{\text{mask}}$} & AP\scriptsize{$^{\text{boundary}}$} & AP\scriptsize{$^{\text{boundary}}_S$} & AP\scriptsize{$^{\text{boundary}}_M$} & AP\scriptsize{$^{\text{boundary}}_L$} \\
      \shline
      Mask R-CNN  & 51.5 \phantom{\dt{+0.1}} & 38.3 \phantom{\dt{+0.1}} & 45.4 \phantom{\dt{+0.1}} & 48.7 \phantom{\dt{+0.1}} & 29.2 \phantom{\dt{+00.1}} \\
      PointRend   & 56.8 \dt{+5.3}           & 45.9 \dt{+7.6}           & 49.6 \dt{+4.2}           & 56.0 \dt{+7.3}           & 42.2 \dt{\bd{+13.0}} \\
      BMask R-CNN & 57.8 \bd{\dt{+6.3}}      & 46.1 \bd{\dt{+7.8}}      & 50.8 \bd{\dt{+5.4}}      & 56.3 \bd{\dt{+7.6}}      & 40.7 \dt{+11.5} \\
    \end{tabular}
    \subcaption{The models are trained on COCO and evaluated on LVIS$^*$v0.5 \texttt{val} set, which has higher annotation quality.}
    \vspace{4mm}
  \end{subtable}
  \begin{subtable}{1.0\linewidth}
    \centering
    \tablestyle{1.5pt}{1.1}
    \begin{tabular}{c | x{34}x{34}x{38}x{34}x{38}}
      Method & AP\scriptsize{$^{\text{mask}}$} & AP\scriptsize{$^{\text{boundary}}$} & AP\scriptsize{$^{\text{boundary}}_S$} & AP\scriptsize{$^{\text{boundary}}_M$} & AP\scriptsize{$^{\text{boundary}}_L$} \\
      \shline
      Mask R-CNN  & 35.5 \phantom{\dt{+0.1}} & 16.4 \phantom{\dt{+0.1}} & 22.3 \phantom{\dt{+0.01}} & 22.0 \phantom{\dt{+0.1}} & 9.7 \phantom{\dt{+0.01}} \\
      PointRend   & 42.2 \dt{+6.7}           & 23.6 \dt{+7.2}           & 29.9 \dt{\phantom{0}+7.6} & 29.0 \dt{+7.0}           & 19.8 \dt{\bd{+10.1}} \\
      BMask R-CNN & 43.3 \bd{\dt{+7.8}}      & 24.0 \bd{\dt{+7.6}}      & 36.0 \bd{\dt{+13.7}}      & 29.4 \bd{\dt{+7.4}}      & 18.7 \dt{\phantom{0}+9.0} \\
    \end{tabular}
    \subcaption{The models are trained and evaluated on Cityscapes.}
  \end{subtable}
\caption{Mask R-CNN comparison with the methods designed to improve the mask quality. All models are fed with ground truth boxes. Boundary AP better captures improvements in the mask quality. BMask R-CNN, which outputs $28 \x 28$ resolution predictions, outperforms PointRend for smaller objects but trails it for large objects where $224 \x 224$ output resolution of PointRend improves boundary quality.}
\label{tab:inference_with_gt_box_detail}
\end{table}

%% file: tables/coco_instance.tex
\begin{table*}[!t]
  \centering
  \tablestyle{3pt}{1.1}
  \begin{tabular}{c c c|x{22}x{22}x{22}x{22}x{22}x{22}|x{22}x{22}x{22}x{22}x{22}x{22}}
    &  &  & \multicolumn{6}{c|}{Mask AP} & \multicolumn{6}{c}{Boundary AP} \\
    Name & Backbone & LRS & AP & AP$_{50}$ & AP$_{75}$ & AP$_S$ &  AP$_M$ &  AP$_L$ & AP & AP$_{50}$ & AP$_{75}$ & AP$_S$ &  AP$_M$ &  AP$_L$ \\
    \shline \rule{0mm}{5mm}
    \multirow{4}{*}{Mask R-CNN~\cite{he2017mask}} & R50~\cite{he2016deep} & $1\x$ & 35.2 & 56.3 & 37.5 & 17.2 & 37.7 & 50.3 & 21.2 & 46.4 & 16.8 & 17.1 & 31.6 & 19.7 \\
    & R50~\cite{he2016deep} & $3\x$ & 37.2 & 58.6 & 39.9 & 18.6 & 39.5 & 53.3 & 23.1 & 49.6 & 19.0 & 18.6 & 33.4 & 22.2 \\
    & R101~\cite{he2016deep} & $3\x$ & 38.6 & 60.4 & 41.3 & 19.5 & 41.3 & 55.3 & 24.5 & 51.7 & 20.3 & 19.4 & 35.0 & 23.9 \\
    & X101-32$\x$8d~\cite{xie2017aggregated} & $3\x$ & 39.5 & 61.7 & 42.6 & 20.7 & 42.0 & 56.5 & 25.4 & 53.2 & 21.0 & 20.6 & 35.8 & 24.7 \\[2.5mm]
    \hline \rule{0mm}{5mm}
    Mask R-CNN with & R50~\cite{he2016deep} & $1\x$ & 37.5 & 59.4 & 40.2 & 18.4 & 39.7 & 54.8 & 22.8 & 49.6 & 18.1 & 18.3 & 33.4 & 22.1 \\
    deformable conv.~\cite{zhu2019deformable} & R50~\cite{he2016deep} & $3\x$ & 38.5 & 60.8 & 41.1 & 19.7 & 40.6 & 55.7 & 24.1 & 51.8 & 19.4 & 19.6 & 34.4 & 23.4 \\[2.5mm]
    \hline \rule{0mm}{5mm}
    Mask R-CNN with & R50~\cite{he2016deep} & $1\x$ & 36.4 & 56.9 & 39.2 & 17.5 & 38.7 & 52.5 & 22.5 & 47.9 & 18.7 & 17.5 & 32.6 & 21.7 \\
    cascade box head~\cite{cai2018cascade} & R50~\cite{he2016deep} & $3\x$ & 38.5 & 59.6 & 41.5 & 19.5 & 41.1 & 54.5 & 24.5 & 51.2 & 20.8 & 19.5 & 34.8 & 23.6 \\[2.5mm]
    \hline \rule{0mm}{5mm}
    \multirow{4}{*}{PointRend~\cite{kirillov2020pointrend}} & R50~\cite{he2016deep} & $1\x$ & 36.2 & 56.6 & 38.6 & 17.1 & 38.8 & 52.5 & 23.5 & 48.4 & 20.2 & 17.1 & 33.0 & 24.1 \\
    & R50~\cite{he2016deep} & $3\x$ & 38.3 & 59.1 & 41.1 & 19.1 & 40.7 & 55.8 & 25.4 & 51.3 & 22.3 & 19.1 & 34.8 & 26.4 \\
    & R101~\cite{he2016deep} & $3\x$ & 40.1 & 61.1 & 43.0 & 20.0 & 42.9 & 58.6 & 27.0 & 54.1 & 24.2 & 19.9 & 37.0 & 28.7 \\
    & X101-32$\x$8d~\cite{xie2017aggregated} & $3\x$ & 41.1 & 62.8 & 44.2 & 21.5 & 43.8 & 59.1 & 28.0 & 55.6 & 25.3 & 21.5 & 37.8 & 29.1 \\[2.5mm]
    \hline \rule{0mm}{3mm}
    \multirow{2}{*}{BMask R-CNN~\cite{ChengWHL20bmaskrcnn}} & R50~\cite{he2016deep} & $1\x$ & 36.6 & 56.7 & 39.4 & 17.3 & 38.8 & 53.8 & 23.5 & 48.4 & 20.2 & 17.2 & 33.0 & 24.5 \\
    & R50~\cite{he2016deep} & $3\x$ & 38.6 & 59.2 & 41.7 & 19.6 & 41.1 & 55.7 & 25.4 & 51.4 & 22.3 & 19.5 & 35.2 & 26.3 \\
  \end{tabular}
\caption{Boundary AP for recent and classic models on COCO \texttt{val}. All models are based on Detectron2. LRS: learning rate schedule, a $1\x$ learning rate schedule refers to 90,000 iterations and a $3\x$ learning rate schedule refers to 270,000 iterations, with batch size 16.}
\label{tab:coco_instance_seg}
\end{table*}

%% file: tables/lvis_instance.tex
\begin{table*}[!t]
  \centering
  \tablestyle{1pt}{1.1}
  \begin{tabular}{c|x{22}x{22}x{22}x{22}x{22}x{22}x{22}x{22}x{22}|x{22}x{22}x{22}x{22}x{22}x{22}x{22}x{22}x{22}}
    & \multicolumn{9}{c|}{Mask AP} & \multicolumn{9}{c}{Boundary AP} \\
    Backbone & AP & AP$_{50}$ & AP$_{75}$ & AP$_S$ &  AP$_M$ &  AP$_L$ & AP$_r$ &  AP$_c$ &  AP$_f$ & AP & AP$_{50}$ & AP$_{75}$ & AP$_S$ &  AP$_M$ &  AP$_L$ & AP$_r$ &  AP$_c$ &  AP$_f$ \\
    \shline
    R50~\cite{he2016deep} & 24.4 & 37.7 & 26.0 & 16.7 & 31.2 & 41.2 & 16.0 & 24.0 & 28.3  &  18.8 & 33.9 & 18.0 & 16.7 & 28.0 & 19.3 & 11.9 & 18.2 & 22.3 \\
    R101~\cite{he2016deep} & 25.8 & 39.7 & 27.3 & 17.6 & 33.0 & 43.7 & 15.5 & 26.0 & 29.6  &  20.1 & 35.2 & 19.8 & 17.6 & 29.9 & 20.8 & 11.8 & 20.1 & 23.5 \\
    X101-32$\x$8d~\cite{xie2017aggregated} & 27.0 & 41.4 & 28.7 & 19.0 & 35.1 & 43.7 & 15.4 & 27.3 & 31.3  &  21.4 & 37.6 & 21.2 & 19.0 & 32.0 & 21.7 & 11.2 & 21.6 & 25.1
  \end{tabular}
\caption{Boundary AP of Mask R-CNN baselines on LVISv0.5 \texttt{val}. All models are from the Detectron2 model zoo.}
\label{tab:lvis_instance_seg}
\end{table*}

%% file: tables/cityscapes_instance.tex
\begin{table}[h!]
  \centering
  \tablestyle{3pt}{1.1}
  \begin{tabular}{c c|x{22}x{22}|x{22}x{22}}
     &  & \multicolumn{2}{c|}{Mask AP} & \multicolumn{2}{c}{Boundary AP} \\
    Method & Backbone & AP & AP$_{50}$ & AP & AP$_{50}$ \\
    \shline
    Mask R-CNN~\cite{he2017mask} & R50~\cite{he2016deep} & 33.8 & 61.5 & 11.4 & 37.4 \\
    PointRend~\cite{kirillov2020pointrend} & R50~\cite{he2016deep} & 35.9 & 61.8 & 16.7 & 47.2 \\
    BMask R-CNN~\cite{ChengWHL20bmaskrcnn} & R50~\cite{he2016deep} & 36.2 & 62.6 & 15.7 & 46.2 \\
    Panoptic-DeepLab~\cite{cheng2020panoptic} & X71~\cite{chollet2017xception} & 35.3 & 57.9 & 16.5 & 47.7 \\
  \end{tabular}
\caption{Boundary AP evaluation on Cityscapes \texttt{val} set for models implemented in Detectron2~\cite{wu2019detectron2}. Note that we set the dilation width to $0.5\%$ image diagonal for Cityscapes.}
\label{tab:cityscapes_instance_seg}
\end{table}

%% file: tables/coco_panoptic.tex
\begin{table*}[h!]
  \vspace{1cm}
  \centering
  \tablestyle{3pt}{1.1}
  \begin{tabular}{c | c c|x{22}x{22}x{22}|x{22}x{22}x{22}}
    & & & \multicolumn{3}{c|}{Mask PQ} & \multicolumn{3}{c}{Boundary PQ} \\
    Dataset & Method & Backbone & PQ & SQ & RQ & PQ & SQ & RQ \\
    \shline \rule{0mm}{5mm}
    \multirow{6}{*}{COCO panoptic~\cite{kirillov2017panoptic}} & \multirow{3}{*}{Panoptic FPN~\cite{kirillov2019panopticfpn}} & R50~\cite{he2016deep} & 41.5 & 79.1 & 50.5 & 30.8 & 70.0 & 41.7 \\
    & & R101~\cite{he2016deep} & 43.0 & 80.0 & 52.1 & 32.5 & 70.9 & 43.7 \\
    & & X101-32$\x$8d~\cite{xie2017aggregated} & 44.4 & 80.4 & 53.8 & 33.9 & 71.4 & 45.5 \\[2.5mm]
    \cline{2-9} \rule{0mm}{5mm}
    & UPSNet~\cite{xiong19upsnet} & R50~\cite{he2016deep} & 42.5 & 78.2 & 52.4 & 31.0 & 68.7 & 43.3 \\
    & DETR~\cite{detr} & R50~\cite{he2016deep} & 43.4 & 79.3 & 53.8 & 32.8 & 71.0 & 45.2  \\[2.5mm]
    \shline \rule{0mm}{5mm}
    \multirow{3}{*}{Cityscapes~\cite{Cordts2016Cityscapes}} & UPSNet~\cite{xiong19upsnet} & R50~\cite{he2016deep} & 59.4 & 79.7 & 73.1 & 33.4 & 63.1 & 51.9 \\[2.5mm]
    \cline{2-9} \rule{0mm}{5mm}
    & \multirow{2}{*}{Panoptic-DeepLab~\cite{cheng2020panoptic}} & R50~\cite{he2016deep} & 59.8 & 80.0 & 73.5 & 36.3 & 64.3 & 55.6 \\
    & & X71~\cite{chollet2017xception} & 63.0 & 81.7 & 76.2 & 41.0 & 65.5 & 61.7 \\
  \end{tabular}
\caption{Reference Boundary PQ evaluation for various models on COCO panoptic \texttt{val} and Cityscapes \texttt{val}.}
\label{tab:coco_panoptic_seg}
\end{table*}

%% file: tables/coco_panoptic_gt_output_resolution.tex
\begin{table}[!h]
  \centering
  \tablestyle{2pt}{1.1}
  \begin{tabular}{cc|x{18}x{18}x{18}|x{18}x{18}x{18}}
    Downscaling &  Evaluation & \multicolumn{3}{c|}{COCO panoptic~\cite{kirillov2017panoptic}} & \multicolumn{3}{c}{Cityscapes~\cite{Cordts2016Cityscapes}} \\
    ratio & metric & PQ & SQ & RQ & PQ & SQ & RQ \\
    \shline \rule{0mm}{5mm}
    \multirow{2}{*}{$8$} &
    Mask PQ & 62.6 & 78.5 & 78.4 & 66.3 & 77.8 & 83.7 \\
    & Boundary PQ & 52.8 & 68.1 & 77.0 & 47.1 & 58.6 & 80.2 \\[2.5mm]
    \hline \rule{0mm}{5mm}
    \multirow{2}{*}{$4$} &
    Mask PQ & 81.0 & 85.9 & 93.7 & 84.3 & 85.7 & 98.2 \\
    & Boundary PQ & 76.6 & 81.4 & 93.7 & 75.0 & 76.3 & 98.2 \\[2.5mm]
    \hline \rule{0mm}{5mm}
    \multirow{2}{*}{$2$} &
    Mask PQ & 92.5 & 93.4 & 99.0 & 94.2 & 94.2 & 99.9 \\
    & Boundary PQ & 90.8 & 91.6 & 99.0 & 90.7 & 90.7 & 99.9 \\
  \end{tabular}
  
\caption{Boundary PQ and Mask PQ evaluated on COCO panoptic \texttt{val} and Cityscapes \texttt{val} sets for synthetic prediction with $8$, $4$, and $2$ downscaling ratios generated from the ground truth. Boundary PQ is more sensitive than Mask PQ in its Segmentation Quality (SQ) component while the Recognition Quality (RQ) component is comparable.}
\label{tab:coco_panoptic_gt_output_resolution}
\end{table}

%% file: boundary_iou.bbl
\begin{thebibliography}{10}\itemsep=-1pt

\bibitem{bodla2017soft}
Navaneeth Bodla, Bharat Singh, Rama Chellappa, and Larry~S Davis.
\newblock Soft-{NMS}--improving object detection with one line of code.
\newblock In {\em ICCV}, 2017.

\bibitem{bolya2020tide}
Daniel Bolya, Sean Foley, James Hays, and Judy Hoffman.
\newblock {TIDE}: A general toolbox for identifying object detection errors.
\newblock In {\em ECCV}, 2020.

\bibitem{caesar2016coco}
Holger Caesar, Jasper Uijlings, and Vittorio Ferrari.
\newblock {COCO-Stuff}: Thing and stuff classes in context.
\newblock In {\em CVPR}, 2018.

\bibitem{cai2018cascade}
Zhaowei Cai and Nuno Vasconcelos.
\newblock {Cascade R-CNN}: Delving into high quality object detection.
\newblock In {\em CVPR}, 2018.

\bibitem{detr}
Nicolas Carion, Francisco Massa, Gabriel Synnaeve, Nicolas Usunier, Alexander
  Kirillov, and Sergey Zagoruyko.
\newblock End-to-end object detection with transformers.
\newblock In {\em ECCV}, 2020.

\bibitem{deeplabV2}
Liang-Chieh Chen, George Papandreou, Iasonas Kokkinos, Kevin Murphy, and Alan~L
  Yuille.
\newblock {DeepLab}: Semantic image segmentation with deep convolutional nets,
  atrous convolution, and fully connected {CRF}s.
\newblock {\em PAMI}, 2018.

\bibitem{cheng2020panoptic}
Bowen Cheng, Maxwell~D Collins, Yukun Zhu, Ting Liu, Thomas~S Huang, Hartwig
  Adam, and Liang-Chieh Chen.
\newblock {Panoptic-DeepLab}: A simple, strong, and fast baseline for bottom-up
  panoptic segmentation.
\newblock In {\em CVPR}, 2020.

\bibitem{ChengWHL20bmaskrcnn}
Tianheng Cheng, Xinggang Wang, Lichao Huang, and Wenyu Liu.
\newblock Boundary-preserving {Mask R-CNN}.
\newblock In {\em ECCV}, 2020.

\bibitem{chollet2017xception}
Fran{\c{c}}ois Chollet.
\newblock Xception: Deep learning with depthwise separable convolutions.
\newblock In {\em CVPR}, 2017.

\bibitem{Cordts2016Cityscapes}
Marius Cordts, Mohamed Omran, Sebastian Ramos, Timo Rehfeld, Markus Enzweiler,
  Rodrigo Benenson, Uwe Franke, Stefan Roth, and Bernt Schiele.
\newblock The {Cityscapes} dataset for semantic urban scene understanding.
\newblock In {\em CVPR}, 2016.

\bibitem{csurka2013good}
Gabriela Csurka, Diane Larlus, Florent Perronnin, and France Meylan.
\newblock What is a good evaluation measure for semantic segmentation?.
\newblock In {\em BMVC}, 2013.

\bibitem{dai2016instance}
Jifeng Dai, Kaiming He, and Jian Sun.
\newblock Instance-aware semantic segmentation via multi-task network cascades.
\newblock In {\em CVPR}, 2016.

\bibitem{everingham2015pascal}
Mark Everingham, SM~Ali Eslami, Luc Van~Gool, Christopher~KI Williams, John
  Winn, and Andrew Zisserman.
\newblock The {PASCAL} visual object classes challenge: A retrospective.
\newblock {\em IJCV}, 2015.

\bibitem{pascal-voc-2007}
M. Everingham, L. Van~Gool, C.~K.~I. Williams, J. Winn, and A. Zisserman.
\newblock The {PASCAL} {V}isual {O}bject {C}lasses {C}hallenge 2007 {(VOC2007)}
  {R}esults.
\newblock
  \url{http://www.pascal-network.org/challenges/VOC/voc2007/workshop/index.html},
  2007.

\bibitem{shapely}
Sean Gillies et~al.
\newblock Shapely: manipulation and analysis of geometric objects, 2007.

\bibitem{gupta2019lvis}
Agrim Gupta, Piotr Dollar, and Ross Girshick.
\newblock {LVIS}: A dataset for large vocabulary instance segmentation.
\newblock In {\em ICCV}, 2019.

\bibitem{he2017mask}
Kaiming He, Georgia Gkioxari, Piotr Doll{\'a}r, and Ross Girshick.
\newblock Mask {R-CNN}.
\newblock In {\em ICCV}, 2017.

\bibitem{he2016deep}
Kaiming He, Xiangyu Zhang, Shaoqing Ren, and Jian Sun.
\newblock Deep residual learning for image recognition.
\newblock In {\em CVPR}, 2016.

\bibitem{huang2019mask}
Zhaojin Huang, Lichao Huang, Yongchao Gong, Chang Huang, and Xinggang Wang.
\newblock Mask scoring {R-CNN}.
\newblock In {\em CVPR}, 2019.

\bibitem{kirillov2019panopticfpn}
Alexander Kirillov, Ross Girshick, Kaiming He, and Piotr Doll{\'a}r.
\newblock Panoptic feature pyramid networks.
\newblock In {\em CVPR}, 2019.

\bibitem{kirillov2017panoptic}
Alexander Kirillov, Kaiming He, Ross Girshick, Carsten Rother, and Piotr
  Doll{\'a}r.
\newblock Panoptic segmentation.
\newblock In {\em CVPR}, 2019.

\bibitem{kirillov2020pointrend}
Alexander Kirillov, Yuxin Wu, Kaiming He, and Ross Girshick.
\newblock {PointRend}: Image segmentation as rendering.
\newblock In {\em CVPR}, 2020.

\bibitem{kohli2009robust}
Pushmeet Kohli, Philip~HS Torr, et~al.
\newblock Robust higher order potentials for enforcing label consistency.
\newblock {\em IJCV}, 2009.

\bibitem{li2020joint}
Zeming Li, Yuchen Ma, Yukang Chen, Xiangyu Zhang, and Jian Sun.
\newblock Joint coco and mapillary workshop at iccv 2019: Coco instance
  segmentation challenge track.
\newblock {\em arXiv:2010.02475}, 2020.

\bibitem{coco2018winner}
Zeming Li, Yueqing Zhuang, Xiangyu Zhang, Gang Yu, and Jian Sun.
\newblock {COCO} instance segmentation challenges 2018: winner.
\newblock
  \url{http://presentations.cocodataset.org/ECCV18/COCO18-Detect-Megvii.pdf},
  2018.

\bibitem{liang2020polytransform}
Justin Liang, Namdar Homayounfar, Wei-Chiu Ma, Yuwen Xiong, Rui Hu, and Raquel
  Urtasun.
\newblock Polytransform: Deep polygon transformer for instance segmentation.
\newblock In {\em CVPR}, 2020.

\bibitem{liberman2015}
Mark Liberman.
\newblock Reproducible research and the common task method.
\newblock
  \url{https://www.simonsfoundation.org/event/reproducible-research-and-the-common-task-method},
  2015.

\bibitem{lin2014coco}
Tsung-Yi Lin, Michael Maire, Serge Belongie, James Hays, Pietro Perona, Deva
  Ramanan, Piotr Doll{\'a}r, and C~Lawrence Zitnick.
\newblock Microsoft {COCO}: Common objects in context.
\newblock In {\em ECCV}, 2014.

\bibitem{marin2019efficient}
Dmitrii Marin, Zijian He, Peter Vajda, Priyam Chatterjee, Sam Tsai, Fei Yang,
  and Yuri Boykov.
\newblock Efficient segmentation: Learning downsampling near semantic
  boundaries.
\newblock In {\em ICCV}, 2019.

\bibitem{martin2003empirical}
David~Royal Martin.
\newblock {\em An empirical approach to grouping and segmentation}.
\newblock University of California Berkeley, 2003.

\bibitem{martin2004learning}
David~R Martin, Charless~C Fowlkes, and Jitendra Malik.
\newblock Learning to detect natural image boundaries using local brightness,
  color, and texture cues.
\newblock {\em PAMI}, 2004.

\bibitem{perazzi2016davis}
F. Perazzi, J. Pont-Tuset, B. McWilliams, L. {Van Gool}, M. Gross, and A.
  Sorkine-Hornung.
\newblock A benchmark dataset and evaluation methodology for video object
  segmentation.
\newblock In {\em CVPR}, 2016.

\bibitem{sengupta2020synthetic}
Akash Sengupta, Ignas Budvytis, and Roberto Cipolla.
\newblock Synthetic training for accurate 3d human pose and shape estimation in
  the wild.
\newblock In {\em BMVC}, 2020.

\bibitem{sharma2020compositional}
Akash Sharma, Wei Dong, and Michael Kaess.
\newblock Compositional scalable object {SLAM}.
\newblock {\em arXiv:2011.02658}, 2020.

\bibitem{wang2019object}
Zian Wang, David Acuna, Huan Ling, Amlan Kar, and Sanja Fidler.
\newblock Object instance annotation with deep extreme level set evolution.
\newblock In {\em CVPR}, 2019.

\bibitem{wu2019detectron2}
Yuxin Wu, Alexander Kirillov, Francisco Massa, Wan-Yen Lo, and Ross Girshick.
\newblock Detectron2.
\newblock \url{https://github.com/facebookresearch/detectron2}, 2019.

\bibitem{xie2017aggregated}
Saining Xie, Ross Girshick, Piotr Doll{\'a}r, Zhuowen Tu, and Kaiming He.
\newblock Aggregated residual transformations for deep neural networks.
\newblock In {\em CVPR}, 2017.

\bibitem{xiong19upsnet}
Yuwen Xiong, Renjie Liao, Hengshuang Zhao, Rui Hu, Min Bai, Ersin Yumer, and
  Raquel Urtasun.
\newblock Upsnet: A unified panoptic segmentation network.
\newblock In {\em CVPR}, 2019.

\bibitem{zhang2020perceiving}
Jason~Y Zhang, Sam Pepose, Hanbyul Joo, Deva Ramanan, Jitendra Malik, and
  Angjoo Kanazawa.
\newblock Perceiving 3d human-object spatial arrangements from a single image
  in the wild.
\newblock In {\em ECCV}, 2020.

\bibitem{zhou2017ade20k}
Bolei Zhou, Hang Zhao, Xavier Puig, Sanja Fidler, Adela Barriuso, and Antonio
  Torralba.
\newblock Scene parsing through {ADE20K} dataset.
\newblock In {\em CVPR}, 2017.

\bibitem{zhu2019deformable}
Xizhou Zhu, Han Hu, Stephen Lin, and Jifeng Dai.
\newblock Deformable convnets v2: More deformable, better results.
\newblock In {\em CVPR}, 2019.

\end{thebibliography}
